\documentclass{article}
\usepackage[numbers]{natbib}

\usepackage{PRIMEarxiv}

\usepackage[utf8]{inputenc} 
\usepackage[T1]{fontenc}    
\usepackage{hyperref}       
\usepackage{url}            
\usepackage{booktabs}       
\usepackage{amsfonts}       
\usepackage{nicefrac}       
\usepackage{microtype}      
\usepackage{fancyhdr}       
\usepackage{graphicx}       
\usepackage{amsmath} 
\usepackage{amsthm}

\usepackage{tikz}
\usetikzlibrary{shapes.symbols}
\usepackage{fontawesome5}
\usepackage{IEEEtrantools}
\usepackage{bbm}
\usepackage{algorithm}
\usepackage{algpseudocode}
\usepackage{booktabs}
\usepackage{multirow}
\usepackage{xurl}
\usepackage{subfigure}

\newcommand*\diff{\mathop{}\!\mathrm{d}}

\DeclareMathOperator{\op}{op}

\DeclareMathOperator{\sign}{sign}
\DeclareMathOperator{\argmin}{argmin}
\DeclareMathOperator{\argmax}{argmax}

\DeclareMathOperator{\diag}{diag}
\DeclareMathOperator{\eig}{eig}

\newtheorem{definition}{Definition}
\newtheorem{theorem}{Theorem}

\newtheorem{assumption}{Assumption}
\newtheorem{proposition}{Proposition}
\newtheorem{remark}{Remark}
\newtheorem{result}{Result}

\title{Operator-Theoretic Framework for Gradient-Free Federated Learning
}

\author{
Mohit Kumar \\
University of Rostock, Germany\\ Software Competence Center Hagenberg GmbH, Austria \\
  Hagenberg, Austria \\
 \texttt{mohit.kumar@uni-rostock.de} 
\And
  Mathias Brucker, Alexander Valentinitsch \\
  Software Competence Center Hagenberg GmbH \\
  Hagenberg, Austria \\
  \texttt{\{mathias.brucker, alexander.valentinitsch\}@scch.at} \\
   \And
  Adnan Husakovic, Ali Abbas \\
  Primetals Technologies Austria GmbH \\
  Linz, Austria \\
  \texttt{\{adnan.husakovic,ali.abbas\}@primetals.com} \\
   \And
Manuela Geiß, Bernhard A. Moser\\
Software Competence Center Hagenberg GmbH \\
  Hagenberg, Austria \\
  \texttt{\{manuela.geiss, bernhard.moser\}@scch.at} \\
}

\begin{document}
\maketitle

\begin{abstract}
{\bf Background:} Federated learning in practice must address client heterogeneity, strict communication and computation requirements, and data privacy, while optimizing performance.{\bf Objectives:} Develop an operator-theoretic framework for federated learning that simultaneously addresses statistical heterogeneity, performance guarantees, and privacy under practical communication and computation constraints. {\bf Methods:} We first map the $L^2$-optimal solution into a reproducing kernel Hilbert space (RKHS) using a forward operator. Using the available data in that RKHS, we approximate the optimal solution. We then map this solution back to the original $L^2$ function space via the inverse operator. This construction yields a gradient-free learning scheme. We derive explicit finite-sample performance bounds for this scheme using concentration inequalities over operator norms. The framework analytically identifies a data-dependent hypothesis space and provides guarantees on risk, prediction error, robustness, and approximation error. Within this space, we design a communication- and computation-efficient model using kernel machines, leveraging the {\em space folding} property of Kernel Affine Hull Machines (KAHMs). Clients transfer knowledge to the server using a novel scalar metric, {\em space folding measure}, derived from KAHMs. Being a scalar, this measure greatly reduces communication overhead. It also supports a simple differentially private FL protocol in which scalar space folding summaries are computed from noise-perturbed data matrices obtained via a single application of a noise-adding mechanism, thereby avoiding per-round gradient clipping and privacy accounting. Finally, the induced global prediction rule can be implemented using a small number of integer \texttt{minimum} and \texttt{equality-comparison} operations per test point, making it structurally compatible with fully homomorphic encryption (FHE) during inference. {\bf Results:} Across four benchmarks (20Newsgroup, XGLUE-NC, CIFAR-10-LT, CIFAR-100-LT), the resulting gradient-free FL method built on fixed encoder embeddings is competitive with, and in several cases outperforms, strong gradient-based federated fine-tuning, with gains of up to 23.7 percentage points on the considered benchmarks. In differentially private experiments, the proposed kernel-based smoothing mechanism partially offsets the accuracy loss caused by noise in high-privacy regimes. The induced global prediction rule admits an FHE realization based on $Q \times C$ encrypted \texttt{minimum} and $C$ \texttt{equality-comparison} operations per test point (where $Q = \#\text{clients}$ and $C = \#\text{classes}$), and our operation-level benchmarks for these primitives indicate latencies compatible with practical secure inference at the evaluated cryptographic parameter settings. {\bf Conclusions:} The operator-theoretic, gradient-free federated learning framework provides provable performance guarantees with low communication overhead, supports differentially private knowledge transfer via scalar space folding summaries, and yields an FHE-compatible prediction rule for which we report operation-level runtimes, offering a mathematically grounded alternative to gradient-based federated learning under client heterogeneity.
\end{abstract}


\section{Introduction}\label{sec_introduction}
Developing a competitive machine learning model often necessitates a substantial amount of high-quality data for training. However, integrating different data sources to utilize all distributed data is challenging due to data privacy requirements and cumbersome exchange procedures. Data privacy is becoming increasingly important due to regulations such as the GDPR (General Data Protection Regulation) \cite{GDPR} and Artificial Intelligence Act (AI Act) \cite{AI-Act}. Federated learning (FL) offers a popular solution for collaborative learning from distributed, privately owned data under the orchestration of a central server, without requiring participating clients to share raw training data. Despite its appeal, FL faces several practical challenges. First, data across clients are often statistically heterogeneous (non-IID), which can degrade the performance of a one-size-fits-all model. Second, devices may have limited computation and communication budgets, making frequent or large exchanges infeasible. Third, strict privacy and security guarantees (e.g., via differential privacy or encryption) must be enforced without significantly degrading the model's learning performance and computation-efficiency. 
\subsection{Requirements}
We identify the following requirements for the development of an effective FL algorithm. \textbf{R1 (Hypothesis Space for Learning from Heterogeneous Distributed Data):} Provide a mathematical framework that, without imposing parametric form or homogeneity assumptions on the client data distributions (beyond mild regularity conditions such as square-integrability), determines a suitable hypothesis space for task learning in a federated setting. \textbf{R2 (Theoretical Guarantees):} Calculate theoretically the error bounds and evaluate the task learning solution in terms of 1) robustness of the prediction error against the {\em disturbances} arising from uncertainties and data noise, and 2) accuracy of the solution and  asymptotic upper bound on the approximation error. \textbf{R3 (Communication Efficiency):} Ensure that the analytically derived task learning solution can be implemented in a federated setting with communication and computational efficiency. Specifically, the optimization of the global model should not require multiple rounds of communication between the server and clients. \textbf{R4 (Efficient Differentially Private Knowledge Transfer Across Global and Local Models):} Define a novel differentially private metric, that allows for knowledge transfer from clients to server by solving the global model optimization problem without requiring exchange of gradients or model parameters (that may be computationally challenging and not optimal for privacy preservation), while optimizing the utility-privacy tradeoff. \textbf{R5 (Computationally Efficient FHE-Based Secure Federated Learning):} Rather than transmitting high-dimensional gradient or parameter update vectors, which entail substantial computational overhead when transmitted and operated on within encrypted domains, define novel low-dimensional attributes. These attributes must enable the inference of the global model with low communication overhead and latency, thereby supporting a computationally efficient realization of fully homomorphically encrypted inference of the global model. Researchers have approached these issues from multiple angles including communication-efficient protocols, robust aggregation methods, differential privacy techniques, and secure computation. We briefly review these developments next.
\paragraph{Problem Setting and Scope} Throughout this work we focus on a practically motivated FL setting in which pretrained encoders (e.g., deep neural networks trained on large public corpora) are available to the participating clients, but are not jointly updated during federated training. This reflects scenarios where 
\begin{enumerate}
\item encoder updates may be constrained by regulatory or validation requirements, 
\item communication and computation budgets preclude repeated updates of large models, or 
\item the clients intend to reuse a common frozen representation while collaboratively learning a task-specific prediction head.
\end{enumerate}
Within this setting, our goal is not to improve representation learning itself, but to design and analyse a gradient-free, communication-efficient, and privacy-/security-aware federated head on top of such fixed encoders.
\subsection{State of the Art}
To address the above challenges, a variety of federated learning strategies have been explored. In this section, we review key state-of-the-art approaches related to communication efficiency and handling heterogeneous data, differentially private learning, and secure aggregation with homomorphic encryption, among others.
\subsubsection{Communication Efficiency and Addressing Statistical Heterogeneity}\label{sebsec_state_of_the_art_1}
A cloud-edge architecture effectively mitigates communication and computation cost challenges by offloading computationally intensive learning tasks to the edge~\cite{9090366}. To accelerate the model convergence and thus to reduce the number of communication rounds during the learning process, device-to-device communication can be leveraged for mitigating the local over-fitting issue~\cite{guo2022hybrid}. To reduce the communication cost, {\em low-rank Hadamard product parametrization} of the model parameters has been suggested~\cite{hyeon-woo2022fedpara}. Instead of training and transmitting full models, sparse models can be considered for computational and communication efficiency~\cite{Bibikar_Vikalo_Wang_Chen_2022}. To enhance robustness against heterogeneity and improve communication efficiency, clients and the server exchange {\em abstract prototypes}, while local prototypes (rather than gradients) are aggregated ~\cite{Tan_Long_LIU_Zhou_Lu_Jiang_Zhang_2022}. To tailor the local model size and consequently the computation, memory, and data exchange requirements to the available client resources, an importance-based pruning mechanism has been proposed to extract lower-footprint nested submodels~\cite{horv2021fjord}. To facilitate FL on heterogeneous devices, a split-mix strategy~\cite{hong2022efficient} enables the learning of base sub-networks with varying sizes and robustness levels, which can be aggregated on-demand to meet specific inference requirements. The challenges of system heterogeneity and connection uncertainty in federated learning can be tackled by developing models that are readily prunable to arbitrary sizes and thus can be structurally decomposed for learning, inference, and transmission~\cite{pmlr-v162-zhu22e}. The training on the devices can be accelerated by introducing sparsity~\cite{qiu2022zerofl}. To achieve a faster convergence rate in theory and practice, adaptive gradient methods has been integrated into the FL~\cite{10.1609/aaai.v37i9.26235}. The sparse and complementary subsets of the dense model are instead exchanged between server and clients to reduce communication and computational cost \cite{10.1609/aaai.v37i7.25977}. The similarities among clients can be assessed to enable personalized FL while reducing communication overhead, achieved through the optimization of aggregation weights~\cite{10.24963/ijcai.2023/444}.

A study~\cite{10.5555/3618408.3618891} identifies {\em local learning bias} as the pitfall of FL with heterogeneous data, and introduces an algorithm that leverages label-distribution-agnostic pseudo-data to reduce the learning bias on local features and classifiers. An empirical study~\cite{zhang2023large} reports that the large sparse convolution kernels can lead to enhanced robustness against distribution shifts in FL. A robust FL approach is to alleviate the {\em worst-case} effect of distribution shifts on the model performance. This approach has been followed~\cite{NEURIPS2020_f5e53608} for the case of {\em affine distribution shifts} by minimizing the maximum possible loss induced by distribution shifts across clients. Adversarial learning approach~\cite{pmlr-v202-li23j} has been considered for addressing distribution shifts, where the server aims to train a discriminator to distinguish the representations of the clients while the clients aim to generate a common representation distribution. {\em Knowledge distillation} is another approach to address heterogeneity, where e.g. the knowledge about the global view of data distribution is extracted by the server and is distilled to guide local models' learning~\cite{pmlr-v139-zhu21b}. The clustered FL approach~\cite{9174890,10.1609/aaai.v37i8.26197} addresses data distribution heterogeneity by grouping clients with similar distributions. This enables clients within the same cluster to mutually benefit from federated learning while reducing harmful interference from clients with dissimilar distributions. A clustered FL algorithm~\cite{9832954} alternately estimates the cluster identities of the clients and optimizes model parameters for the client clusters. Assuming that each client's data follows a mixture of multiple distributions, a method~\cite{Ruan2021FedSoftSC} facilitates the simultaneous training of cluster models and personalized local models.  

A personalized FL approach to tackle statistical heterogeneity involves clients and the server aiming at learning a global representation together, while each client learns its unique head locally~\cite{pmlr-v139-collins21a}. Along this line, a study~\cite{10.5555/3540261.3540903} suggests learning a common kernel function (parameterized by a neural network) across all clients, while each client employs a personalized Gaussian Process model. A simple personalization mechanism can be provided using local k-nearest neighbors model based on the shared representation provided by the global model~\cite{pmlr-v162-marfoq22a}. The personalized FL problem can be studied under the {\em model agnostic meta-learning framework}~\cite{10.5555/3495724.3496024}, where the goal is to find an initial shared model that clients can easily adapt to their local datasets. Instead of using a {\em meta-model} as the initialization, both the personalized and global models can be pursued in parallel by formulating a bi-level optimization problem using the Moreau envelope as a regularized loss function~\cite{10.5555/3495724.3497520}. To enable pairwise collaborations between clients with similar data, a mechanism has been proposed for personalized federated learning, which exchanges weighted model-aggregation messages between personalized models and personalized cloud models~\cite{Huang_Chu_Zhou_Wang_Liu_Pei_Zhang_2021}. To address the computational limitations of heterogeneous devices in personalized FL, optimized masking vectors (derived by minimizing the bias term in the convergence bound) can be employed to train a sub-network of the learning model for each device, tailored to device's computational capacity~\cite{setayesh2023perfedmask}. A personalized FL method~\cite{10.1609/aaai.v37i9.26330} adaptively aggregates the global model and the previous local model to initialize the local model. 
\subsubsection{Differentially Private Learning}\label{sebsec_state_of_the_art_2}
Differential privacy is the gold-standard approach to provide anonymization guarantees for the data used in FL \cite{10.1145/3494834.3500240}. Differential privacy can be enforced within machine learning pipeline at any of three stages: on the input training data, during model training, or on model predictions \cite{articleDP-fy}. Releasing a differentially private version of the training data would enable any training algorithm to be applied to it, thanks to differential privacy’s post-processing property. However, a large amount of noise would typically be needed in this setting to be added into the data for achieving differential privacy guarantee, leading to the loss of model utility. To address the utility-loss issue, the notion of differential privacy can be relaxed to allow defining privacy in terms of the distinguishability level between inputs by means of a distance function \cite{10.1145/3336191.3371856,10.1145/3459637.3482281}. Generating differentially private synthetic data for the training of models is a promising approach to privacy-preserving machine learning. Differentially private synthetic data can be generated using random projections \cite{pmlr-v124-gondara20a}, Bayesian networks \cite{10.1145/3134428}, Markov random fields \cite{10.14778/3476249.3476272}, GANs \cite{yoon2018pategan}, iterative methods \cite{10.5555/3540261.3540314}, neural tangent kernels \cite{10.1613/jair.1.15985}, and {\em Kernel Affine Hull Machines}~\cite{KAHM}. Perturbing the objective function of the optimization problem with noise is another approach to enforce differential privacy in models with strong convexity \cite{Phan_Wang_Wu_Dou_2016,8835258}. The study in \cite{DBLP:journals/corr/PapernotAEGT16} introduced Private Aggregation of Teacher Ensembles (PATE) method based on knowledge aggregation from ``teacher'' models and transfer to a ``student'' model in differentially private fashion. The authors suggested to train the student model on public unlabeled data using GAN-like approach for semi-supervised learning. This approach achieves differential privacy by injecting noise in the aggregation of teacher models (that have been trained on disjoint splits of a private dataset) without placing any restrictions on teacher models, thus allowing any models to be used in {\em model-agnostic} manner. Further, a rigorous theoretical analysis of the PATE approach is available \cite{NEURIPS2018_aa97d584,JMLR:v22:20-1251}.   

As it is typical that large-scale machine learning models are optimized using gradient-based algorithms, the gradient perturbation-based methods such as DP-SGD (differentially private stochastic gradient descent) are widely used for achieving rigorous differential privacy guarantees \cite{10.1145/2976749.2978318}. DP-SGD operates by running stochastic gradient descent on noisy mini-batch gradients. The DP-SGD method's privacy analysis relies on the concept of {\em privacy amplification by sampling} requiring that each mini-batch is sampled with replacement on each iteration. This requirement may be infeasible to achieve in distributed settings like FL. Thus, the authors in \cite{pmlr-v139-kairouz21b} introduced a solution, referred to as DP-FTRL algorithm, that does not rely on random sampling for privacy amplification, instead leverages the differentially private streaming of of the cumulative sum of gradients. The problem of continual-release of cumulative sums was connected to the matrix mechanism in \cite{denissov2022improved}, yielding improvements in federated learning with user-level differential privacy. The matrix mechanism was extended in \cite{10.5555/3618408.3618644} to the multi-epoch setting, allowing for differentially private gradient-based machine learning with multiple epochs over a dataset.

The integration of FL and differential privacy is potentially an effective approach to privacy-preserving collaborative learning from distributed datasets \cite{naseri2022local}. In practice, the edge device heterogeneity may cause straggler effect that can be mitigated by an {\em asynchronous} approach allowing clients to synchronize with the central server independently and at different times \cite{10316599}. FL models have typically deep learning architecture, where estimating the sensitivity of gradients (which is required for differential privacy), is difficult, and thus gradients are clipped (whenever their norm exceed some threshold) to control the sensitivity \cite{10.1145/2976749.2978318,brendan2018learning}. Gradient clipping has been theoretically proven and empirically observed to accelerate gradient descent optimization in training \cite{Zhang2020Why}. However, the clipping may lead to a bias on the convergence to a stationary point and the clipping bias can be quantified with a disparity measure between the gradient distribution and a geometrically symmetric distribution \cite{10.5555/3495724.3496879}. The choice of clipping threshold is crucial to the performance of differentially private models, necessitating the development of automatic \cite{bu2023automatic} and adaptive \cite{10316599} gradient clipping methods.  
\subsubsection{Secure Learning with Homomorphic Encryption}\label{sebsec_state_of_the_art_3}
Fully Homomorphic Encryption (FHE) allows arbitrary computation in encrypted space, and thus FHE can be applied to FL for protecting the privacy of the data (shared by clients with the server during training) against a server eavesdropper. Traditional single-key FHE (where all clients share one key) poses a security risk in the event of a client colluding with the server, and thus multi-key FHE schemes \cite{10.5555/3081738.3081764} have been considered for FL \cite{electronics13224478}, where the suggested scheme remains secured against collusion attacks involving up to all but one participant. Moreover, their scheme reduces the computational load by reducing the complexity of the NAND gate. In order to reduce computational and communication overhead during HE-based secure model aggregation in federated training, the authors in \cite{jin2023fedmlhe} suggest to selectively encrypt only the most privacy-sensitive parameters. The CKKS scheme \cite{cheon2017ckks} is a leveled homomorphic encryption scheme designed specifically for approximate arithmetic on real or complex numbers, and thus has been considered for privacy-preserving FL \cite{pan2024fedshe}. The SPDZ protocol \cite{damgaard2012spdz} uses somewhat homomorphic encryption and provides low-latency secure multi party computation due to its fast online phase, and thus can be considered for FL \cite{Truhn2024EncryptedFL}. Smart network interface cards can be leveraged as hardware accelerators to offload compute-intensive HE operations of FL \cite{choi2024fednic}. To implement FHE-based secure FL with reduced communication overhead and latency, the authors in \cite{10208145} have experimented with different approaches in which data can be encrypted and transmitted. TFHE (Torus Fully Homomorphic Encryption) is a lattice-based FHE scheme \cite{chillotti2020tfhe} designed for fast gate-by-gate computation on encrypted data with very fast bootstrapping. TFHE leads to a computationally efficient FHE-based FL \cite{10012502,kumar2023secureFLKAHM} by reformulating the FL problem in such a way that 1) exchange of only low-dimensional attributes is required between clients and server, and 2) the inference of the global model in encrypted space is not computationally heavy.                    
\subsection{Research Gap}
In summary, most existing FL algorithms address only a subset of requirements R1–R5. For example, the classic FedAvg algorithm~\cite{pmlr-v54-mcmahan17a} (and similar gradient-based variants) requires many SGD epochs and communication rounds, yet still lacks formal guarantees for heterogeneous data. Approaches using differentially private SGD enhance privacy, but in federated settings they face other issues (e.g. performance drop and reliance on random sampling for privacy amplification). Methods using homomorphic encryption can secure model updates, however, they incur high computational cost and often still depend on exchanging gradients. Recently, kernel-based and prototype-based methods (e.g. distillation of local prototypes, or KAHM-based aggregation) have been explored to reduce communication and improve robustness, but a unified theoretical framework is lacking. In particular, the state of the art (as reviewed in Section~\ref{sebsec_state_of_the_art_1}, \ref{sebsec_state_of_the_art_2}, \ref{sebsec_state_of_the_art_3}) doesn't jointly solve R1–R5. A recent kernel-based FL scheme~\cite{kumar2024geometricallyinspiredkernelmachines} addressed aspects of R2–R3 and enabled privacy \cite{KAHM} and security \cite{kumar2023secureFLKAHM} by exchanging compact task-sufficient information, but it did not address R1 by deriving the underlying hypothesis space for learning from heterogeneous distributed data. Existing FL algorithms either rely on iterative gradient descent, repeated communication, or do not fully protect privacy, and they do not come with end-to-end learning guarantees under heterogeneity. To the best of our knowledge, no existing framework concurrently addresses all requirements R1–R5 within a single, mathematically unified treatment.
\subsection{Contributions}
To address these limitations, we argue that a fundamentally different (gradient-free) approach is needed to satisfy
requirements R1-R5. In this paper, we therefore propose a rigorous operator-theoretic framework that is designed to address R1-R5 within a single, unified treatment. Our framework generalizes the methodology of~\cite{kumar2024geometricallyinspiredkernelmachines} and extends the methodology to address not only R1 but also simultaneously R2, R3, R4, and R5 in a unified manner. Our main contributions are:
\paragraph{Operator-Theoretic Formulation.} We formulate the FL problem in the $L^2$ function space as solving for the minimizer of mean-squared error. We define an invertible forward operator that maps the $L^2-$optimal solution into an RKHS. This RKHS is associated with a generalized kernel whose feature-map serves as an estimator of class posterior probability. We derive a sample-based estimator in the RKHS and prove non-asymptotic upper bound on its risk (Theorem~\ref{theorem_040220251005}) using kernel and operator theory  and concentration inequalities. Notably, under mild regularity conditions on the data-generating distributions (i.e., square-integrability of the class-posterior function, as stated in Section \ref{section_240320251411}), we achieve a risk bound of $\mathcal{O}(1/\sqrt{N})$, where $N$ is the total number of training data samples distributed across clients. We map the RKHS learning solution back to $L^2$ function space using the inverse-operator to obtain a generalized learning solution.
\paragraph{Performance Guarantees.} The generalized learning solution (provided in Section~\ref{section_generalized_learning_solution}) is evaluated for its performance in terms of risk (in Theorem~\ref{theorem_090220251546}), prediction error (in Theorem~\ref{theorem_090220251718}), robustness (in Remark~\ref{remark_100220251448}), and approximation error (in Theorem~\ref{theorem_100220251831}). Under the same regularity condition on the data-generating distributions as in Section \ref{section_240320251411}, it is shown that risk, prediction error, and approximation error bounds are of $\mathcal{O}(1/\sqrt{N})$. {\em The robustness property of the generalized learning solution is established by showing that small disturbances cannot lead to large prediction error}. 
\paragraph{Determination of Hypothesis Space and Theoretical Analysis.} A data-dependent hypothesis space for task learning is determined (in Section~\ref{section_hypothesis_space_determination}) from the generalized learning solution by tuning the kernel to the scale of the data. We then analyze the Rademacher complexity of the hypothesis space (in Theorem~\ref{theorem_110220250938}) and derive upper bound on prediction error (in Theorem~\ref{theorem_120220250928}) and approximation error (in Theorem~\ref{theorem_120220251923}). It is shown (in Remark~\ref{rem_230420251610}) that the achieved error bounds are tighter than the existing bounds \cite{kumar2024geometricallyinspiredkernelmachines}.    
\paragraph{KAHM-Based Space Folding Kernel and Classification:} 
We introduce a novel {\em space folding measure} for KAHMs (Definition~\ref{definition_150220251240}). It quantifies how much a data point must be ``folded'' to fit it into the subspace spanned by the training data (i.e. the KAHM’s data subspace). This space folding measure induces a new kernel whose feature-map aligns with the class membership (Remark~\ref{remark_240320250940}). In fact, the feature value can be interpreted as an estimate of class posterior probability (Proposition~\ref{proposition_160220251115}). 
\subsection{The Proposed Operator-Theoretic Framework}\label{section_240320251428}
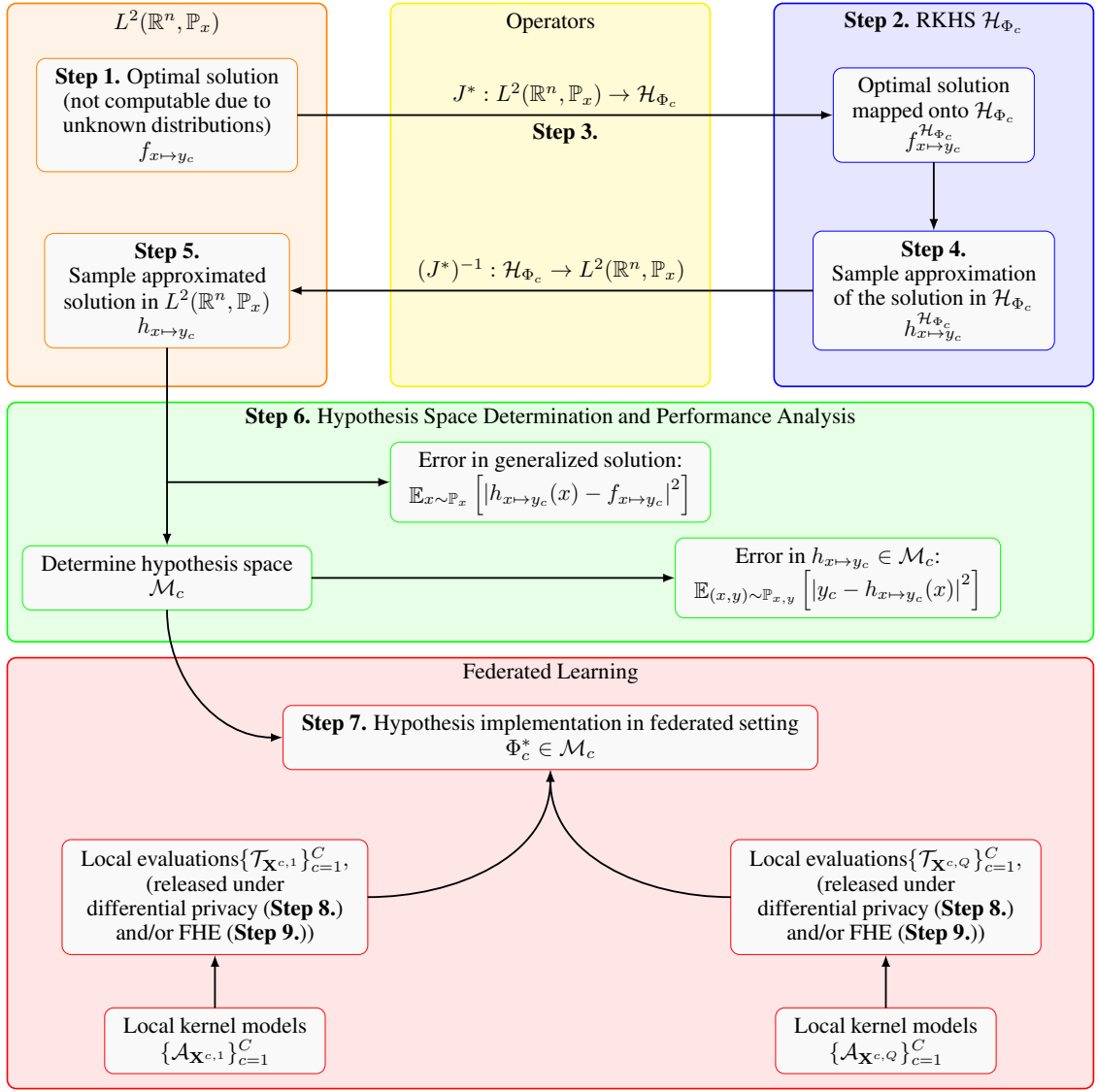
\begin{figure}[!h]  
\centering
\scalebox{0.875}{
\begin{tikzpicture}[scale=1]
\path[fill=orange!10,rounded corners](0,0)--(5,0)--(5,-6)--(0,-6)--cycle;
\draw[orange,line width = 0.25mm,rounded corners](0,0)--(5,0)--(5,-6)--(0,-6)--cycle;
\draw (2.5,-0.3) node[]{ $L^2(\mathbb{R}^n,\mathbb{P}_{x})$};
\draw (2.5,-1.75) node[rounded corners,draw=orange,fill=gray!5](n1){ $\begin{array}{c}\mbox{\textbf{Step 1.} Optimal solution} \\ \mbox{(not computable due to} \\ \mbox{unknown distributions)} \\  f_{x \mapsto y_c} \end{array}$};
\path[fill=yellow!20,rounded corners](0+6,0)--(5+6,0)--(5+6,-6)--(0+6,-6)--cycle;
\draw[yellow,line width = 0.25mm,rounded corners](0+6,0)--(5+6,0)--(5+6,-6)--(0+6,-6)--cycle;
\draw (2.5+6,-0.3) node[]{Operators};
\path[fill=blue!10,rounded corners](0+12,0)--(5+12,0)--(5+12,-6)--(0+12,-6)--cycle;
\draw[blue,line width = 0.25mm,rounded corners](0+12,0)--(5+12,0)--(5+12,-6)--(0+12,-6)--cycle;
\draw (2.5+12,-0.3) node[]{\textbf{Step 2.} RKHS $\mathcal{H}_{\Phi_c}$};
\draw (2.5+12,-1.75) node[rounded corners,draw=blue,fill=gray!5](n2){ $\begin{array}{c} \mbox{Optimal solution}  \\ \mbox{mapped onto $\mathcal{H}_{\Phi_c}$} \\ f_{x \mapsto y_c}^{\mathcal{H}_{\Phi_c}} \end{array}$};
 \draw[-latex,thick,line cap=round] (n1) to [out=0,in=180]  node[above,sloped] {$J^*: L^2(\mathbb{R}^n,\mathbb{P}_{x}) \rightarrow \mathcal{H}_{\Phi_c}$} node[below,sloped]{\textbf{Step 3.}} (n2); 
\draw (2.5+12,-4.5) node[rounded corners,draw=blue,fill=gray!5](n3){ $\begin{array}{c} \mbox{\textbf{Step 4.}} \\  \mbox{Sample approximation}  \\ \mbox{of the solution in $\mathcal{H}_{\Phi_c}$} \\ h_{x \mapsto y_c}^{\mathcal{H}_{\Phi_c}} \end{array}$};
  \draw[-latex,thick,line cap=round] (n2) to [out=-90,in=90]  node[above,sloped] {} (n3); 
\draw (2.5,-4.5) node[rounded corners,draw=orange,fill=gray!5](n4){ $\begin{array}{c} \mbox{\textbf{Step 5.}} \\  \mbox{Sample approximated}  \\ \mbox{solution in $L^2(\mathbb{R}^n,\mathbb{P}_{x})$} \\ h_{x \mapsto y_c} \end{array}$};
 \draw[-latex,thick,line cap=round] (n3) to [out=180,in=0]  node[above,sloped] { $(J^*)^{-1}: \mathcal{H}_{\Phi_c} \rightarrow L^2(\mathbb{R}^n,\mathbb{P}_{x})$}  (n4); 
\path[fill=green!10,rounded corners](0,0-6.25)--(5+12,0-6.25)--(5+12,-10)--(0,-10)--cycle;
\draw[green,line width = 0.25mm,rounded corners](0,0-6.25)--(5+12,0-6.25)--(5+12,-10)--(0,-10)--cycle;
\draw (2.5+6,-6.5) node[]{\textbf{Step 6.} Hypothesis Space Determination and Performance Analysis};
\draw (8.5,-7.5) node[rounded corners,draw=green,fill=gray!5](n5){ $\begin{array}{c}\mbox{Error in generalized solution:} \\  \mathop{\mathbb{E}}_{x \sim \mathbb{P}_x}\left[ \left| h_{x \mapsto y_c}(x) - f_{x \mapsto y_c} \right|^2\right]  \end{array}$};
\draw[-latex,thick,line cap=round] (2.5,-7.5) to [out=0,in=180]  node[above,sloped] {} (n5); 
\draw (2.5,-9) node[rounded corners,draw=green,fill=gray!5](n6){$\begin{array}{c}\mbox{Determine hypothesis space} \\ \mathcal{M}_c  \end{array}$};
\draw[-latex,thick,line cap=round] (n4) to [out=-90,in=90]  node[above,sloped] {} (n6); 
\draw (13,-9) node[rounded corners,draw=green,fill=gray!5](n7){ $\begin{array}{c}\mbox{Error in $h_{x \mapsto y_c} \in \mathcal{M}_c$:} \\  \mathop{\mathbb{E}}_{(x,y) \sim \mathbb{P}_{x,y}} \left [ \left | y_c - h_{x \mapsto y_c}(x) \right |^2  \right]    \end{array}$};
\draw[-latex,thick,line cap=round] (n6) to [out=0,in=180]  node[above,sloped] {} (n7); 
\path[fill=red!10,rounded corners](0,0-10.25)--(5+12,0-10.25)--(5+12,-17)--(0,-17)--cycle;
\draw[red,line width = 0.25mm,rounded corners](0,0-10.25)--(5+12,0-10.25)--(5+12,-17)--(0,-17)--cycle;
\draw (2.5+6,-10.5) node[]{Federated Learning};
\draw (8.5,-11.5) node[rounded corners,draw=red,fill=gray!5](n8){ $\begin{array}{c}\mbox{\textbf{Step 7.} Hypothesis implementation in federated setting} \\  \Phi^*_c \in \mathcal{M}_c   \end{array}$};
\draw[-latex,thick,line cap=round] (n6) to [out=-90,in=180] (n8); 
\draw (3.25,-14) node[rounded corners,draw=red,fill=gray!5](n9){$\begin{array}{c}\mbox{Local evaluations}  \{\mathcal{T}_{ \mathbf{X}^{c,1}}\}_{c= 1}^C, \\ \mbox{(released under} \\ \mbox{differential privacy (\textbf{Step 8.})} \\ \mbox{and/or FHE (\textbf{Step 9.}))}  \end{array}$};
\draw[-latex,thick,line cap=round] (n9) to [out=0,in=-90] (n8); 
\draw (3+10.75,-14) node[rounded corners,draw=red,fill=gray!5](n10){$\begin{array}{c}\mbox{Local evaluations}  \{\mathcal{T}_{ \mathbf{X}^{c,Q}}\}_{c= 1}^C, \\ \mbox{(released under} \\ \mbox{differential privacy (\textbf{Step 8.})} \\ \mbox{and/or FHE (\textbf{Step 9.}))}  \end{array}$};
\draw[-latex,thick,line cap=round] (n10) to [out=180,in=-90] (n8); 
\draw (3.25,-16.25) node[rounded corners,draw=red,fill=gray!5](n11){$\begin{array}{c}\mbox{Local kernel models} \\ \{\mathcal{A}_{ \mathbf{X}^{c,1}}\}_{c= 1}^C  \end{array}$};
\draw[-latex,thick,line cap=round] (n11) to [out=90,in=-90] (n9); 
\draw (3+10.75,-16.25) node[rounded corners,draw=red,fill=gray!5](n12){$\begin{array}{c}\mbox{Local kernel models} \\ \{\mathcal{A}_{ \mathbf{X}^{c,Q}}\}_{c= 1}^C  \end{array}$};
\draw[-latex,thick,line cap=round] (n12) to [out=90,in=-90] (n10); 
\end{tikzpicture}
}
\caption{The operator-theoretic kernel FL framework is developed by 1) considering the optimal learning solution in $L^2(\mathbb{R}^n,\mathbb{P}_{x})$, 2) mapping the optimal solution onto a RKHS (associated to a generalized kernel) using an operator, 3) approximating the optimal solution using available data samples in RKHS, 4) mapping the sample approximated solution onto $L^2(\mathbb{R}^n,\mathbb{P}_{x})$ using the inverse-operator, 5) analyzing the sample approximated solution in $L^2(\mathbb{R}^n,\mathbb{P}_{x})$ and identifying conditions on kernel choice to define hypothesis space, 6) implementing a suitable hypothesis with the minimum computational and communication cost in the federated setting using kernel models.}
\label{fig_framework}
\end{figure}

Below, we outline the key steps of our approach (illustrated in Fig.~\ref{fig_framework}) before delving into the technical details.
\paragraph{Step 1. Analytical Formulation of the Learning Problem in $L^2$ Function Space} The task learning problem is mathematically formulated in the $L^2$ function space and the optimal solution is analytically derived. Since the probability distributions involved in the analytical solution are unknown, the analytically derived solution cannot be practically computed. 
\paragraph{Step 2. A Reproducing Kernel Hilbert Space (RKHS) for the Learning Problem} We consider the RKHS induced by a generalized kernel to approximate the optimal solution. Because the kernel choice is not obvious a priori, we adopt a generalized kernel whose feature map is derived from theoretical analysis aimed at estimating the class-posterior probability. 
\paragraph{Step 3. An Invertible Operator for Mapping Optimal Solution onto RKHS} For an approximation and analysis of the optimal solution in RKHS using powerful kernel theory, an integral kernel operator with its inverse existing is defined from $L^2$ function space to RKHS.   
\paragraph{Step 4. Sample Approximation and Analysis in RKHS:} The mapped (onto RKHS) optimal solution is approximated by means of training data samples distributed across clients to obtain the {\em RKHS learning solution}. Further, kernel theory and concentration inequalities are applied to derive upper bound on the risk for the RKHS learning solution.
\paragraph{Step 5. Generalized Learning Solution and Analysis} The learning solution, obtained in RKHS via sample approximation, is mapped onto the $L^2$ function space through the inverse-operator to obtain the {\em generalized learning solution}. The risk and error bounds for the generalized learning solution are easily obtained, thanks to the derived risk bound for the RKHS learning solution. 
\paragraph{Step 6. Determination of the Hypothesis Space and Analysis} The generalized learning solution is modulated by the kernel choice. We identify conditions on kernel under which the generalized learning solution captures the data's scale. The set of learning solutions with the kernel satisfying the identified conditions defines the hypothesis space. Rademacher complexity of the hypothesis space is calculated to derive upper bounds on prediction error and approximation error for the hypothesis space.    
\paragraph{Step 7. Choosing a Communication Efficient Hypothesis} Having determined the hypothesis space and provided theoretical performance guarantees, the next step is to choose a suitable hypothesis that can be efficiently implemented in the federated setting. It is highlighted that it is possible to define such a hypothesis by means of Kernel Affine Hull Machines (KAHMs)~\cite{KAHM,kumar2024geometricallyinspiredkernelmachines}. Specifically, the {\em space folding property} of the KAHMs (where the space folding property refers to the mapping of an arbitrary point by a KAHM onto the data subspace represented by the KAHM) is leveraged to implement a communication efficient hypothesis. This is done by specifying the feature-map of the kernel as an estimator of class posterior probability from the space folding measures, enabling gradient-free FL protocol where local KAHM-based models are aggregated by means of space folding measures without requiring rounds of communication between server and the clients.  
\paragraph{Step 8. Differentially Private Release of Space Folding Measures} Privacy-preserving knowledge transfer from clients to server is enabled by providing differentially private approximations to the space folding measures. Since estimating the sensitivity of space folding measure is challenging, we consider the differentially private release of data samples using an optimized noise adding mechanism \cite{kumar2019deriving}. The adverse effect of the added noise is mitigated by leveraging the post-processing property of differential privacy for a smoothing of noise added data samples. The study introduces a kernel-based smoothing function, with the degree of smoothing optimized to minimize the deviation of smoothed data points from original data points. 
\paragraph{Step 9. Computationally Efficient Secure Inference of Global Model Using FHE} Since the space folding measure (unlike high-dimensional gradient or parameter update vectors) is scalar-valued, inference for a $C-$class problem with $Q$ participating clients reduces to evaluating $Q \times C$ space folding measures and performing $Q \times C$ \texttt{minimum} operations and $C$ \texttt{equality-comparison} operations per test point. This fixed and low-dimensional operation pattern is well suited to secure implementation under fully homomorphic encryption. In our experiments, we instantiate this using the TFHE scheme and report runtimes for the encrypted minimum and equality-comparison primitives to characterize the computational profile of secure inference.
\subsection{Novelty} 
Kernel methods, empowered by a strong mathematical theory on kernel machine learning, have been considered for FL~\cite{9625795,ghari2022personalized}. However, only a recent study~\cite{kumar2024geometricallyinspiredkernelmachines} has introduced a kernel FL method that departs from gradient descent. That study proposed a KAHM-based federated scheme which considered a particular convex-hull hypothesis and derived Rademacher-complexity-based error bounds, and the study is complemented by separate works on KAHM-based differentially private~\cite{KAHM} and FHE-secured~\cite{kumar2023secureFLKAHM} FL protocols. Taken together, these works partially address R2–R3 and demonstrate that KAHM-based scalar, task-sufficient summaries can support privacy and security. They do not, however, provide a unified operator-theoretic formulation that derives the hypothesis space from first principles, nor do they jointly address requirements R1–R5 within a single framework.  

The present paper goes substantially beyond those earlier works in several ways:
\paragraph{Operator-Theoretic Formulation and Hypothesis-Space Derivation} Instead of postulating a particular convex hypothesis set as in~\cite{kumar2024geometricallyinspiredkernelmachines}, we formulate the learning problem in the $L^2$ function space, derive the $L^2-$optimal solution, map it into an RKHS via an invertible operator, and then map a sample-based RKHS approximation back to $L^2$. This forward-inverse operator construction yields a generalized learning solution and, by identifying conditions on the generalized kernel, induces a data-dependent hypothesis space tailored to heterogeneous client distributions. 
\paragraph{Integrated Operator- and Complexity-Based Analysis with Tighter Bounds} We derive non-asymptotic risk, prediction-error, robustness, and approximation-error bounds for the generalized solution via operator-theoretic arguments, and then combine these with new Rademacher-complexity bounds. The resulting bounds are strictly tighter than those in \cite{kumar2024geometricallyinspiredkernelmachines}.
\paragraph{Generalized space folding Kernel and Probabilistic Interpretation} We extend the KAHM-induced distance used for aggregation in \cite{kumar2024geometricallyinspiredkernelmachines,KAHM,kumar2023secureFLKAHM} to a new space folding measure and associated generalized kernel whose feature map admits an interpretation as an estimator of the class posterior probability. This construction is derived from the operator-theoretic framework rather than chosen ad hoc.
\paragraph{Unified Treatment of Communication Efficiency, DP, and FHE} Within the same operator-theoretic framework, we show that scalar space folding summaries are sufficient statistics for the global prediction rule, that they admit an optimal univariate noise distribution for $(\epsilon,\delta)-$differential privacy together with a kernel-based post-processing smoother, and that the resulting decision rule has a fixed gate-level structure compatible with FHE-based secure inference. Earlier works \cite{kumar2024geometricallyinspiredkernelmachines,KAHM,kumar2023secureFLKAHM} studied these aspects in isolation, without connecting them to a formally derived hypothesis space.
\paragraph{Expanded Empirical Study} We provide a new empirical evaluation on four benchmarks (20Newsgroup, XGLUE-NC, CIFAR-10-LT, CIFAR-100-LT) under heterogeneous and long-tailed data partitions, including ablations on space folding, batch size, and embedding combinations. These experiments go beyond those reported in \cite{kumar2024geometricallyinspiredkernelmachines,KAHM,kumar2023secureFLKAHM}.
\subsection{Organization}
The remainder of the paper is organized as follows: Section~\ref{sec_background} introduces the necessary notations and formal problem setup. Section~\ref{sec_main} presents the development of our operator-theoretic framework, including theoretical analysis of its performance. Section~\ref{sec_experiments} reports experimental results on benchmarks, and ~\ref{sec_conclusion} concludes with a discussion of the findings and future work.

\section{Mathematical Prerequisites}\label{sec_background}
This section introduces the used notations, presents the considered distributed data setting, and reviews the necessary definitions.
\subsection{Notations}
We use the boldface font to denote the matrices. The following notations are introduced:
\begin{itemize}
\item Let $n,N,c,C,q,Q \in \mathbb{Z}_{+}$ be the positive integers. 
\item For a scalar $a\in \mathbb{R}$, $|a|$ denotes its absolute value. For a set $A$, $|A|$ denotes its cardinality. For a real matrix $\mathbf{X}$, $\mathbf{X}^T$ is the transpose of $\mathbf{X}$. 
\item For a vector $y \in \mathbb{R}^C$, $\| y\|$ denotes the Euclidean norm and $y_j$ (and also $(y)_j$) denotes the $j^{th}$ element. For a matrix $\mathbf{X}\in \mathbb{R}^{N \times n}$, $\|\mathbf{X}\|_2$ denotes the spectral norm, $\| \mathbf{X} \|_F$ denotes the Frobenius norm, $(\mathbf{X})_{i,:}$ denotes the $i^{th}$ row, $(\mathbf{X})_{:,j}$ denotes the $j^{th}$ column, and $(\mathbf{X})_{i,j}$ denotes the $(i,j)^{th}$ element.  
\item For a set $\{x^1,\cdots,x^N \} \subset  \mathbb{R}^n$, its affine hull is denoted as $\mathrm{aff}\left(\{x^1,\cdots,x^N \}\right)$. 
\item The square brackets are used to represent the construction of a matrix from columns e.g. $\left[\begin{IEEEeqnarraybox*}[][c]{,c/c/c,} x^1 & \cdots & x^N \end{IEEEeqnarraybox*} \right]$ is a matrix with vectors $x^1,\cdots,x^N$ as the columns.
\end{itemize}
\subsection{Definitions}
\begin{itemize}
\item Let $(\Omega_{x},\mathcal{F}_{x},\mu_{x})$ be a {\em probability space} and $x:\Omega_{x} \rightarrow \mathbb{R}^n$ be a random vector on $\Omega_{x}$. Let $\mathcal{B}(\mathbb{R}^n)$ be the {\em Borel $\sigma-$algebra} on $\mathbb{R}^n$. Let $\mathbb{P}_{x}: \mathcal{B}(\mathbb{R}^n) \rightarrow [0,1]$ be the distribution of $x$ given as   
      \begin{IEEEeqnarray}{rCl}
\label{eq_290820251436}\mathbb{P}_{x} & := &  \mu_{x} \circ x^{-1}.
      \end{IEEEeqnarray}
  \item Let $(\Omega_{x,y},\mathcal{F}_{x,y},\mu_{x,y})$ be a probability space and $(x,y): \Omega_{x,y} \rightarrow \mathbb{R}^n \times \{0,1\}^C$ be a random vector on $\Omega_{x,y}$. Let $\mathcal{B}(\mathbb{R}^n \times \{0,1\}^C)$ denote the Borel $\sigma-$algebra on $\mathbb{R}^n \times \{0,1\}^C$. Let $\mathbb{P}_{x,y}: \mathcal{B}(\mathbb{R}^n \times \{0,1\}^C) \rightarrow [0,1]$ be the distribution of $(x,y)$ given as         
      \begin{IEEEeqnarray}{rCl}
\mathbb{P}_{x,y} & := &  \mu_{x,y} \circ (x,y)^{-1}.
      \end{IEEEeqnarray}
\item Let$(\Omega_{x,y,q},\mathcal{F}_{x,y,q},\mu_{x,y,q})$ be a probability space and $(x,y,q): \Omega_{x,y,q} \rightarrow \mathbb{R}^n \times \{0,1\}^C \times \{1,2,\cdots,Q \}$ be a random vector on $\Omega_{x,y,q}$. Let $\mathcal{B}(\mathbb{R}^n \times \{0,1\}^C \times \{1,2,\cdots,Q \})$ denote the Borel $\sigma-$algebra on $\mathbb{R}^n \times \{0,1\}^C \times \{1,2,\cdots,Q \}$. Let $\mathbb{P}_{x,y,q}: \mathcal{B}(\mathbb{R}^n \times \{0,1\}^C \times \{1,2,\cdots,Q \}) \rightarrow \mathbb{R}$ be the distribution of $(x,y,q)$ given as         
      \begin{IEEEeqnarray}{rCl}
\mathbb{P}_{x,y,q} & := &  \mu_{x,y,q} \circ (x,y,q)^{-1}.
      \end{IEEEeqnarray}    
\item Let $L^2(\mathbb{R}^n,\mathbb{P}_{x})$ be the space of all complex-valued measurable functions on $\mathbb{R}^n$ that satisfy
   \begin{IEEEeqnarray}{rCl}
\int_{\mathbb{R}^n} |f(x)|^2 \diff{\mathbb{P}_{x}}(x) & < & \infty.
        \end{IEEEeqnarray}
The norm of a $f \in L^2(\mathbb{R}^n,\mathbb{P}_{x})$ is given as
   \begin{IEEEeqnarray}{rCl}
\| f \|_{L^2(\mathbb{R}^n,\mathbb{P}_{x})} & : = & \left( \int_{\mathbb{R}^n} |f(x)|^2 \diff{\mathbb{P}_{x}}(x) \right)^{1/2}.
     \end{IEEEeqnarray}
\end{itemize}
\subsection{Statistically Heterogeneously Distributed Data Setting}
Let $\mathcal{D}$ be a set consisting of $N$ number of samples drawn IID according to the distribution $\mathbb{P}_{x,y}$:
      \begin{IEEEeqnarray}{rCCCl}
\label{eq_080620242021}   \mathcal{D} & := &  \{(x^i,y^i) \in \mathbb{R}^n\times \{0,1\}^C \; \mid \;i \in \{1,2,\cdots,N\} \}  & \sim & (\mathbb{P}_{x,y})^N.
      \end{IEEEeqnarray}
Let $\mathcal{I}^{c}$ be the set of indices of those samples in the sequence $\left(\left(x^i,y^i\right) \in \mathcal{D}\right)_{i=1}^N$ which are $c^{th}$ class labelled, i.e.,
    \begin{IEEEeqnarray}{rCl}
   \mathcal{I}^{c} & : = & \left \{ i \in \{1,2,\cdots,N\} \; \mid \; y^i_c = 1  \right \}. 
      \end{IEEEeqnarray}
Let $N_c$ be the number of $c^{th}$ class labelled samples, i.e.,
     \begin{IEEEeqnarray}{rCl}
N_c &  = & |\mathcal{I}^{c}|.
      \end{IEEEeqnarray} 
Let $\mathrm{I}^{c} = (\mathrm{I}_1^{c},\cdots,\mathrm{I}_{N_c}^{c})$ be the sequence of elements of $\mathcal{I}^{c}$ in ascending order, i.e., 
\begin{IEEEeqnarray}{rCl}
\mathrm{I}_1^{c} & = & \min(\mathcal{I}^{c}) \\
\mathrm{I}_i^{c} & = & \min(\mathcal{I}^{c} \setminus \{\mathrm{I}_1^{c}, \cdots, \mathrm{I}_{i-1}^{c}\} ),\; \forall i \in \{2,\cdots,N_c\}.
\end{IEEEeqnarray}
Let $\mathbf{X}^c  \in \mathbb{R}^{N_c \times n}$ be the matrix storing $c^{th}$ class labelled samples as its rows, i.e., 
     \begin{IEEEeqnarray}{rCl}
\mathbf{X}^c &  = & \left[x^{\mathrm{I}_1^{c}} \cdots x^{\mathrm{I}_{N_c}^{c}}\right]^T.
      \end{IEEEeqnarray} 
We consider the distributed data setting where total data samples are distributed among $Q$ ($Q > 1$) different clients. Let $q^i \in \{1,2,\cdots,Q \}$ be the client characterizing variable associated to the $i^{th}$ sample pair $(x^i,y^i)$ indicating which of the $Q$ clients owns the $i^{th}$ sample pair. Let $\mathcal{I}^{c,q}$ be the set of indices of those samples in the sequence $\left(\left(x^i,y^i\right) \in \mathcal{D} \right)_{i=1}^N$ which are $c^{th}$ class labelled and owned by client $q$, i.e.,
    \begin{IEEEeqnarray}{rCl}
 \label{eq_310120251532}    \mathcal{I}^{c,q} & : = & \left \{ i \in \{1,2,\cdots,N\} \; \mid \; (y^i)_c = 1,\; q^i = q  \right \}.
      \end{IEEEeqnarray} 
Let $(\mathrm{I}_1^{c,q},\cdots,\mathrm{I}_{|\mathcal{I}^{c,q}|}^{c,q})$ be the sequence of elements of $\mathcal{I}^{c,q}$ in ascending order, i.e., 
\begin{IEEEeqnarray}{rCl}
\mathrm{I}_1^{c,q} & = & \min(\mathcal{I}^{c,q}), \\
\mathrm{I}_i^{c,q} & = & \min(\mathcal{I}^{c,q} \setminus \{\mathrm{I}_1^{c,q}, \cdots, \mathrm{I}_{i-1}^{c,q}\} ),
\end{IEEEeqnarray}
for $i \in \{2,\cdots, |\mathcal{I}^{c,q}|\}$. Let $\mathbf{X}^{c,q} \in \mathbb{R}^{ |\mathcal{I}^{c,q}| \times n}$ be the matrix storing the $c^{th}$ class labelled and $q^{th}$ client owned samples, i.e., 
\begin{IEEEeqnarray}{rCl}
\label{eq_310120251534} \mathbf{X}^{c,q} &=& \left[\begin{IEEEeqnarraybox*}[][c]{,c/c/c,} x^{\mathrm{I}_1^{c,q}} & \cdots & x^{\mathrm{I}_{|\mathcal{I}^{c,q}|}^{c,q}} \end{IEEEeqnarraybox*} \right]^T.
\end{IEEEeqnarray}
Since the $c^{th}$ class labelled samples are distributed among $Q$ clients, we have
\begin{IEEEeqnarray}{rCl}
N_c & = & |\mathcal{I}^{c,1}| +|\mathcal{I}^{c,2}|+ \cdots + |\mathcal{I}^{c,Q}|.
\end{IEEEeqnarray}
\begin{remark}[Data Heterogeneity across Clients]\label{rem_111320251253}
We assume that data samples are statistically heterogeneously distributed, i.e., for arbitrary clients $q^i$ and $q^j$ with $i \neq j$, we assume that 
 \begin{IEEEeqnarray}{rCl}
 \mathbb{P}_{x,y| q}(\cdot,\cdot | q = q^i) & \neq &  \mathbb{P}_{x,y| q}(\cdot,\cdot | q = q^j) ,\\
 \mathbb{P}_{y| x, q}(\cdot | x, q = q^i) & \neq &  \mathbb{P}_{y| x, q}(\cdot | x, q = q^j).
 \end{IEEEeqnarray} 
 \end{remark}
 \subsection{Kernel Affine Hull Machine (KAHM)}  
The KAHMs, originally defined in \cite{KAHM}, have been considered for automated machine learning in \cite{kumar2024geometricallyinspiredkernelmachines}. Given a finite number of samples: $\mathbf{X} = \left[\begin{IEEEeqnarraybox*}[][c]{,c/c/c,} x^1 & \cdots & x^N \end{IEEEeqnarraybox*} \right]^T$ with $x^1,\cdots,x^N \in \mathbb{R}^n$, a KAHM $\mathcal{A}_{\mathbf{X}}: \mathbb{R}^n \rightarrow \mathrm{aff}(\{x^1,\cdots,x^N \})$ is defined as
  \begin{IEEEeqnarray}{rCl}
\label{eq_220420251843}\mathcal{A}_{\mathbf{X}}(x) & := & \frac{h_{\mathbf{X}}^1(\mathbf{P}_{\mathbf{X}}x)}{\sum_{i=1}^Nh_{\mathbf{X}}^i(\mathbf{P}_{\mathbf{X}}x)}x^1 + \cdots + \frac{h_{\mathbf{X}}^N(\mathbf{P}_{\mathbf{X}}x)}{\sum_{i=1}^Nh_{\mathbf{X}}^i(\mathbf{P}_{\mathbf{X}}x)}x^N.
   \end{IEEEeqnarray} 
Appendix~A presents a comprehensive description of the variables and functions associated with (\ref{eq_220420251843}).

 \section{Operator-Theoretic Kernel Federated Learning Framework}\label{sec_main}
This section provides an operator-theoretic framework for kernel FL. As stated previously in Section~\ref{section_240320251428}, the framework development approach consists of 7 steps. Each step is described separately in a subsection.  
\subsection{Step 1: Learning Problem in $L^2(\mathbb{R}^n,\mathbb{P}_{x})$}\label{section_240320251411}
We consider the learning problem in $L^2(\mathbb{R}^n,\mathbb{P}_{x})$. Our goal is to learn a function $f_{x \mapsto y_c}: \mathbb{R}^n \rightarrow \mathbb{R}$ that minimizes the mean squared error:
 \begin{IEEEeqnarray}{rCl}
f_{x \mapsto y_c} & : = & \mathop{\argmin}_{g \in L^2(\mathbb{R}^n,\mathbb{P}_{x})} \mathop{\mathbb{E}}_{(x,y) \sim \mathbb{P}_{x,y} } \left[ |y_c - g(x)|^2 \right]\\
& = & \mathop{\argmin}_{g \in L^2(\mathbb{R}^n,\mathbb{P}_{x})} \left( \int_{\mathbb{R}^n\times \{0,1\}^C} |y_c - g(x)|^2 \diff{\mathbb{P}_{x,y}}(x,y) \right).
   \end{IEEEeqnarray}  
It is well-known and also shown in Appendix B that the conditional expectation, also known as regression function, minimizes the mean squared error. That is,
 \begin{IEEEeqnarray}{rCl}
\label{eq_040920240934}f_{x \mapsto y_c}(x) &  = & \mathop{\mathbb{E}}_{y \sim \mathbb{P}_{y | x}} \left[ y_c | x  \right],
   \end{IEEEeqnarray}
where we have made the following regularity assumption:
\begin{assumption}[Square-Integrability of the Regression Function]\label{assumption_regularity}
 \begin{IEEEeqnarray}{rCl}
 \mathop{\mathbb{E}}_{y \sim \mathbb{P}_{y | x}} \left[ y_c | x  \right] & \in & L^2(\mathbb{R}^n,\mathbb{P}_{x}).
    \end{IEEEeqnarray}
\end{assumption}
Due to $y_c \in \{0,1\}$, we have 
 \begin{IEEEeqnarray}{rCl}
\label{eq_100920241321}f_{x \mapsto y_c}(x) &  \in & [0,1].
   \end{IEEEeqnarray}
For an analysis, the {\em disturbance function}, $\xi_c:\mathbb{R}^n\times \{0,1\}^C \rightarrow \mathbb{R}$, is defined as
 \begin{IEEEeqnarray}{rCl}
\xi_c(x,y) & := & y_c - f_{x \mapsto y_c}(x) \\
\label{eq_101220241056}& = & y_c - \mathop{\mathbb{E}}_{y \sim \mathbb{P}_{y | x}} \left[ y_c | x  \right].
    \end{IEEEeqnarray}
It is obvious that
 \begin{IEEEeqnarray}{rCl}
\label{eq_240920241406}\mathop{\mathbb{E}}_{y \sim \mathbb{P}_{y | x}} \left[ \xi_c(x,y) \right] & = & 0,
    \end{IEEEeqnarray}
    and
     \begin{IEEEeqnarray}{rCl}
\label{eq_240920241255} \xi_c(x,y) & \in & [-1,1].
    \end{IEEEeqnarray}
\subsection{Step 2: A RKHS Associated to a Generalized Kernel Function} 
We consider a generalized kernel function such that for each class $c \in \{1,2,\cdots,C \}$,
 \begin{IEEEeqnarray}{rCl}
\label{eq_260920241342} \mathcal{K}_{\Phi_c}(x,x')& : = & \Phi_c(x) \Phi_c(x'),
  \end{IEEEeqnarray} 
where $\Phi_c:\mathbb{R}^n \rightarrow [0,1]$ is the feature-map (which will be determined based on theoretical analysis to estimate class posterior probability). 
\begin{remark}[Rational for the Restrictive Feature-Map]\label{rem_031120250851}
The rational for the restrictive nature of the feature-map, $\Phi_c:\mathbb{R}^n \rightarrow [0,1]$, is the intent of setting it as an estimator of the class posterior probability, i.e., $\Phi_c(x)  \approx  \mathbb{P}_{y | x}(y_c = 1 | x)$. The bound on the error in approximating class posterior probability through the feature-map will be derived (in Proposition~\ref{proposition_160220251115}).  
\end{remark}
Since for any $x \in \mathbb{R}^n$, $\Phi_c(x) \in [0,1]$, thus we have 
    \begin{IEEEeqnarray}{rCl}
\label{eq_270920241923}\| \Phi_c \|_{L^2(\mathbb{R}^n,\mathbb{P}_{x})}^2 & \geq & 0 \\
\| \Phi_c \|_{L^2(\mathbb{R}^n,\mathbb{P}_{x})}^2 & \leq & 1.
     \end{IEEEeqnarray} 
It is shown in Appendix C that $\mathcal{K}_{\Phi_c}$ is a positive semi-definite kernel. Now the RKHS associated to $\mathcal{K}_{\Phi_c}$ is given as
\begin{IEEEeqnarray}{rCl}
\mathcal{H}_{\Phi_c} & := & \left\{f = \sum_{i=1}^{\infty} \alpha_i \mathcal{K}_{\Phi_c}(\cdot,x^i) \: \mid \: \alpha_i \in \mathbb{R},\; x^i \in \mathbb{R}^n,\; \| f \|_{\mathcal{H}_{\Phi_c}}^2 := \sum_{i,j=1}^{\infty} \alpha_i \alpha_j \mathcal{K}_{\Phi_c}(x^i,x^j) < \infty   \right \} \IEEEeqnarraynumspace
\end{IEEEeqnarray} 
with inner product for any $f = \sum_{i=1}^{N}a_i \mathcal{K}_{\Phi_c}(\cdot,s^i) \in \mathcal{H}_{\Phi_c}$ and $g = \sum_{j=1}^{M}b_j \mathcal{K}_{\Phi_c}(\cdot,t^j) \in \mathcal{H}_{\Phi_c}$ defined as
 \begin{IEEEeqnarray}{rCl}
 \langle f,g\rangle_{\mathcal{H}_{\Phi_c}} & : = & \sum_{i=1}^N \sum_{j=1}^M a_i b_j \mathcal{K}_{\Phi_c}(s^i,t^j). 
 \end{IEEEeqnarray}  
\subsection{Step 3: Operators between $L^2(\mathbb{R}^n,\mathbb{P}_{x})$ and RKHS}
To enable kernel-based approximation, we map the $L^2-$optimal function into an RKHS using a bounded linear operator. We introduce an operator from $L^2(\mathbb{R}^n,\mathbb{P}_{x})$ to $\mathcal{H}_{\Phi_c}$ such that it is invertible. For defining such an operator, we first consider the inclusion operator $J:\mathcal{H}_{\Phi_c} \hookrightarrow L^2(\mathbb{R}^n,\mathbb{P}_{x})$, its adjoint operator $J^*: L^2(\mathbb{R}^n,\mathbb{P}_{x}) \rightarrow \mathcal{H}_{\Phi_c}$, and the inverse of the adjoint operator $(J^*)^{-1}: \mathcal{H}_{\Phi_c} \rightarrow L^2(\mathbb{R}^n,\mathbb{P}_{x})$. It is shown in Appendix D that $J$ is well defined. Consider for any $f \in \mathcal{H}_{\Phi_c}$,
 \begin{IEEEeqnarray}{rCl}
\left \langle Jf, g \right \rangle_{L^2(\mathbb{R}^n,\mathbb{P}_{x})} & = & \mathop{\mathbb{E}}_{x \sim \mathbb{P}_{x}} \left[ f(x) g(x) \right] \\
& = & \mathop{\mathbb{E}}_{x \sim \mathbb{P}_{x}} \left[ \left \langle f, \mathcal{K}_{\Phi_c}(x,\cdot) \right \rangle_{\mathcal{H}_{\Phi_c}} g(x) \right] \\
& = & \left \langle f, \mathop{\mathbb{E}}_{x \sim \mathbb{P}_{x}} \left[ \mathcal{K}_{\Phi_c}(x,\cdot)g(x) \right] \right \rangle_{\mathcal{H}_{\Phi_c}}.
   \end{IEEEeqnarray}      
It follows that the adjoint of $J$, $J^*: L^2(\mathbb{R}^n,\mathbb{P}_{x}) \rightarrow \mathcal{H}_{\Phi_c}$, is given as
 \begin{IEEEeqnarray}{rCl}
(J^*g)(x) & : = & \mathop{\mathbb{E}}_{x' \sim \mathbb{P}_{x}} \left[ \mathcal{K}_{\Phi_c}(x',x)g(x') \right] \\
& = & \int_{\mathbb{R}^n} \mathcal{K}_{\Phi_c}(x',x)g(x')\diff{\mathbb{P}_{x}}(x').
  \end{IEEEeqnarray}
Consider for any $f = \sum_{i=1}^{\infty} \alpha_i \mathcal{K}_{\Phi_c}(\cdot,x^i) \in \mathcal{H}_{\Phi_c}$, 
 \begin{IEEEeqnarray}{rCl}
\left(J^*\frac{f}{\| \Phi_c \|_{L^2(\mathbb{R}^n,\mathbb{P}_{x})}^2}\right)(x) &  = & \frac{1}{\| \Phi_c \|_{L^2(\mathbb{R}^n,\mathbb{P}_{x})}^2} \mathop{\mathbb{E}}_{x' \sim \mathbb{P}_{x}} \left[ \mathcal{K}_{\Phi_c}(x',x)f(x') \right] \\
& = & \frac{\Phi_c(x)}{\| \Phi_c \|_{L^2(\mathbb{R}^n,\mathbb{P}_{x})}^2} \mathop{\mathbb{E}}_{x' \sim \mathbb{P}_{x}} \left[ \Phi_c(x')f(x') \right] \\
& = & \frac{\Phi_c(x)}{\| \Phi_c \|_{L^2(\mathbb{R}^n,\mathbb{P}_{x})}^2} \mathop{\mathbb{E}}_{x' \sim \mathbb{P}_{x}} \left[ \Phi_c(x')\sum_{i=1}^{\infty}\alpha_i \mathcal{K}_{\Phi_c}(x',x^i) \right] \\
& = & \frac{\Phi_c(x)}{\| \Phi_c \|_{L^2(\mathbb{R}^n,\mathbb{P}_{x})}^2} \mathop{\mathbb{E}}_{x' \sim \mathbb{P}_{x}} \left[ |\Phi_c(x')|^2 \right] \sum_{i=1}^{\infty}\alpha_i \Phi_c(x^i) \\
& = & \Phi_c(x) \sum_{i=1}^{\infty}\alpha_i \Phi_c(x^i) \\
& = & f(x).
  \end{IEEEeqnarray}
It follows that the inverse of $J^*$, $(J^*)^{-1}: \mathcal{H}_{\Phi_c} \rightarrow L^2(\mathbb{R}^n,\mathbb{P}_{x})$, is given as
 \begin{IEEEeqnarray}{rCl}
\label{eq_270920241900}\left( (J^*)^{-1} f\right)(x) & := & \frac{f(x)}{\| \Phi_c \|_{L^2(\mathbb{R}^n,\mathbb{P}_{x})}^2},
 \end{IEEEeqnarray}
where $\Phi_c:\mathbb{R}^n\rightarrow [0,1]$ characterizes the kernel $\mathcal{K}_{\Phi_c}$, as stated in (\ref{eq_260920241342}). It is shown in Appendix E that $(J^*)^{-1}$ is well defined on the range of $J^*$.
\subsubsection{A Few Propositions}
It is shown in Appendix F that $J^*J$ is given as 
 \begin{IEEEeqnarray}{rCl}
\label{eq_080220251753}J^*J & = & \mathop{\mathbb{E}}_{x' \sim \mathbb{P}_{x}} \left[ (\mathcal{K}_{\Phi_c}(x',\cdot)  \otimes \mathcal{K}_{\Phi_c}(x',\cdot) )\right],
    \end{IEEEeqnarray}
where for any $f,f' \in \mathcal{H}_{\Phi_c}$, $f \otimes f': \mathcal{H}_{\Phi_c} \rightarrow \mathcal{H}_{\Phi_c}$ is defined as
 \begin{IEEEeqnarray}{rCl}
(f \otimes f')(g) & := & \langle g,f' \rangle_{\mathcal{H}_{\Phi_c}} f. 
  \end{IEEEeqnarray}
It is shown in Appendix G that the norm of the operator $J^*J$ is upper bounded as
 \begin{IEEEeqnarray}{rCl}
\label{eq_260720241423}\|J^*J \|_{\op}& \leq  & 1. 
   \end{IEEEeqnarray} 
It is clear that $J^*J$ is self-adjoint. $J^*J$ is a positive operator, since for all $f \in \mathcal{H}_{\Phi_c}$,
 \begin{IEEEeqnarray}{rCl}
\left \langle f, (J^*J) f \right \rangle_{\mathcal{H}_{\Phi_c}} & = & \left \langle J f, J f \right \rangle_{L^2(\mathbb{R}^n,\mathbb{P}_{x})} \\
& = & \| Jf \|^2_{L^2(\mathbb{R}^n,\mathbb{P}_{x})} \\
& \geq & 0.
    \end{IEEEeqnarray}
Since $J^*J$ is a positive self-adjoint operator, there is a unique positive self-adjoint square root of $J^*J$ and is denoted by $(J^*J)^{1/2}$. It follows from (\ref{eq_260720241423}) that
  \begin{IEEEeqnarray}{rCl}
\label{eq_260720241424}\|(J^*J)^{1/2}\|_{\op}& \leq  & 1. 
   \end{IEEEeqnarray} 
\subsubsection{Sample Approximation of Operators}
For a given sample sequence $(x^i \in \mathbb{R}^n)_{i=1}^N$, let $\widehat{S}_{(x^i)_{i=1}^N}:\mathcal{H}_{\Phi_c} \rightarrow (\mathbb{R}^N, \langle \cdot, \cdot \rangle_{\mathbb{R}^N} )$ be the {\em sample evaluation} operator defined as
 \begin{IEEEeqnarray}{rCl}
\widehat{S}_{(x^i)_{i=1}^N}f & := & (f(x^1),\cdots,f(x^N)),
    \end{IEEEeqnarray}
with inner product for any $u,v\in \mathbb{R}^N$ defined as
 \begin{IEEEeqnarray}{rCl}
\langle u, v \rangle_{\mathbb{R}^N} & := & \frac{1}{N} \sum_{i=1}^Nu_i v_i.
    \end{IEEEeqnarray}
$\widehat{S}_{(x^i)_{i=1}^N}$ is viewed as the sample approximation of $J$. For any $u \in \mathbb{R}^N$, we have
 \begin{IEEEeqnarray}{rCl}
\langle \widehat{S}_{(x^i)_{i=1}^N}f, u \rangle_{\mathbb{R}^N} & = & \frac{1}{N} \sum_{i=1}^N f(x^i).u_i\\
& = & \frac{1}{N}\sum_{i=1}^N \langle f, \mathcal{K}_{\Phi_c}(x^i,\cdot) \rangle_{\mathcal{H}_{\Phi_c}} u_i \\
& = & \left \langle f, \frac{1}{N} \sum_{i=1}^N u_i \mathcal{K}_{\Phi_c}(x^i,\cdot)  \right \rangle_{\mathcal{H}_{\Phi_c}}.
    \end{IEEEeqnarray}
It follows that adjoint of $\widehat{S}_{(x^i)_{i=1}^N}$, $\widehat{S}^*_{(x^i)_{i=1}^N} : (\mathbb{R}^N, \langle \cdot, \cdot \rangle_{\mathbb{R}^N} ) \rightarrow \mathcal{H}_{\Phi_c}$, is given as
 \begin{IEEEeqnarray}{rCl}
\widehat{S}^*_{(x^i)_{i=1}^N} (u_1, \cdots, u_N) & := & \frac{1}{N} \sum_{i=1}^N u_i \mathcal{K}_{\Phi_c}(x^i,\cdot). 
    \end{IEEEeqnarray} 
$\widehat{S}^*_{(x^i)_{i=1}^N}$ is viewed as the sample approximation of $J^*$.
\subsection{Step 4: RKHS Learning Solution}
The optimal solution (\ref{eq_040920240934}), $f_{x \mapsto y_c} \in L^2(\mathbb{R}^n,\mathbb{P}_{x})$, is mapped onto $\mathcal{H}_{\Phi_c}$ through the operator $J^*$, i.e.,  
 \begin{IEEEeqnarray}{rCl}
\label{eq_040220250916} f_{x \mapsto y_c}^{\mathcal{H}_{\Phi_c}} &  := &  J^* f_{x \mapsto y_c}.
\end{IEEEeqnarray}
To approximate $f_{x \mapsto y_c}^{\mathcal{H}_{\Phi_c}}$ in RKHS, a natural approach is of approximating $J^*$ in (\ref{eq_040220250916}) using available data samples $\mathcal{D}$ (\ref{eq_080620242021}). This leads to
\begin{IEEEeqnarray}{rCl}
\label{eq_100920240903} h_{x \mapsto y_c}^{\mathcal{H}_{\Phi_c}}& := & \widehat{S}^*_{(x^i)_{i=1}^N} (y_c^{1},  \cdots,  y_c^{N} ),
\end{IEEEeqnarray}
where $ h_{x \mapsto y_c}^{\mathcal{H}_{\Phi_c}} \in \mathcal{H}_{\Phi_c}$ is viewed as the sample approximation of $f_{x \mapsto y_c}^{\mathcal{H}_{\Phi_c}}$. For a given sequence $(x^i \in \mathbb{R}^n)_{i=1}^N$, let $\mathrm{Ev}_{(x^i)_{i=1}^N}:L^2(\mathbb{R}^n,\mathbb{P}_x) \rightarrow (\mathbb{R}^N, \langle \cdot, \cdot \rangle_{\mathbb{R}^N} )$ be the function evaluation operator defined as
 \begin{IEEEeqnarray}{rCl}
\mathrm{Ev}_{(x^i)_{i=1}^N}g & := & (g(x^1),\cdots,g(x^N)).
    \end{IEEEeqnarray}
The given data samples can be represented using evaluation operator as
 \begin{IEEEeqnarray}{rCl}
\label{eq_240920240956}\left( y^1_c ,  \cdots,  y^N_c \right) &  = &  \mathrm{Ev}_{(x^i)_{i=1}^N} f_{x \mapsto y_c}  + \left( \xi_c(x^1,y^1),  \cdots,  \xi_c(x^N,y^N) \right).
\end{IEEEeqnarray}
Combining (\ref{eq_100920240903}) and (\ref{eq_240920240956}), we get
 \begin{IEEEeqnarray}{rCl}
\label{eq_090220151301}h_{x \mapsto y_c}^{\mathcal{H}_{\Phi_c}} &  = & ( \widehat{S}^*_{(x^i)_{i=1}^N} \mathrm{Ev}_{(x^i)_{i=1}^N}) f_{x \mapsto y_c}  +  \widehat{S}^*_{(x^i)_{i=1}^N} (\xi_c(x^{1}, y^{1}),  \cdots,  \xi_c(x^{N}, y^{N}) ).
\end{IEEEeqnarray}
\begin{theorem}[Risk for Sample Approximation of the Optimal Solution in RKHS]\label{theorem_040220251005}
The following holds with probability at least $1-\delta$ for any $\delta \in (0,1)$: 
 \begin{IEEEeqnarray}{rCl}
\label{eq_240920242010}\mathop{\mathbb{E}}_{x \sim \mathbb{P}_{x}} \left [ \left | h_{x \mapsto y_c}^{\mathcal{H}_{\Phi_c}}(x)-f_{x \mapsto y_c}^{\mathcal{H}_{\Phi_c}}(x) \right |^2 \right] & \leq & \frac{3}{\sqrt{N}} + \sqrt{\frac{8  \log(1/\delta)}{N}}.
   \end{IEEEeqnarray} 
  \begin{proof}
The proof is provided in five parts:
\paragraph{Part 1:} It is shown in Appendix H that
   \begin{IEEEeqnarray}{rCl}
\label{eq_240920241954} \mathop{\mathbb{E}}_{x \sim \mathbb{P}_{x}} \left [ \left | h_{x \mapsto y_c}^{\mathcal{H}_{\Phi_c}}-f_{x \mapsto y_c}^{\mathcal{H}_{\Phi_c}}(x) \right |^2 \right]& \leq & \left \|  \left(  \widehat{S}^*_{(x^i)_{i=1}^N} \mathrm{Ev}_{(x^i)_{i=1}^N}  - J^* \right) f_{x \mapsto y_c}   \right \|_{\mathcal{H}_{\Phi_c}} +  \left \|     \widehat{S}^*_{(x^i)_{i=1}^N} \left( \xi_c(x^{1}, y^{1}),  \cdots,  \xi_c(x^{N}, y^{N}) \right)  \right \|_{\mathcal{H}_{\Phi_c}}. \IEEEeqnarraynumspace
   \end{IEEEeqnarray}
\paragraph{Part 2:} It is shown in Appendix I that
\begin{IEEEeqnarray}{rCl}
\label{eq_100920241949} \mathop{\mathbb{E}}_{\left((x^i,y^i) \sim \mathbb{P}_{x,y} \right)_{i=1}^{N}} \left [ \left \|  \left(  \widehat{S}^*_{(x^i)_{i=1}^N} \mathrm{Ev}_{(x^i)_{i=1}^N}  - J^* \right) f_{x \mapsto y_c}  \right \|_{\mathcal{H}_{\Phi_c}} \right ]
 & \leq & \frac{2}{\sqrt{N}}.
     \end{IEEEeqnarray}
\paragraph{Part 3:} It is shown in Appendix J that
 \begin{IEEEeqnarray}{rCl}
\label{eq_011020241149}\mathop{\mathbb{E}}_{\left((x^i, y^i) \sim \mathbb{P}_{x, y} \right)_{i=1}^{N}} \left [ \left \| \widehat{S}^*_{(x^i)_{i=1}^N} ( \xi_c(x^{1}, y^{1}),  \cdots,  \xi_c(x^{N}, y^{N}) ) \right \|_{\mathcal{H}_{\Phi_c}} \right ]
 & \leq & \frac{1}{\sqrt{N}}.
   \end{IEEEeqnarray} 
\paragraph{Part 4:} It is shown in Appendix K that we have with probability at least $1-\delta$,
 \begin{IEEEeqnarray}{rCl}
\label{eq_011020241156} \frac{3}{\sqrt{N}} + \sqrt{\frac{8\log(1/\delta)}{N}} & \geq & \left \|  \left(  \widehat{S}^*_{(x^i)_{i=1}^N} \mathrm{Ev}_{(x^i)_{i=1}^N}  - J^* \right) f_{x \mapsto y_c} \right \|_{\mathcal{H}_{\Phi_c}} + \left \| \widehat{S}^*_{(x^i)_{i=1}^N} \left ( \xi_c(x^{1}, y^1),  \cdots,   \xi_c(x^{N}, y^N) \right ) \right \|_{\mathcal{H}_{\Phi_c}}.
   \end{IEEEeqnarray}  
\paragraph{Part 5:} Finally, we get  (\ref{eq_240920242010}) by using (\ref{eq_011020241156}) in (\ref{eq_240920241954}).
 \end{proof}  
\end{theorem}
\subsection{Step 5: Generalized Learning Solution in $L^2(\mathbb{R}^n,\mathbb{P}_{x})$}\label{section_generalized_learning_solution}
The learning solution $ h_{x \mapsto y_c}^{\mathcal{H}_{\Phi_c}} \in \mathcal{H}_{\Phi_c}$ is mapped onto $L^2(\mathbb{R}^n,\mathbb{P}_{x})$ using inverse-operator $(J^*)^{-1}$, i.e., 
 \begin{IEEEeqnarray}{rCl}
\label{eq_011020241911}h_{x \mapsto y_c}  & := & (J^*)^{-1} h_{x \mapsto y_c}^{\mathcal{H}_{\Phi_c}}.
 \end{IEEEeqnarray}
The obtained solution $h_{x \mapsto y_c} $ is referred to as {\em generalized learning solution} reflecting upon the considered generalized kernel function. The learning solution is evaluated in Theorem~\ref{theorem_090220251546} for its risk with respect to the optimal solution $f_{x \mapsto y_c}$.
\begin{theorem}[Risk for Generalized Learning Solution] \label{theorem_090220251546}
The following holds with probability at least $1-\delta$ for any $\delta \in (0,1)$: 
 \begin{IEEEeqnarray}{rCl}
\label{eq_011020241430}\mathop{\mathbb{E}}_{x \sim \mathbb{P}_{x}} \left [ \left |  h_{x \mapsto y_c}(x)-f_{x \mapsto y_c}(x) \right |^2 \right] & \leq & \frac{1}{\left(\| \Phi_c \|_{L^2(\mathbb{R}^n,\mathbb{P}_{x})}^2 \right)^2} \left( \frac{3}{\sqrt{N}} + \sqrt{\frac{8  \log(1/\delta)}{N}} \right).
   \end{IEEEeqnarray} 
  \begin{proof}
Consider
 \begin{IEEEeqnarray}{rCl}
\mathop{\mathbb{E}}_{x \sim \mathbb{P}_{x}} \left [ \left |  h_{x \mapsto y_c}(x)-f_{x \mapsto y_c}(x) \right |^2 \right]   & = & \mathop{\mathbb{E}}_{x \sim \mathbb{P}_{x}} \left [ \left |  \left( \left(J^*\right)^{-1} h_{x \mapsto y_c}^{\mathcal{H}_{\Phi_c}} \right)(x)-\left( \left(J^*\right)^{-1} f_{x \mapsto y_c}^{\mathcal{H}_{\Phi_c}} \right)(x) \right |^2 \right]  \\
\label{eq_011020241427}& = & \frac{1}{\left(\| \Phi_c \|_{L^2(\mathbb{R}^n,\mathbb{P}_{x})}^2 \right)^2} \mathop{\mathbb{E}}_{x \sim \mathbb{P}_{x}} \left [ \left |  h_{x \mapsto y_c}^{\mathcal{H}_{\Phi_c}}(x)-f_{x \mapsto y_c}^{\mathcal{H}_{\Phi_c}}(x) \right |^2 \right].
   \end{IEEEeqnarray} 
Using Theorem (\ref{theorem_040220251005}) in (\ref{eq_011020241427}), we get the result (\ref{eq_011020241430}). 
  \end{proof}
\end{theorem}
Theorem~\ref{theorem_090220251546} allows to bound the error associated to $ h_{x \mapsto y_c}$ in predicting the output $y_c$, as stated in Theorem~\ref{theorem_090220251718}.   
\begin{theorem}[Prediction Error Bound for Generalized Learning Solution]\label{theorem_090220251718}
The following holds with probability at least $1-\delta$ for any $\delta \in (0,1)$:
 \begin{IEEEeqnarray}{rCl}
\nonumber \underbrace{\mathop{\mathbb{E}}_{(x,y) \sim \mathbb{P}_{x,y}} \left [ \left | y_c - h_{x \mapsto y_c}(x) \right |^2 \right]}_{\mbox{mean-squared prediction error}}  & \leq &   \underbrace{\mathop{\mathbb{E}}_{(x,y) \sim \mathbb{P}_{x,y}} \left [ \left | y_c - \mathop{\mathbb{E}}_{y \sim \mathbb{P}_{y | x}} \left[ y_c | x  \right] \right |^2 \right]}_{\mbox{mean-squared disturbance magnitude}} \\
 \label{eq_091220241931} & & {+}\: \frac{1}{\left(\| \Phi_c \|_{L^2(\mathbb{R}^n,\mathbb{P}_{x})}^2 \right)^2} \left( \frac{3}{\sqrt{N}} + \sqrt{\frac{8  \log(1/\delta)}{N}} \right).
   \end{IEEEeqnarray}
\begin{proof}
Consider the prediction error
\begin{IEEEeqnarray}{rCl}
\nonumber \mathop{\mathbb{E}}_{(x,y) \sim \mathbb{P}_{x,y}} \left [ \left | y_c - h_{x \mapsto y_c}(x) \right |^2 \right]  & = & \mathop{\mathbb{E}}_{x \sim \mathbb{P}_{x}} \left [ \left | f_{x \mapsto y_c}(x) - h_{x \mapsto y_c}(x) \right |^2 \right] + \mathop{\mathbb{E}}_{(x,y) \sim \mathbb{P}_{x,y}} \left [ \left | \xi_c(x,y) \right |^2 \right] \\
&& {+}\: 2 \mathop{\mathbb{E}}_{(x,y) \sim \mathbb{P}_{x,y}} \left [ (f_{x \mapsto y_c}(x) - h_{x \mapsto y_c}(x) ) \xi_c(x,y)  \right].
  \end{IEEEeqnarray}
Using (\ref{eq_240920241406}), we get
\begin{IEEEeqnarray}{rCl}
\mathop{\mathbb{E}}_{(x,y) \sim \mathbb{P}_{x,y}} \left [ \left | y_c - h_{x \mapsto y_c}(x) \right |^2 \right]  & = & \mathop{\mathbb{E}}_{x \sim \mathbb{P}_{x}} \left [ \left | f_{x \mapsto y_c}(x) - h_{x \mapsto y_c}(x) \right |^2 \right] + \mathop{\mathbb{E}}_{(x,y) \sim \mathbb{P}_{x,y}} \left [ \left | \xi_c(x,y) \right |^2 \right]. \IEEEeqnarraynumspace
  \end{IEEEeqnarray}
Using (\ref{eq_101220241056}) and Theorem~\ref{theorem_090220251546}, the result is obtained.
\end{proof}
\end{theorem}
\begin{remark}[Robustness of Generalized Learning Solution]\label{remark_100220251448}
  Inequality (\ref{eq_091220241931}) establishes the robustness property in the sense that if the disturbances are small (i.e. $\mathop{\mathbb{E}}_{(x,y) \sim \mathbb{P}_{x,y}} \left [ \left | y_c - \mathop{\mathbb{E}}_{y \sim \mathbb{P}_{y | x}} \left[ y_c | x  \right]\right |^2 \right]$ is small), then the prediction error cannot be large (i.e. $\mathop{\mathbb{E}}_{(x,y) \sim \mathbb{P}_{x,y}} \left [ \left | y_c - h_{x \mapsto y_c}(x) \right |^2 \right]$ cannot be large). As the number of training samples $N$ increases, the upper bound on prediction error decreases. In the limiting case, we have
 \begin{IEEEeqnarray}{rCl}
\lim_{N \to \infty} \frac{\underbrace{\mathop{\mathbb{E}}_{(x,y) \sim \mathbb{P}_{x,y}} \left [ \left | y_c - h_{x \mapsto y_c}(x) \right |^2 \right]}_{\mbox{mean-squared prediction error}}}{ \underbrace{\mathop{\mathbb{E}}_{(x,y) \sim \mathbb{P}_{x,y}} \left [ \left | y_c - \mathop{\mathbb{E}}_{y \sim \mathbb{P}_{y | x}} \left[ y_c | x  \right] \right |^2 \right]}_{\mbox{mean-squared disturbance magnitude}}} & \leq & 1.
  \end{IEEEeqnarray}     
\end{remark}
Theorem~\ref{theorem_100220251831} provides an upper bound on the error in approximating the target function $\mathbb{P}_{y | x}(y_c = 1 | x)$ through $h_{x \mapsto y_c}$.   
\begin{theorem}[Approximation Error Bound for Generalized Learning Solution]\label{theorem_100220251831}
The following holds with probability at least $1-\delta$ for any $\delta \in (0,1)$:
\begin{IEEEeqnarray}{rCl}
\label{eq_070120251306}\underbrace{\mathop{\mathbb{E}}_{x \sim \mathbb{P}_x}\left[ \left| h_{x \mapsto y_c}(x) - \mathbb{P}_{y | x}(y_c = 1 | x) \right|^2\right]}_{\mbox{mean-squared approximation error}} & \leq & \frac{1}{\left(\| \Phi_c \|_{L^2(\mathbb{R}^n,\mathbb{P}_{x})}^2 \right)^2} \left( \frac{3}{\sqrt{N}} + \sqrt{\frac{8  \log(1/\delta)}{N}} \right).
   \end{IEEEeqnarray}  
\begin{proof}
Since $y_c \in \{ 0,1\}$, we have 
 \begin{IEEEeqnarray}{rCl}
\mathop{\mathbb{E}}_{y \sim \mathbb{P}_{y | x}} \left[ y_c | x  \right] & = & \mathbb{P}_{y | x}(y_c = 1 | x).
  \end{IEEEeqnarray}  
Using (\ref{eq_040920240934}), we get 
 \begin{IEEEeqnarray}{rCl}
\label{eq_110220250941} \mathbb{P}_{y | x}(y_c = 1 | x) & = & f_{x \mapsto y_c}(x).
  \end{IEEEeqnarray}
Using (\ref{eq_110220250941}) in Theorem~\ref{theorem_090220251546} leads to the result.
\end{proof}
\end{theorem}
\begin{remark}[Asymptotic Convergence of Generalized Learning Solution]\label{remark_100220251905}
Theorem~\ref{theorem_100220251831} indicates that the approximation error bound decays with an increasing number of training samples. In the limiting case, the approximation error reduces to zero i.e.
\begin{IEEEeqnarray}{rCl}
\lim_{N \to \infty} \; \mathop{\mathbb{E}}_{x \sim \mathbb{P}_x}\left[ \left| h_{x \mapsto y_c}(x) - \mathbb{P}_{y | x}(y_c = 1 | x) \right|^2\right] & = & 0.
   \end{IEEEeqnarray}
\end{remark}
\subsection{Step 6: Identification of Conditions on Kernel and Determination of Hypothesis Space}\label{section_hypothesis_space_determination}
It can be seen from (\ref{eq_011020241911}) and (\ref{eq_100920240903}) that the obtained solution $h_{x \mapsto y_c} $ is given as
\begin{IEEEeqnarray}{rCl}
h_{x \mapsto y_c}(x) & =  & \frac{1}{\| \Phi_c \|_{L^2(\mathbb{R}^n,\mathbb{P}_{x})}^2} \frac{1}{N} \sum_{i=1}^N y_c^i \mathcal{K}_{\Phi_c}(x^i,x) \\
& = & \frac{1}{N\| \Phi_c \|_{L^2(\mathbb{R}^n,\mathbb{P}_{x})}^2}  \sum_{i=1}^{N_c} \mathcal{K}_{\Phi_c}(x^{\mathrm{I}_i^{c}},x) \\
\label{eq_131020241644} & = & \frac{\Phi_c(x)}{N\| \Phi_c \|_{L^2(\mathbb{R}^n,\mathbb{P}_{x})}^2}  \sum_{i=1}^{N_c} \Phi_c(x^{\mathrm{I}_i^{c}}),
   \end{IEEEeqnarray} 
where $(x^{\mathrm{I}_1^{c}},\cdots,x^{\mathrm{I}_{N_c}^{c}} )$ is the sequence of $c^{th}$ class labelled samples, i.e., $y_c^{\mathrm{I}_1^{c}} = \cdots = y_c^{\mathrm{I}_{N_c}^{c}} = 1$. Since $\Phi_c:\mathbb{R}^n \rightarrow [0,1]$, we have for any $x \in \mathbb{R}^n$,
\begin{IEEEeqnarray}{rCl}
\label{eq_050120251431}h_{x \mapsto y_c}(x) & \in & \left[0,\frac{N_c}{N\| \Phi_c, \|_{L^2(\mathbb{R}^n,\mathbb{P}_{x})}^2} \right].
  \end{IEEEeqnarray} 
Since $h_{x \mapsto y_c}$ is viewed as the predictor of $c^{th}$ class label $y_c \in \{0,1\}$, we ensure that $h_{x \mapsto y_c}(x) \in [0,1]$ by constraining the feature-map $\Phi_c$ as
\begin{IEEEeqnarray}{rCl}
\label{eq_100220251930}\| \Phi_c \|_{L^2(\mathbb{R}^n,\mathbb{P}_{x})}^2 & = & \frac{N_c}{N},
  \end{IEEEeqnarray}
resulting in
\begin{IEEEeqnarray}{rCl}
\label{eq_251220241757}h_{x \mapsto y_c}(x) & \in & \left[0,1\right].
  \end{IEEEeqnarray}
\begin{remark}[Justification of the Normalization (\ref{eq_100220251930})]\label{rem_031120250936}
Condition (\ref{eq_100220251930}) is not an ad hoc constraint but a normalization that aligns the feature-map $\Phi_c$ with the class prior. The KAHM-based construction in Section~\ref{section_hypothesis_identification} is designed such that $\Phi_c(x)\in[0,1]$ behaves as an (almost) binary membership score for class $c$: in the idealized case encoded by (\ref{eq_010120252052}) we have $\Phi_c(x)\in\{0,1\}$, and Proposition~\ref{proposition_160220251115} shows that $\Phi_c(x)$ provides a calibrated approximation of the posterior probability $\mathbb{P}_{y | x}(y_c = 1 | x)$. In the binary idealization $\Phi_c(x)\in\{0,1\}$ we obtain $|\Phi_c(x)|^2 = \Phi_c(x)$ and hence $\mathop{\mathbb{E}}_{x \sim \mathbb{P}_x}\left[  |\Phi_c(x)|^2  \right ] = \mathop{\mathbb{E}}_{x \sim \mathbb{P}_x}\left[  \Phi_c(x)  \right ]$. Under the posterior interpretation $\Phi_c(x)\approx \mathbb{P}_{y | x}(y_c = 1 | x)$ we further have, by the law of total expectation,  $ \mathop{\mathbb{E}}_{x \sim \mathbb{P}_x}\left[  \Phi_c(x)  \right ] \approx \mathop{\mathbb{E}}_{x \sim \mathbb{P}_x}\left[  \mathbb{P}_{y | x}(y_c = 1 | x)  \right ] = \mathbb{P}_{y}(y_c = 1)$. Combining these relations yields $\mathop{\mathbb{E}}_{x \sim \mathbb{P}_x}\left[  |\Phi_c(x)|^2  \right ]  \approx \mathbb{P}_{y}(y_c = 1)$. Imposing (\ref{eq_100220251930}) for our design, i.e.
$\mathop{\mathbb{E}}_{x \sim \mathbb{P}_x}\left[  |\Phi_c(x)|^2  \right ] = N_c/N$, is therefore equivalent to requiring $\mathbb{P}_{y}(y_c = 1) \approx N_c/N$, which is the standard assumption that the empirical class frequency $N_c/N$ approximates the true class prior. Thus, in our design of $\Phi_c$, condition (\ref{eq_100220251930}) encodes a natural and statistically consistent alignment with the empirical class prior, rather than an artificial restriction on the hypothesis space. 
\end{remark}
With the kernel feature-map normalization condition (\ref{eq_100220251930}), we define our hypothesis space as
\begin{IEEEeqnarray}{rCl}
\mathcal{M}_{c} & := & \left \{ h_{x \mapsto y_c} =   \frac{\Phi_c(\cdot)}{N\| \Phi_c \|_{L^2(\mathbb{R}^n,\mathbb{P}_{x})}^2} \sum_{i=1}^{N_c} \Phi_c(x^{\mathrm{I}_i^{c}})  \; \mid \; \Phi_c: \mathbb{R}^n \rightarrow [0,1],\; \| \Phi_c \|_{L^2(\mathbb{R}^n,\mathbb{P}_{x})}^2 = \frac{N_c}{N}         \right \} \IEEEeqnarraynumspace \\
\label{eq_150220251829}& = & \left \{ h_{x \mapsto y_c} =   \frac{\Phi_c(\cdot)}{N_c} \sum_{i=1}^{N_c} \Phi_c(x^{\mathrm{I}_i^{c}})  \; \mid \; \Phi_c: \mathbb{R}^n \rightarrow [0,1],\; \| \Phi_c \|_{L^2(\mathbb{R}^n,\mathbb{P}_{x})}^2 = \frac{N_c}{N}         \right \} \\
& = & \left \{ h_{x \mapsto y_c} =   \sum_{i=1}^{N_c} \frac{1}{N_c} \mathcal{K}_{\Phi_c}(\cdot,x^{\mathrm{I}_i^{c}})   \; \mid \; \Phi_c: \mathbb{R}^n \rightarrow [0,1],\; \| \Phi_c \|_{L^2(\mathbb{R}^n,\mathbb{P}_{x})}^2 = \frac{N_c}{N}         \right \} \\
& \subset & \mathcal{H}_{\Phi_c}.
   \end{IEEEeqnarray}
For any $h_{x \mapsto y_c} \in \mathcal{M}_{c} $, we have
\begin{IEEEeqnarray}{rCl}
\label{eq_111220241416}\| h_{x \mapsto y_c} \|_{\mathcal{H}_{\Phi_c}} & = & \frac{\sum_{i=1}^{N_c} \Phi_c(x^{\mathrm{I}_i^{c}})}{N_c},\; \forall \: h_{x \mapsto y_c} \in \mathcal{M}_{c}.
  \end{IEEEeqnarray} 
\subsubsection{Rademacher Complexity of the Hypothesis Space}
For a given data set $\mathcal{D}$ (defined in (\ref{eq_080620242021})), the empirical Rademacher complexity of the hypothesis space $ \mathcal{M}_{c} $ is given as
 \begin{IEEEeqnarray}{rCl}
\widehat{\mathcal{R}}_{\mathcal{D}}( \mathcal{M}_{c} ) & = & \frac{1}{N} \mathop{\mathbb{E}}_{\sigma } \left[ \sup_{h_{x \mapsto y_c} \in \mathcal{M}_{c}  } \sum_{i=1}^N \sigma_i \: h_{x \mapsto y_c}(x^i) \right],
  \end{IEEEeqnarray} 
where $\sigma = (\sigma_1,\cdots,\sigma_N)$ with  $\sigma_1,\cdots,\sigma_N$ as the independent random variables drawn from the Rademacher distribution.  
\begin{theorem}[Bound on Rademacher Complexity of the Hypothesis Space]\label{theorem_110220250938}
Given a dataset $\mathcal{D}$, as defined in (\ref{eq_080620242021}), we have
\begin{IEEEeqnarray}{rCl}
\label{eq_121220240925}\widehat{\mathcal{R}}_{\mathcal{D}}( \mathcal{M}_{c} ) & \leq & \frac{1}{ \sqrt{N}}.
 \end{IEEEeqnarray} 
Thus,
\begin{IEEEeqnarray}{rCl}
\mathop{\mathbb{E}}_{\mathcal{D} \sim (\mathbb{P}_{x,y})^N}\left[ \widehat{\mathcal{R}}_{\mathcal{D}}( \mathcal{M}_{c} ) \right] & \leq & \frac{1}{ \sqrt{N}}.
\end{IEEEeqnarray} 
 \begin{proof}
The proof is provided in Appendix~L.
 \end{proof}
\end{theorem}
\subsubsection{Error Bounds}
Theorem~\ref{theorem_090220251718} and Theorem~\ref{theorem_100220251831} have provided error bounds for the generalized solution using operator-theoretic analysis. Further, the Rademacher complexity of the hypothesis space can be used to derive error bounds for the hypothesis space as in~\cite{kumar2024geometricallyinspiredkernelmachines}. The results obtained by the two approaches can be combined to obtain the tighter bounds as in Theorem~\ref{theorem_120220250928} and Theorem~\ref{theorem_120220251923}.   
\begin{theorem}[Prediction Error Bound for Hypothesis Space]\label{theorem_120220250928}
Given a data set $\mathcal{D}  =   \{(x^i,y^i)  \; \mid \;i \in \{1,2,\cdots,N\} \}   \sim  (\mathbb{P}_{x,y})^N$, for any $h_{x \mapsto y_c} \in \mathcal{M}_{c}$, we have with probability at least $1-\delta$ for any $\delta \in (0,1)$, 
\begin{IEEEeqnarray}{rCl}
\nonumber \lefteqn{\mathop{\mathbb{E}}_{(x,y) \sim \mathbb{P}_{x,y}} \left [ \left | y_c - h_{x \mapsto y_c}(x) \right |^2 \right]} \\
\nonumber & \leq & \min\left( \frac{1}{N} \sum_{i=1}^N |y_c^i - h_{x \mapsto y_c}(x^i) |^2 +  \frac{4}{\sqrt{N}} +  \sqrt{\frac{\log(1/\delta)}{2N}},   \right. \\
 \label{eq_120220251951} & & \left. \mathop{\mathbb{E}}_{(x,y) \sim \mathbb{P}_{x,y}} \left [ \left | y_c - \mathop{\mathbb{E}}_{y \sim \mathbb{P}_{y | x}} \left[ y_c | x  \right] \right |^2 \right] + \frac{1}{\left(N_c/N \right)^2} \left( \frac{3}{\sqrt{N}} + \sqrt{\frac{8  \log(1/\delta)}{N}} \right)  \right).
   \end{IEEEeqnarray}
 \begin{proof}
     The proof is provided in Appendix M.
 \end{proof}
\end{theorem}
\begin{theorem}[Approximation Error Bound for Hypothesis Space]\label{theorem_120220251923}
Given a data set $\mathcal{D}  =   \{(x^i,y^i)  \; \mid \;i \in \{1,2,\cdots,N\} \}   \sim  (\mathbb{P}_{x,y})^N$, for any $h_{x \mapsto y_c} \in \mathcal{M}_{c}$, we have with probability at least $1-\delta$ for any $\delta \in (0,1)$,
\begin{IEEEeqnarray}{rCl}
\nonumber \mathop{\mathbb{E}}_{x \sim \mathbb{P}_x}\left[ \left| h_{x \mapsto y_c}(x) - \mathbb{P}_{y | x}(y_c = 1 | x) \right|^2\right] & \leq & \min\left( \frac{1}{N} \sum_{i=1}^N |y_c^i - h_{x \mapsto y_c}(x^i) |^2 +  \frac{4}{\sqrt{N}} +  \sqrt{\frac{\log(1/\delta)}{2N}}, \right. \\
\label{eq_130220250858}&& \left. \frac{1}{\left(N_c/N \right)^2} \left( \frac{3}{\sqrt{N}} + \sqrt{\frac{8  \log(1/\delta)}{N}} \right)  \right).
\end{IEEEeqnarray}
\begin{proof}
The proof is provided in Appendix N.
\end{proof}
\end{theorem}
\begin{remark}[Comparison with the Existing Error Bounds]\label{rem_230420251610}
Since
\begin{IEEEeqnarray}{rCl}
\mbox{r.h.s. of inequality (\ref{eq_120220251951})}& \leq & \underbrace{\frac{1}{N} \sum_{i=1}^N |y_c^i - h_{x \mapsto y_c}(x^i) |^2 +  \frac{4}{\sqrt{N}} +  \sqrt{\frac{\log(1/\delta)}{2N}}}_{= \mbox{prediction error bound of \cite{kumar2024geometricallyinspiredkernelmachines}}},\\
\mbox{r.h.s. of inequality (\ref{eq_130220250858})}& \leq & \underbrace{\frac{1}{N} \sum_{i=1}^N |y_c^i - h_{x \mapsto y_c}(x^i) |^2 +  \frac{4}{\sqrt{N}} +  \sqrt{\frac{\log(1/\delta)}{2N}}}_{= \mbox{approximation error bound of \cite{kumar2024geometricallyinspiredkernelmachines}}},
\end{IEEEeqnarray}
we achieve tighter bounds on prediction and approximation errors.
\end{remark}
\subsection{Step 7: Hypothesis Implementation in Federated Setting}
We now describe how the RKHS-based method is deployed across decentralized clients in a federated setup. Till-now, we have not fixed the kernel feature-map $\Phi_c$ and the corresponding hypothesis for an implementation in federated setting. Our idea is to leverage KAHMs for defining $\Phi_c$ in such a way that the corresponding hypothesis can be evaluated efficiently from the distributed training data.
\subsubsection{Space Folding Property of a KAHM}
KAHMs exhibit the space folding property in a sense that a KAHM associated to a given set of data samples, folds any arbitrary point in the data space around the data samples.      
\begin{theorem}[KAHM as a Bounded Function~\cite{KAHM}]\label{result_kahm_bounded_function}
The KAHM $\mathcal{A}_{\mathbf{X}}$,  associated to $\mathbf{X} = \left[\begin{IEEEeqnarraybox*}[][c]{,c/c/c,} x^1 & \cdots & x^N \end{IEEEeqnarraybox*} \right]^T$ with $x^1,\cdots,x^N \in \mathbb{R}^n$,  is a bounded function on $\mathbb{R}^n$ such that for any $x \in \mathbb{R}^n$, 
\begin{IEEEeqnarray}{rCl}
\label{eq_100120231400} \| \mathcal{A}_{\mathbf{X}}(x)\| & < &   \left\|\mathbf{X} \right \|_2\left(1 + \frac{nN^2}{2\|\mathbf{X}\|_F^2} \right).
\end{IEEEeqnarray}  
Thus, the image of $\mathcal{A}_{\mathbf{X}}$ is bounded such that 
 \begin{IEEEeqnarray}{rCl}
 \mathcal{A}_{\mathbf{X}}[\mathbb{R}^n]& \subset &\left\{ x \in \mathbb{R}^n \; \mid \;  \| x\| < \left\|\mathbf{X} \right \|_2 \left(1 + \frac{nN^2}{2\|\mathbf{X}\|_F^2} \right) \right \}.
   \end{IEEEeqnarray} 
\end{theorem} 
\begin{theorem}[KAHM Induced Distance Measure~\cite{KAHM}]\label{result_ratio_distances}
The ratio of the distance of a point $x\in \mathbb{R}^n$ from its image under $\mathcal{A}_{\mathbf{X}}$ to the distance of $x$ from $\{x^1,\cdots,x^N\}$ evaluated as $\left \|\left[\begin{IEEEeqnarraybox*}[][c]{,c/c/c,} x - x^1 & \cdots & x - x^N \end{IEEEeqnarraybox*} \right] \right \|_2$ remains upper bounded as
 \begin{IEEEeqnarray}{rCl}
\label{eq_100120231432} \frac{ \left \| x - \mathcal{A}_{\mathbf{X}}(x) \right \| }{\left \|\left[\begin{IEEEeqnarraybox*}[][c]{,c/c/c,} x - x^1 & \cdots & x - x^N \end{IEEEeqnarraybox*} \right] \right \|_2} & < &   1 + \frac{n N^2}{2 \|\mathbf{X}\|_F^2}.
   \end{IEEEeqnarray}     
   \end{theorem}
Theorem~\ref{result_kahm_bounded_function} establishes the boundedness property of the KAHM. Theorem~\ref{result_ratio_distances} states that if a point $x$ is close to the points $\{x^1,\cdots,x^N\}$, then the value $\left \| x - \mathcal{A}_{\mathbf{X}}(x) \right \|$ cannot be large. Theorem~\ref{result_kahm_bounded_function} and Theorem~\ref{result_ratio_distances} indicate that any arbitrary point $x \in \mathbb{R}^n$ can be mapped to a point closer to data samples $\mathbf{X}$ through a KAHM, and thus the space folding property is established. This is illustrated in Fig.~\ref{fig_demo_space_folding_1}, where the KAHM folds the data space around given data samples. 
\begin{definition}[Space Folding Measure]\label{definition_150220251240}
To evaluate the amount of folding required for an arbitrary point $x \in \mathbb{R}^n$ to map it (by the KAHM $\mathcal{A}_{\mathbf{X}}$) to a point closer to data samples $\mathbf{X}$, we define a space folding measure, $\mathcal{T}_{ \mathbf{X}}: \mathbb{R}^n \rightarrow [0,1]$, associated to data samples $\mathbf{X}$, as
 \begin{IEEEeqnarray}{rCl}
\label{eq_170220251438} \mathcal{T}_{ \mathbf{X}}(x) & := & \begin{cases}
 \sqrt{\frac{1}{2} \left(\left|\mathcal{T}_{ \mathbf{X}}^{\text{Euc}}(x) \right|^2 + \left| \mathcal{T}_{ \mathbf{X}}^{\text{Cos}}(x) \right|^2\right)} & \mbox{option 1} \\
  \mathcal{T}_{ \mathbf{X}}^{\text{Euc}}(x) \mathcal{T}_{ \mathbf{X}}^{\text{Cos}}(x) & \mbox{option 2} \\
   \min\left( \mathcal{T}_{ \mathbf{X}}^{\text{Euc}}(x) , \mathcal{T}_{ \mathbf{X}}^{\text{Cos}}(x) \right) & \mbox{option 3} \\
  \max\left( \mathcal{T}_{ \mathbf{X}}^{\text{Euc}}(x) , \mathcal{T}_{ \mathbf{X}}^{\text{Cos}}(x) \right) &  \mbox{option 4}, 
\end{cases}
    \end{IEEEeqnarray}  
where 
 \begin{IEEEeqnarray}{rCl}
\mathcal{T}_{ \mathbf{X}}^{\text{Euc}}(x) &:= & 1 - \exp\left( - \|x - \mathcal{A}_{\mathbf{X}}(x) \|  \right) \\
\mathcal{T}_{ \mathbf{X}}^{\text{Cos}}(x) & := &  \frac{1}{\pi} \arccos{\left(\frac{ \left(\mathcal{A}_{\mathbf{X}}(x)\right)^T x}{ \|\mathcal{A}_{\mathbf{X}}(x) \| \| x\| }\right)}. 
 \end{IEEEeqnarray} 
 The space folding measure $\mathcal{T}_{\mathbf{X}}$ combines both Euclidean distance and cosine distance to define a composite measure of the distance between $x$ and $\mathcal{A}_{\mathbf{X}}(x)$. Since there are different possibilities to combine the Euclidean and cosine distances resulting in different possible definitions of the space folding measure, (\ref{eq_170220251438}) provides four different possibilities among others. Fig.~\ref{fig_demo_space_folding_2} displays the color-plot of the space folding measure function associated to a set of 2-dimensional data samples.
\begin{figure*}[!h]
\centerline{\subfigure[The given 2-dimensional data samples $\mathbf{X}$ have been marked in blue color as ``*'', and a red line connects a point to its image under KAHM (i.e. $x$ is connected to $\mathcal{A}_{\mathbf{X}}(x)$ through a line).]{\includegraphics[width=0.48\textwidth]{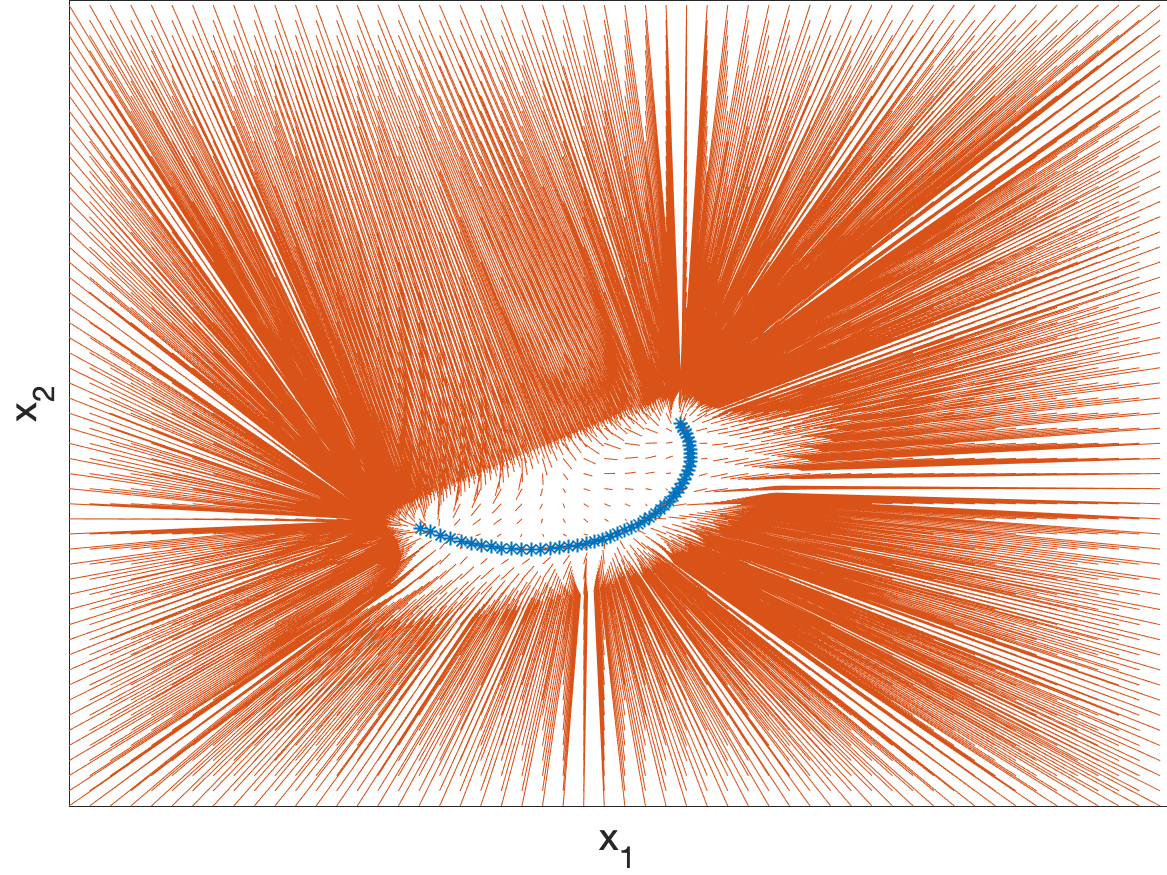}\label{fig_demo_space_folding_1}} \hfil
\subfigure[The color plot of the space folding measure function $\mathcal{T}_{ \mathbf{X}}$ (option 1). The given 2-dimensional data samples $\mathbf{X}$ have been marked in white color as ``*''.]{\includegraphics[width=0.48\textwidth]{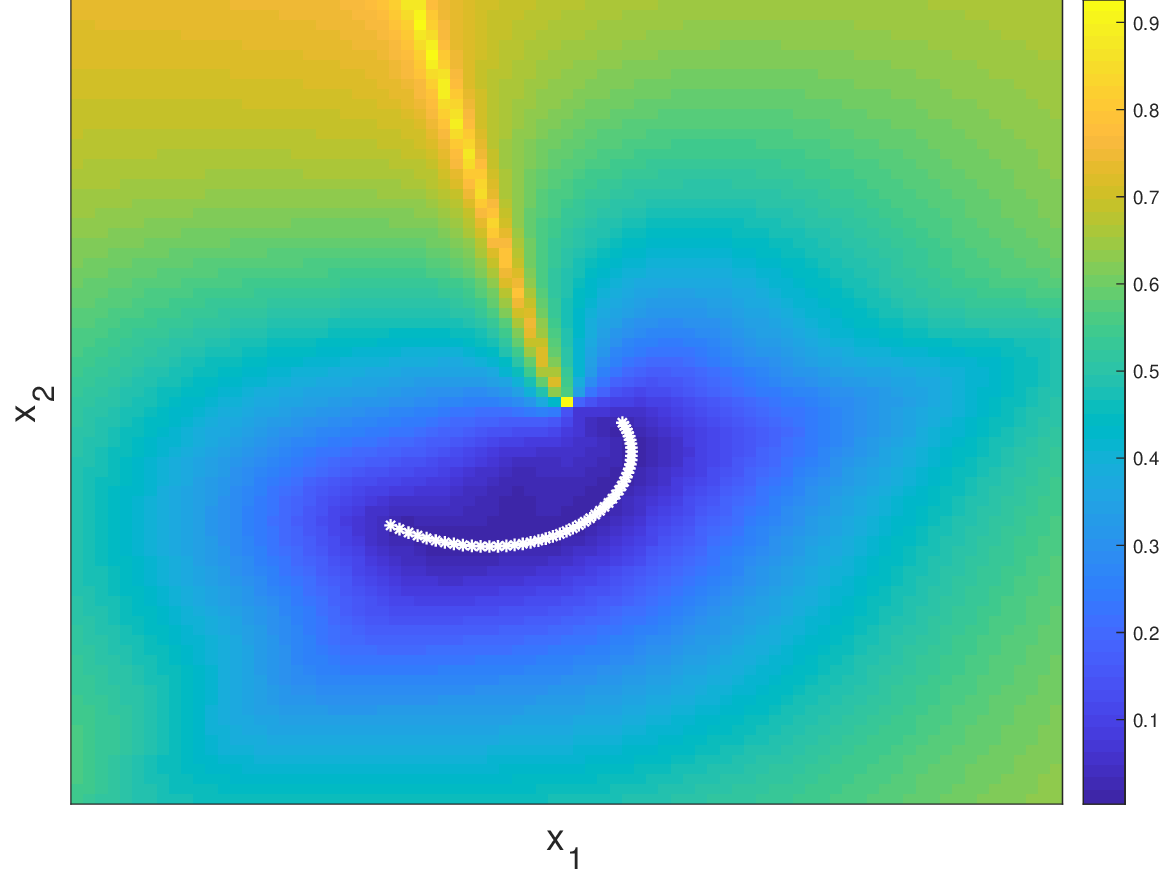}\label{fig_demo_space_folding_2}}}
\caption{An illustration of the space folding property possessed by a KAHM.}
\label{fig_demonstration_space_folding}
\end{figure*}
\end{definition}
\begin{definition}[Space Folding Measure Associated to Distributed Data]\label{definition_150220251244}
Under the scenario that total data samples are distributed among $Q$ number of parties such that matrix $\mathbf{X}^q$ represents local data samples owned by $q^{th}$ party, one possible way to define a global space folding measure, $\mathcal{T}_{ \mathbf{X}^1,\cdots,\mathbf{X}^Q}:\mathbb{R}^n \rightarrow [0,1]$, associated to distributed data samples $\mathbf{X}^1,\cdots,\mathbf{X}^Q$, is as follows
\begin{IEEEeqnarray}{rCl}
\mathcal{T}_{ \mathbf{X}^1,\cdots,\mathbf{X}^Q}(x) & := & \min_{q \in \{1,\cdots,Q\}} \; \mathcal{T}_{ \mathbf{X}^q}(x).
 \end{IEEEeqnarray}
  \begin{figure*}[!h]
 \centerline{\subfigure[The color plot of the space folding measure function $\mathcal{T}_{ \mathbf{X}^1}$ (option 1) associated to data samples $\mathbf{X}^1$ (marked in white color as ``*'').]{\includegraphics[width=0.23\textwidth]{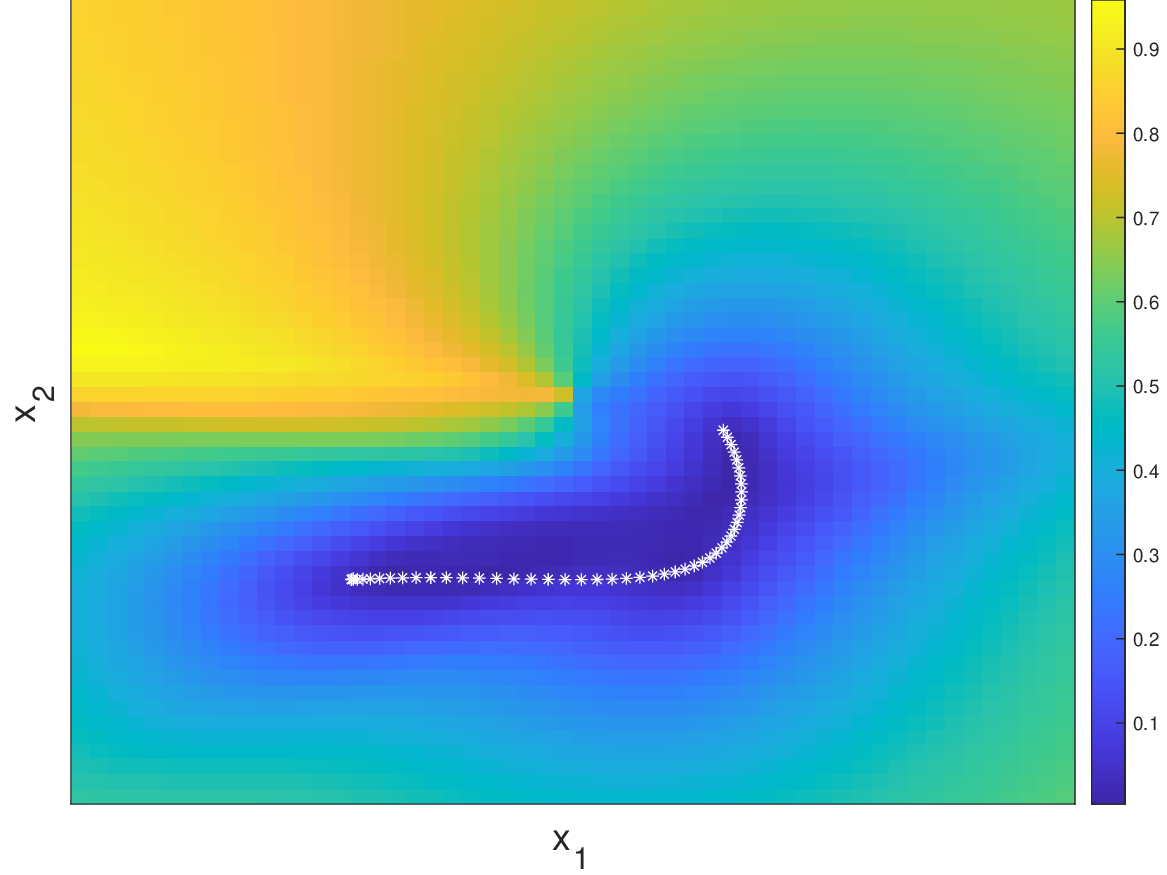}\label{fig_demo_distributed_space_folding_1}} \hfil 
 \subfigure[The color plot of the space folding measure function $\mathcal{T}_{ \mathbf{X}^2}$ (option 1) associated to data samples $\mathbf{X}^2$ (marked in white color as ``*'').]{\includegraphics[width=0.23\textwidth]{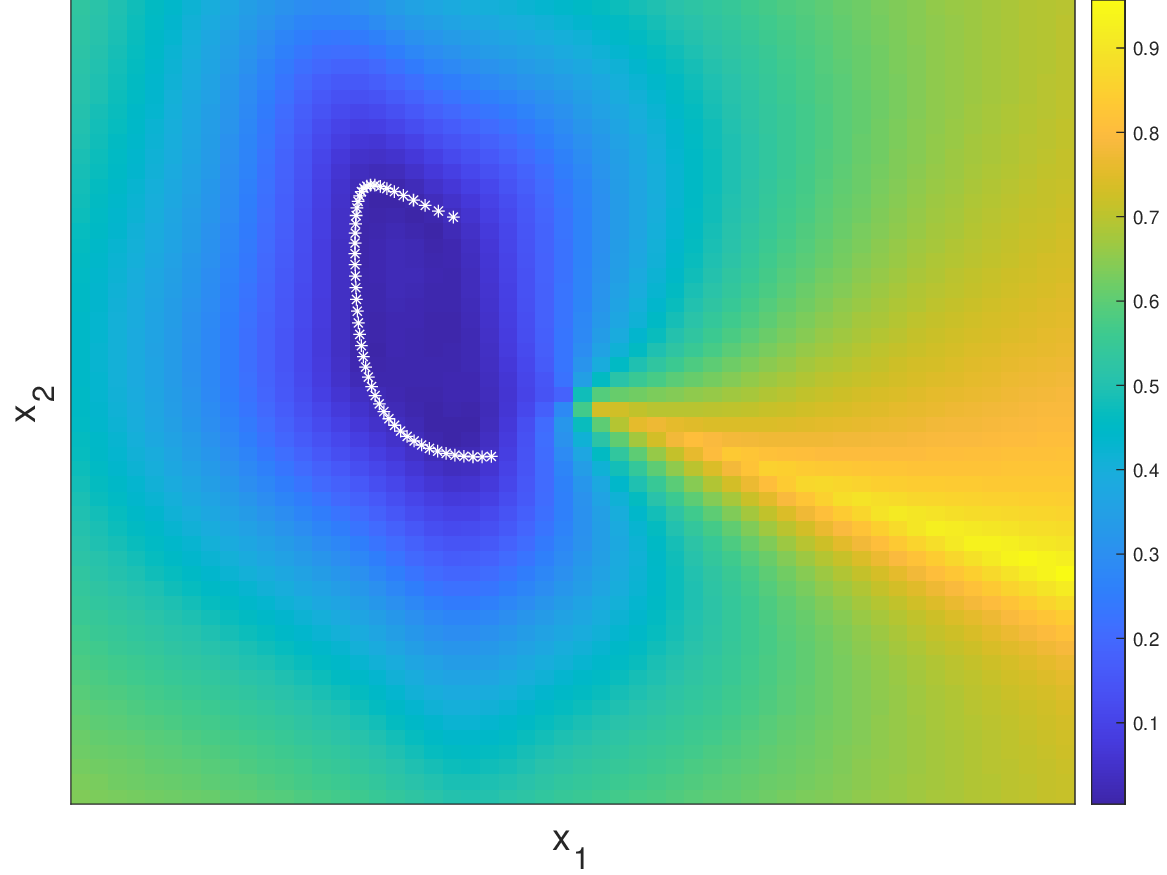}\label{fig_demo_distributed_space_folding_2}} \hfil
 \subfigure[The color plot of the space folding measure function $\mathcal{T}_{ \mathbf{X}^3}$ (option 1) associated to data samples $\mathbf{X}^3$ (marked in white color as ``*'').]{\includegraphics[width=0.23\textwidth]{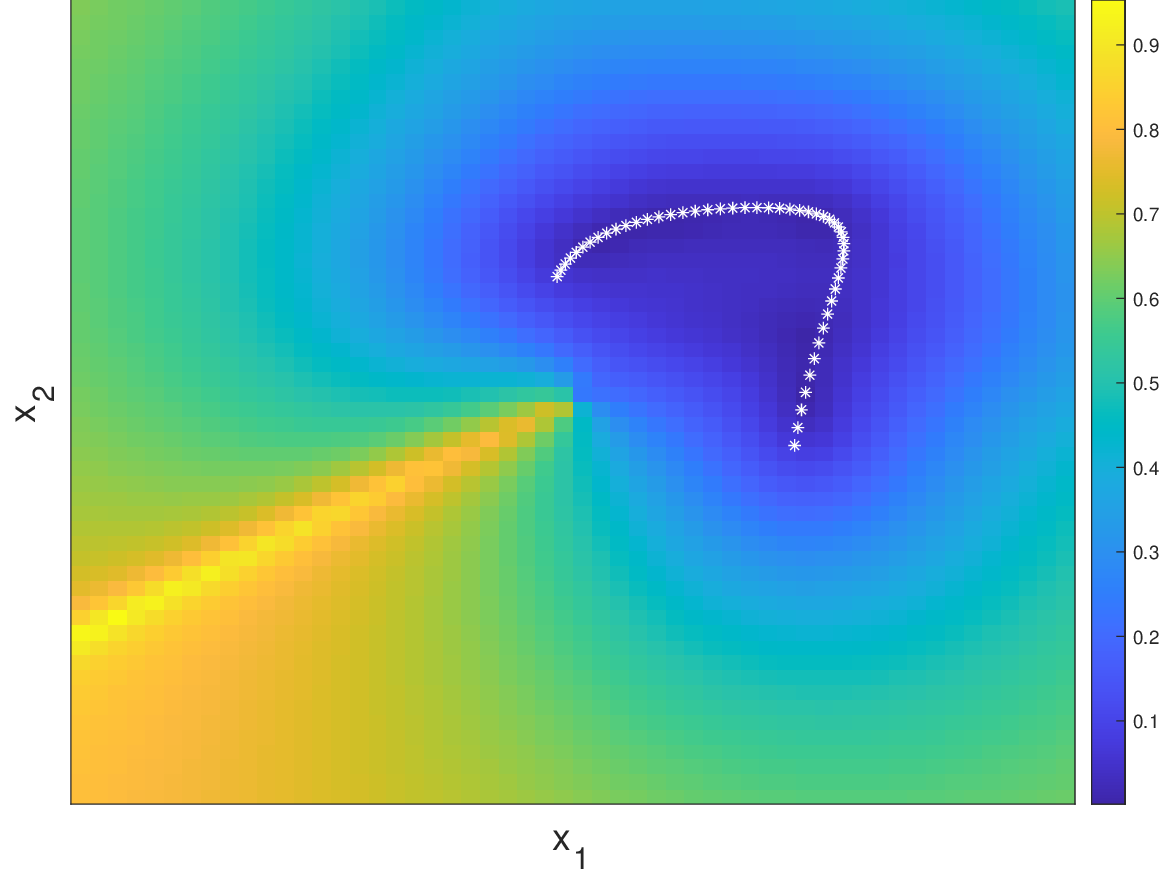}\label{fig_demo_distributed_space_folding_3}} \hfil
 \subfigure[The color plot of the global space folding measure function $\mathcal{T}_{ \mathbf{X}^1,\mathbf{X}^2,\mathbf{X}^3}$ associated to total data samples (marked in white color as ``*'').]{\includegraphics[width=0.23\textwidth]{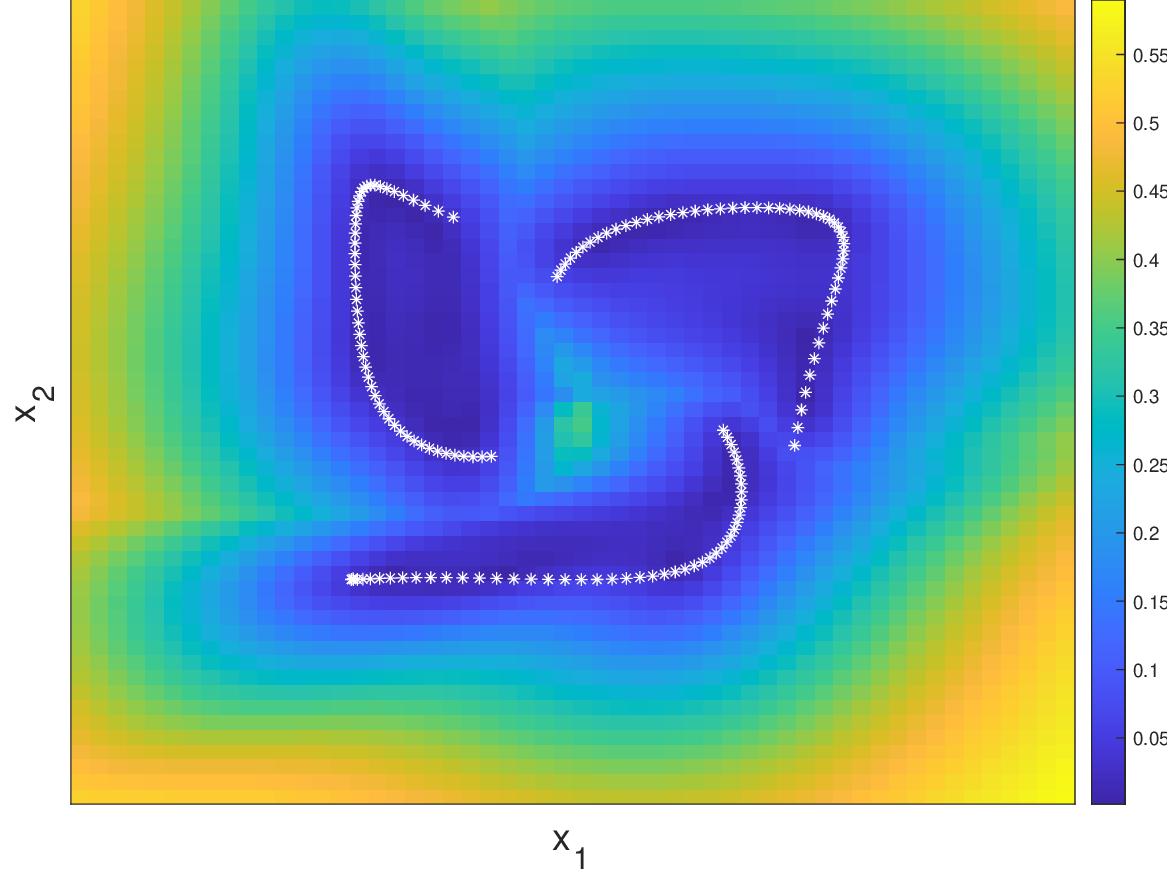}\label{fig_demo_distributed_space_folding_4}}}
\caption{An example of the global space folding measure associated to the distributed data.}
\label{fig_demonstration_distributed_space_folding}
 \end{figure*}
Fig.~\ref{fig_demonstration_distributed_space_folding} shows an example of the global space folding measure associated to distributed 2-dimensional data samples. 
\end{definition}
\subsubsection{Defining Kernel Feature-Map Using Space Folding Measure and a Hypothesis}\label{section_hypothesis_identification}
Our approach is to define the kernel feature-map based on the space folding measure as
\begin{equation}
\label{eq_010120252052}\Phi_c^*(x)  := 
\begin{cases}
1 & \text{if } \mathcal{T}_{ \mathbf{X}^{c,1}, \mathbf{X}^{c,2},\cdots,\mathbf{X}^{c,Q}}(x) = \mathop{\min}_{c \in \{1,2,\cdots,C\}}\;\mathcal{T}_{ \mathbf{X}^{c,1}, \mathbf{X}^{c,2},\cdots,\mathbf{X}^{c,Q}}(x),
\\
0 & \text{otherwise},
\end{cases}
\end{equation}
where $\mathcal{T}_{ \mathbf{X}^{c,1}, \mathbf{X}^{c,2},\cdots,\mathbf{X}^{c,Q}}$ is the global space folding measure associated to $c^{th}$ class labelled samples that are distributed among $Q$ different clients, and $\mathbf{X}^{c,q}$ is the matrix of $c^{th}$ class labelled and $q^{th}$ client owned samples. {\em Definition of the kernel feature-map (\ref{eq_010120252052}) implies that $\Phi_c^*(x)$ is equal to 1, if, among all the classes, $x$ can be mapped (by the KAHMs) to the point closer to the samples of $c^{th}$ class with the least amount of folding}.
\begin{remark}[The Interpretation of the Space Folding Kernel $\mathcal{K}_{\Phi_c^*}$]\label{remark_240320250940}
$\mathcal{K}_{\Phi_c^*}(x,x')$ will be equal to 1, if, among all the classes, both $x$ and $x'$ can be mapped (by the KAHMs) to the points closer to the samples of $c^{th}$ class with the least amount of folding. Similarly, $\mathcal{K}_{\Phi_c^*}(x,x')$ will be equal to 0, if, either of $x$ and $x'$ cannot be  mapped (by the KAHMs) to the points closer to the samples of $c^{th}$ class with the least amount of folding. Thus, $\mathcal{K}_{\Phi_c^*}(x,x')$ estimates an association (of $x$ and $x'$ as together) to the $c^{th}$ class.       
\end{remark}
\begin{assumption}[]\label{assumption_1}
It is reasonable to assume for the training data samples that the global space folding measure associated to $c^{th}$ class (i.e. $\mathcal{T}_{ \mathbf{X}^{c,1}, \mathbf{X}^{c,2},\cdots,\mathbf{X}^{c,Q}}$) will take minimum values for $c^{th}$ class labelled samples, i.e., 
\begin{equation}
\label{eq_260120251904}\Phi_c^*(x^i)  = 
\begin{cases}
1 & \text{if } i  \in \mathcal{I}^{c},
\\
0 & \text{if } i  \notin \mathcal{I}^{c}.
\end{cases},\; \forall i \in \{1,\cdots,N\}.
\end{equation}
\end{assumption}
\begin{assumption}[]\label{assumption_2}
The number of training data samples is sufficiently large i.e. $N \gg 1$, so that $\| \Phi_c^* \|_{L^2(\mathbb{R}^n,\mathbb{P}_{x})}^2$ can be approximated by sample-averaging: 
\begin{IEEEeqnarray}{rCl}
\label{eq_020120252131}\| \Phi_c^* \|_{L^2(\mathbb{R}^n,\mathbb{P}_{x})}^2 & = & \frac{1}{N} \sum_{i=1}^N |\Phi_c^*(x^i)|^2.
 \end{IEEEeqnarray} 
\end{assumption}
\begin{remark}[Consistency of Normalization Condition (\ref{eq_100220251930}) with the Space Folding Kernel]
Using (\ref{eq_260120251904}) in (\ref{eq_020120252131}), we have
\begin{IEEEeqnarray}{rCl}
\label{eq_030120251445}\| \Phi_c^* \|_{L^2(\mathbb{R}^n,\mathbb{P}_{x})}^2 & = & \frac{N_c}{N}.
 \end{IEEEeqnarray} 
 That is, KAHM-induced feature-map $\Phi_c^*$ satisfies the normalization condition (\ref{eq_100220251930}) under Assumptions~\ref{assumption_1}-\ref{assumption_2}. 
\end{remark}
\begin{proposition}[]\label{proposition_150220251924}
Under Assumption~\ref{assumption_1} and Assumption~\ref{assumption_2}, we have
\begin{IEEEeqnarray}{rCl}
\label{eq_150220252029}\Phi_c^* & \in & \mathcal{M}_c.
 \end{IEEEeqnarray}
\begin{proof}
Define
\begin{IEEEeqnarray}{rCl}
\label{eq_150220251831}h_{x \mapsto y_c}^* & := &   \frac{\Phi_c^*(\cdot)}{N_c} \sum_{i=1}^{N_c} \Phi_c^*(x^{\mathrm{I}_i^{c}}).
 \end{IEEEeqnarray}
It follows from (\ref{eq_030120251445}) and (\ref{eq_150220251831}) that 
\begin{IEEEeqnarray}{rCl}
\label{eq_150220252027}h_{x \mapsto y_c}^* & \in & \mathcal{M}_{c}.
 \end{IEEEeqnarray}
Using (\ref{eq_260120251904}) in (\ref{eq_150220251831}), we have
\begin{IEEEeqnarray}{rCl}
\label{eq_150220252028} h_{x \mapsto y_c}^* & = & \Phi_c^*.
 \end{IEEEeqnarray}
Combining (\ref{eq_150220252027}) and (\ref{eq_150220252028}) leads to (\ref{eq_150220252029}).
\end{proof}
\end{proposition}
\begin{proposition}[Approximation Error Bound for $\Phi_c^*$]\label{proposition_160220251115}
Given a data set $\mathcal{D}  =   \{(x^i,y^i)  \; \mid \;i \in \{1,2,\cdots,N\} \}   \sim  (\mathbb{P}_{x,y})^N$, under Assumption~\ref{assumption_1} and Assumption~\ref{assumption_2}, we have with probability at least $1-\delta$ for any $\delta \in (0,1)$,
\begin{IEEEeqnarray}{rCl}
\mathop{\mathbb{E}}_{x \sim \mathbb{P}_x}\left[ \left| \Phi_c^*(x) - \mathbb{P}_{y | x}(y_c = 1 | x) \right|^2\right]
 & \leq & \min\left(  \frac{4}{\sqrt{N}} +  \sqrt{\frac{\log(1/\delta)}{2N}},  \frac{1}{\left(N_c/N \right)^2} \left( \frac{3}{\sqrt{N}} + \sqrt{\frac{8  \log(1/\delta)}{N}} \right)  \right).
\end{IEEEeqnarray}
\begin{proof}
Since $\Phi_c^*  \in  \mathcal{M}_c$, it follows from Theorem~\ref{theorem_120220251923} that
we have with probability at least $1-\delta$ for any $\delta \in (0,1)$,
\begin{IEEEeqnarray}{rCl}
\nonumber \mathop{\mathbb{E}}_{x \sim \mathbb{P}_x}\left[ \left| \Phi_c^*(x) - \mathbb{P}_{y | x}(y_c = 1 | x) \right|^2\right] & \leq & \min\left( \frac{1}{N} \sum_{i=1}^N |y_c^i - \Phi_c^*(x^i) |^2 +  \frac{4}{\sqrt{N}} +  \sqrt{\frac{\log(1/\delta)}{2N}}, \right. \\
\label{eq_160220251131}&& \left. \frac{1}{\left(N_c/N \right)^2} \left( \frac{3}{\sqrt{N}} + \sqrt{\frac{8  \log(1/\delta)}{N}} \right)  \right).
\end{IEEEeqnarray}
Due to (\ref{eq_260120251904}),
\begin{IEEEeqnarray}{rCl}
\label{eq_030120251521}\Phi_c^*(x^i) & = & y^i_c,\; \forall i \in \{1,\cdots,N\}.
 \end{IEEEeqnarray} 
Hence the result follows.
\end{proof}
\end{proposition}
\begin{remark}[Practical Significance of Proposition~\ref{proposition_160220251115}]\label{remark_240320251147}
Proposition~\ref{proposition_160220251115} allows to make the following approximation:
\begin{IEEEeqnarray}{rCl}
\label{eq_160220251523}\Phi_c^*(x) & \approx & \mathbb{P}_{y | x}(y_c = 1 | x).
 \end{IEEEeqnarray} 
That is, $\Phi_c^*(x)$ estimates the probability that $x$ is associated to the $c^{th}$ class.
\end{remark}
\subsubsection{Federated Learning Applications}
The fact, $\Phi^*_c(x)$ (which is the estimated probability of $x$ being associated to the $c^{th}$ class) can be evaluated from the distributed data, is leveraged for FL, as illustrated in Fig.~\ref{fig_federated_learning}. The hypothesis $\Phi^*_c$ is inferred for all $c \in \{1,2,\cdots,C\}$ from the locally computed space folding measures using (\ref{eq_010120252052}). The global classifier, $\widehat{\mathcal{C}}: \mathbb{R}^n \rightarrow \{1,2,\cdots,C \}$, is defined as
\begin{IEEEeqnarray}{rCl}
\label{eq_250320251819} \widehat{\mathcal{C}}(x) & := & \mathop{\argmax}_{c \in \{1,2,\cdots,C\}} \Phi_c^*(x).
 \end{IEEEeqnarray} 
\begin{figure}[!h]  
\centering
\scalebox{0.9}{\begin{tikzpicture}[scale=0.9]
\path[fill=green!10](-3,-10.25)--(3,-10.25)--(3,-0.5)--(-3,-0.5)--cycle;
\draw[green,line width = 0.25mm](-3,-10.25)--(3,-10.25)--(3,-0.5)--(-3,-0.5)--cycle;
\draw (0,-9.75) node[]{\bfseries $\begin{array}{c} \mbox{\scriptsize Client 1} \\ \mbox{\scriptsize \faMale\:\faFemale} \end{array}$};
\draw (0,-8.5) node[rounded corners,draw](n1){\footnotesize $\begin{array}{c}\mbox{training data samples}  \\ \mbox{\small $\{\mathbf{X}^{c,1}  \}_{c=1}^C$} \end{array}$};
\draw (0,-6.55) node[rounded corners,draw](nadd0){\footnotesize $\begin{array}{c}\mbox{rows of $\mathbf{X}^{c,1}$ divided into} \\ \mbox{\small $S_{c,1}$ parts of nearly same size}  \\ \mbox{\small $\{ \mathbf{X}^{c,1}_1,\cdots,\mathbf{X}^{c,1}_{S_{c,1}}\}_{c=1}^C$} \end{array}$};
\draw (0,-4.35) node[rounded corners,draw](nadd1){\footnotesize $\begin{array}{c}\mbox{local space folding measures}  \\ \mbox{defined by (\ref{eq_080920251412})} \\ \mbox{\small $\{ \mathcal{T}_{\mathbf{X}^{c,1}_1,\cdots,\mathbf{X}^{c,1}_{S_{c,1}}} \}_{c=1}^C$}\end{array}$};
\draw[-latex,line width=0.2mm] (n1) to [out=90,in=-90] (nadd0);  
\draw[-latex,line width=0.2mm] (nadd0) to [out=90,in=-90] (nadd1); 
\draw (0,-2.15) node[rounded corners,draw](n4){ \footnotesize $\begin{array}{c}\mbox{local evaluations}  \\ \mbox{\small $\{  \mathcal{T}_{\mathbf{X}^{c,1}_1,\cdots,\mathbf{X}^{c,1}_{S_{c,1}}}(x) \}_{c=1}^C$} \mbox{ released} \\ \mbox{under differential privacy or FHE} \end{array}$};
\draw[-latex,line width=0.2mm] (nadd1) to [out=90,in=-90] (n4);
\path[fill=yellow!10](10-1,-10.25)--(14+1,-10.25)--(14+1,-0.5)--(10-1,-0.5)--cycle;
\draw[yellow,line width = 0.25mm](10-1,-10.25)--(14+1,-10.25)--(14+1,-0.5)--(10-1,-0.5)--cycle;
\draw (11.75,-9.75) node[]{\bfseries $\begin{array}{c} \mbox{\scriptsize Client Q} \\ \mbox{\scriptsize \faMale\:\faFemale} \end{array}$};
\draw (12,-2.15) node[rounded corners,draw](n11){ \footnotesize $\begin{array}{c}\mbox{local evaluations} \\ \mbox{\small $\{ \mathcal{T}_{\mathbf{X}^{c,Q}_1,\cdots,\mathbf{X}^{c,Q}_{S_{c,Q}}} (x) \}_{c=1}^C$}   \mbox{ released} \\ \mbox{under differential privacy or FHE} \end{array}$};
\draw (12,-4.35) node[rounded corners,draw](nadd2){\footnotesize $\begin{array}{c}\mbox{local space folding measures}  \\ \mbox{defined by (\ref{eq_080920251412})} \\ \mbox{\small $\{ \mathcal{T}_{\mathbf{X}^{c,Q}_1,\cdots,\mathbf{X}^{c,Q}_{S_{c,Q}}} \}_{c=1}^C$}\end{array}$};
\draw[-latex,line width=0.2mm] (nadd2) to [out=90,in=-90] (n11);  
\draw (12,-6.55) node[rounded corners,draw](nadd_1){\footnotesize $\begin{array}{c}\mbox{rows of $\mathbf{X}^{c,Q}$ divided into} \\ \mbox{\small $S_{c,Q}$ parts of nearly same size}  \\ \mbox{\small $\{ \mathbf{X}^{c,Q}_1,\cdots,\mathbf{X}^{c,Q}_{S_{c,Q}}\}_{c=1}^C$} \end{array}$};
\draw (12,-8.5) node[rounded corners,draw](n15){\footnotesize $\begin{array}{c}\mbox{training data samples}  \\ \mbox{\small $\{\mathbf{X}^{c,Q}  \}_{c=1}^C$} \end{array}$};
\draw[-latex,line width=0.2mm] (n15) to [out=90,in=-90] (nadd_1);  
\draw[-latex,line width=0.2mm] (nadd_1) to [out=90,in=-90] (nadd2); 
\path[fill=blue!10](4,-8.25)--(8,-8.25)--(8,-0.5)--(4,-0.5)--cycle;  
\draw[blue!40,line width = 0.25mm](4,-8.25)--(8,-8.25)--(8,-0.5)--(4,-0.5)--cycle; 
\draw (6,-7.75) node[]{\bfseries $\begin{array}{c} \mbox{\scriptsize User} \\ \mbox{\scriptsize \faUser} \end{array}$};
  \draw (6,-6.25) node[](n20){ \footnotesize $\begin{array}{c}\mbox{input} \\ \mbox{\small $x$} \end{array}$};
\draw[thick,line width=0.2mm](n20) -- (6,-4);
\draw[thick,line width=0.2mm](3.65,-4) -- (8.35,-4);
\draw[-latex,thick,line width=0.2mm] (3.65,-4) to [out=180,in=0] (n4);   
 \draw[-latex,thick,line width=0.2mm] (8.35,-4) to [out=0,in=180] (n11); 
  \node[cloud,
    draw = gray,
    fill = cyan!5,
    minimum width = 11cm,
    minimum height = 3.5cm,
    cloud puffs = 18] (c) at (6,2) {};
\draw (6,1.85) node[rounded corners,draw, fill=gray!5](n8){ \footnotesize $\begin{array}{c} \mbox{\small $ \{\Phi^*_c(x) \}_{c=1}^C$} \\ \mbox{evaluated using (\ref{eq_010120252052}) and (\ref{eq_150920251220})}  \end{array}$};
\draw[-latex,line width=0.2mm,cyan] (n4) to [out=90,in=180] (n8);  
\draw (6,2.8) node[]{\bfseries {\scriptsize $\begin{array}{c} \mbox{hypothesis inference using space folding measures} \end{array}$  }};
\draw[-latex,line width=0.2mm,cyan] (n11) to [out=90,in=0] (n8);  
\draw (6,-1.5) node[rounded corners,draw, fill = gray!5](n17){ \footnotesize $\begin{array}{c}\mbox{output} \\ \mbox{\small $\{\Phi^*_c(x) \}_{c=1}^C$}  \end{array}$};
\draw[-latex,line width=0.2mm,cyan] (n8) to [out=-90,in=90] (n17);  
\end{tikzpicture}}
\caption{The proposed space folding measure-based methodology, referred to as \texttt{SFM}, estimates for a given input $x$ the probability of $c^{th}$ class, $\Phi^*_c(x)$, without imposing statistical assumptions on clients' data distributions, thereby ensuring robustness towards statistical heterogeneity.}
\label{fig_federated_learning}
\end{figure}
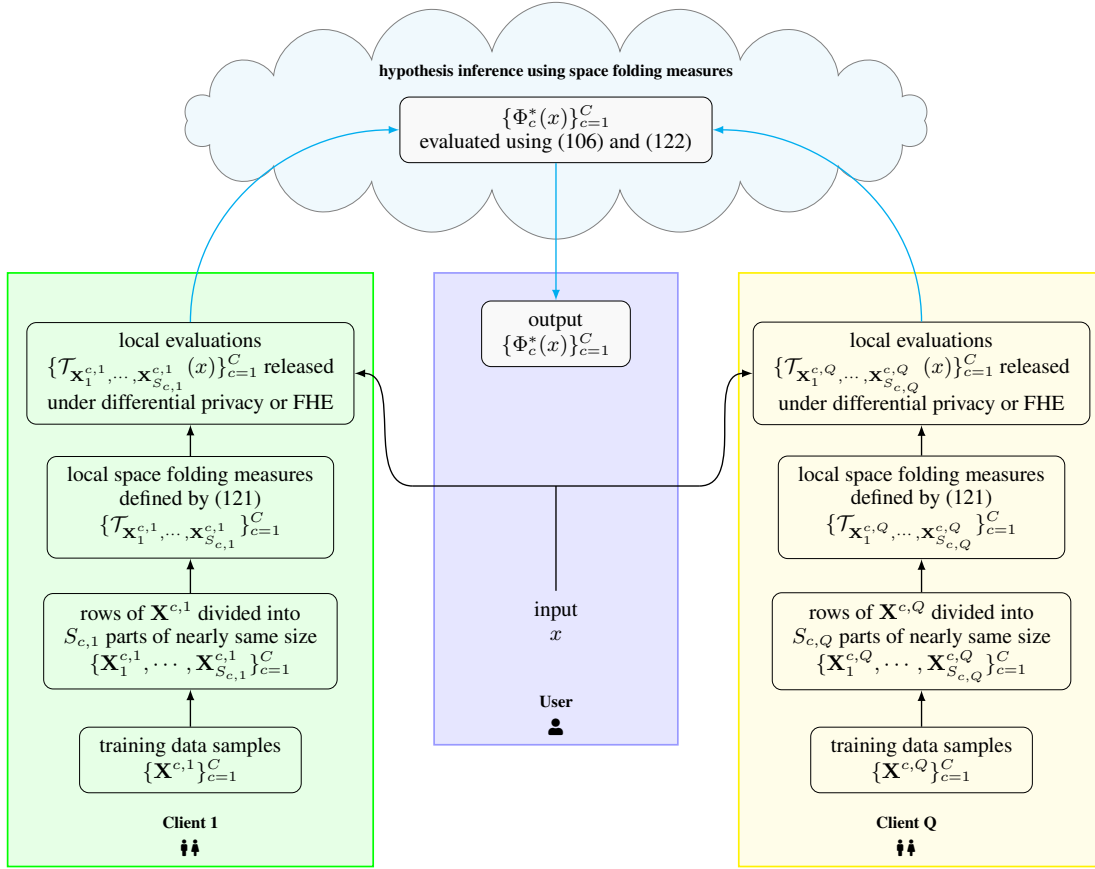
\begin{remark}[Batch KAHM Modeling for Enhanced Accuracy]\label{remark_batch_processing}
To enhance the KAHM modeling accuracy of each class's data samples (which is crucial for datasets with long-tailed imbalance), the total samples are partitioned into subsets and each subset is modelled through a separate KAHM. If $ |\mathcal{I}^{c,q}|$ (i.e. the number of $c^{th}$ class labelled samples that are owned by client $q$) is more than a specified number $N_b$ (representing the batch-size of samples to be modeled by a KAHM), then we have  
\begin{IEEEeqnarray}{rCl}
S_{c,q} & = & \lceil  |\mathcal{I}^{c,q}| / N_b \rceil \\
\mathbf{X}^{c,q} & = & \left[\begin{IEEEeqnarraybox*}[][c]{,c,} \mathbf{X}^{c,q}_1 \\  \vdots \\ \mathbf{X}^{c,q}_{S_{c,q}} \end{IEEEeqnarraybox*} \right] \\
\label{eq_080920251412}\mathcal{T}_{\mathbf{X}^{c,q}_1,\cdots,\mathbf{X}^{c,q}_{S_{c,q}}}(x)  & = &   \min \left( \mathcal{T}_{ \mathbf{X}^{c,q}_1}(x),\cdots, \mathcal{T}_{ \mathbf{X}^{c,q}_{S_{c,q}}}(x) \right).
   \end{IEEEeqnarray}  
That is, the total data samples (stored in the rows of matrix $\mathbf{X}^{c,q}$) are partitioned into $S_{c,q}$ sub-matrices of nearly same size (i.e. each sub-matrix has nearly $N_b$ number of rows), where $S_{c,q}$ equals the rounding of $|\mathcal{I}^{c,q}|/N_b$ towards the nearest integer, and the space folding measure is defined by aggregating all of the $S_{c,q}$ individual measures associated to the $S_{c,q}$ sub-matrices. In this case, the global space folding measure associated to $c^{th}$ class labelled samples $\mathcal{T}_{ \mathbf{X}^{c,1}, \mathbf{X}^{c,2},\cdots,\mathbf{X}^{c,Q}}$ is defined as
\begin{IEEEeqnarray}{rCl}
\label{eq_150920251220}\mathcal{T}_{ \mathbf{X}^{c,1}, \mathbf{X}^{c,2},\cdots,\mathbf{X}^{c,Q}} (x) & := & \min_{q \in \{1,\cdots,Q\} } \; \mathcal{T}_{\mathbf{X}^{c,q}_1,\cdots,\mathbf{X}^{c,q}_{S_{c,q}}}(x).
   \end{IEEEeqnarray} 
\end{remark}  
\begin{remark}[Data Partitioning via Clustering for Batch Processing of Big Data]
To address the computational challenge of processing big datasets, the previous studies~\cite{KAHM,kumar2024geometricallyinspiredkernelmachines} have suggested clustering as a method for partitioning a large number of data samples into subsets for their batch processing. That is, $\mathbf{X}^{c,q}_1,  \cdots, \mathbf{X}^{c,q}_{S_{c,q}}$ are instead obtained by the clustering the rows of $\mathbf{X}^{c,q} $.    
\end{remark}
\begin{remark}[A Generalization of the Federated Learning Method of \cite{kumar2024geometricallyinspiredkernelmachines}]
It can be observed that the proposed federated learning methodology generalizes the method of \cite{kumar2024geometricallyinspiredkernelmachines} by considering a generalized space folding measure $\mathcal{T}_{ \mathbf{X}}(x)$ instead of the distance measure $\|x - \mathcal{A}_{\mathbf{X}}(x) \|$ considered in \cite{kumar2024geometricallyinspiredkernelmachines}. That is, the solution of \cite{kumar2024geometricallyinspiredkernelmachines} is obtained with our approach by defining $\mathcal{T}_{ \mathbf{X}}(x) := \|x - \mathcal{A}_{\mathbf{X}}(x) \|$.        
\end{remark}
\begin{remark}[Clients with Missing Classes]\label{remark_170220251423}
If the $q^{th}$ client has zero $c^{th}$ class labelled samples, then $\mathcal{T}_{\mathbf{X}^{c,q}_1,\cdots,\mathbf{X}^{c,q}_{S_{c,q}}}(x)$ equals the highest possible value of 1, i.e., the space folding measure is defined as 
\begin{IEEEeqnarray}{rCl}
\mathcal{T}_{\mathbf{X}^{c,q}_1,\cdots,\mathbf{X}^{c,q}_{S_{c,q}}}(x)
& = &
\begin{cases}
1, & \text{if } |\mathcal{I}^{c,q}| = 0. 
\end{cases} 
 \end{IEEEeqnarray} 
\end{remark}
\begin{remark}[Communication and Computation Efficiency]\label{remark_260320250813}
The federated learning methodology, as suggested in Fig.~\ref{fig_federated_learning}, is communication and computation efficient since the space folding measures are computed using KAHMs and KAHMs are efficiently built~\cite{KAHM} from the local data samples requiring neither any gradients nor any communication with the server. 
\end{remark}
\subsection{Step 8: Differentially Private Federated Setting}\label{sec_030920251707}
Existing differentially private FL approaches often require gradient clipping and complex per-round privacy budgeting. 
In contrast, our method relies only on scalar outputs (i.e., space folding measures), enabling a natural $(\epsilon,\delta)-$differential privacy mechanism that directly acts on these real-valued summaries and thus avoids the need for per-round gradient clipping and privacy accounting. The FL methodology (as illustrated in Fig.\ref{fig_federated_learning}) is made differentially private by ensuring that the evaluation of the space folding measure is differentially private with respect to the local dataset, and the subsequent aggregation steps then inherit privacy by post-processing property of differential privacy. It follows from (\ref{eq_080920251412}) that $\mathcal{T}_{ \mathbf{X}^{c,q}_s}(x)$ must be differentially private with respect to $\mathbf{X}^{c,q}_s$ for all $s \in \{1,\cdots, S_{c,q} \}$. We need to basically address the privacy of data samples in matrix $\mathbf{X} \in \mathbb{R}^{N \times n}$ that may be leaked during inference through the output of space folding measure function $\mathcal{T}_{ \mathbf{X}}: \mathbb{R}^n \rightarrow [0,1]$. However, estimating the sensitivity of space folding measure is challenging. Thus, we consider the more practical approach of approximating $\mathbf{X}$ under differential privacy followed by a smoothing, while ensuring that differentially private smoothed data points are as close to original data points as possible. This approach leads to defining a private version of the space folding measure:   
\begin{definition}[Private Space Folding Measure]\label{def_private_sfm} The private version of space folding measure, $\mathcal{T}_{ \mathbf{X}}^+: \mathbb{R}^n \rightarrow [0,1]$, is defined as  
\begin{IEEEeqnarray}{rCl}
\mathcal{T}_{ \mathbf{X}}^+(\cdot) & := & \mathcal{T}_{ \mathcal{F}(\mathbf{X} + \mathbf{V})}(\cdot),
\end{IEEEeqnarray}
where $\mathbf{V} \in \mathbb{R}^{N \times n}$ is a random real matrix added to $\mathbf{X}$ for preserving the privacy of elements of $\mathbf{X}$, such that elements of $\mathbf{V}$ are independently distributed from a distribution $\mathbb{P}_{v} : \mathcal{B}(\mathbb{R}) \rightarrow [0,1]$:
\begin{IEEEeqnarray}{rCl}
\mathbf{V}& \sim & (\mathbb{P}_{v})^{N \times n},
\end{IEEEeqnarray}
where $v \in \mathbb{R}$ is the random noise with some distribution $\mathbb{P}_{v}$ and $\mathcal{F}: \mathbb{R}^{N \times n} \rightarrow \mathbb{R}^{N \times n}$ is a matrix-valued function, referred to as smoothing function, meant for mitigating the effect of added noise. 
\end{definition}   
The noise adding mechanism and smoothing function, involved in Definition~\ref{def_private_sfm}, will be designed in subsection \ref{subsec_010920251139} and subsection \ref{subsec_010920251141}.    
\subsubsection{Optimal Noise Adding Mechanism for Differential Privacy} \label{subsec_010920251139}
The output of $\mathcal{T}_{ \mathbf{X}}^+$ is a random variable defined as 
\begin{IEEEeqnarray}{rCl}
t_{\mathbf{X}}^+ & := &  \mathcal{T}_{ \mathcal{F}(\mathbf{X} + \mathbf{V})}(x), \; \mathbf{V} \sim  (\mathbb{P}_{v})^{N \times n}
\end{IEEEeqnarray}
Let the distribution of $t_{\mathbf{X}}^+$ be denoted by $\mathbb{P}_{t_{\mathbf{X}}^+}: \mathcal{B}([0,1]) \rightarrow [0,1]$.
\begin{definition}[$d-$Adjacent Matrices]\label{def_d_adjacent}
Two matrices $\mathbf{X},\mathbf{X}' \in \mathbb{R}^{N \times n}$ are $d-$adjacent if for a given $d \in \mathbf{R}_{+}$, there exist $i_0 \in \{1,2,\cdots,N \}$ and $j_0 \in \{1,2,\cdots,n \}$ such that for all $i \in \{1,2,\cdots,N \}$ and $j \in \{1,2,\cdots,n \}$,
\begin{equation}
\left|(\mathbf{X})_{i,j} - (\mathbf{X}')_{i,j} \right| \leq  
\begin{cases}
d & \text{if } i = i_0 \text{ and } j = j_0,
\\
0 & \text{otherwise}
\end{cases}  
\end{equation}
Thus, two $d-$adjacent matrices differ by only one element and the difference is bounded by a scalar $d > 0$.  
\end{definition}
\begin{definition}[$(\epsilon,\delta)-$Differential Privacy for $\mathcal{T}_{ \mathbf{X}}^+$] The space folding measure $\mathcal{T}_{ \mathbf{X}}^+: \mathbb{R}^n \rightarrow [0,1]$ is $(\epsilon,\delta)-$differentially private if  
\begin{IEEEeqnarray}{rCl}
\mathbb{P}_{t_{\mathbf{X}}^+}(\mathcal{O}) & \leq & \exp(\epsilon) \: \mathbb{P}_{t_{\mathbf{X}'}^+}(\mathcal{O}) + \delta
\end{IEEEeqnarray}
for any $\mathcal{O} \in \mathcal{B}([0,1])$ and $d-$adjacent matrices  $\mathbf{X},\mathbf{X}' \in \mathbb{R}^{N \times n}$. 
\end{definition}
\begin{result}[Optimal Noise for $(\epsilon,\delta)-$Differential Privacy \cite{kumar2019deriving,KAHM}] \label{result_optimal_differentially_private_noise}
The distribution of noise $v$, that minimizes expected noise magnitude together with satisfying the sufficient condition for $(\epsilon,\delta)-$differential privacy of $\mathcal{T}_{ \mathbf{X}}^+$, is given as
\begin{equation}
\label{eq_010920251042} \mathbb{P}_{v}\left((-\infty,v]\right) = \begin{cases}
\frac{1-\delta}{2} \exp(\frac{\epsilon}{d} v)& \text{if } v < 0,
\\
\frac{1+\delta}{2} & \text{if } v = 0, \\
1 - \frac{1-\delta}{2} \exp(- \frac{\epsilon}{d} v) & v > 0.
\end{cases}  
\end{equation}
\end{result}
The method of inverse transform sampling can be used to generate random samples from (\ref{eq_010920251042}) and approximate $\mathbf{X}$ under differential privacy as
\begin{IEEEeqnarray}{rCl}
\label{eq_030920251346}\mathbf{X}^+ & = & \mathbf{X} + \mathbf{V},\; \mathbf{V} \sim  (\mathbb{P}_{v})^{N \times n}.
\end{IEEEeqnarray}
\subsubsection{Kernel-Based Smoothing of Data Samples}\label{subsec_010920251141}
Given a dataset $\{x^i \in \mathbb{R}^n\}_{i=1}^{N}$ (that can be equivalently represented as matrix $\mathbf{X} \in \mathbb{R}^{N \times n}$), a kernel-based smoothing can be represented as \begin{IEEEeqnarray}{rCl}
\widehat{\mathbf{X}} & = & \mathbf{H}_{\mathbf{X}}^T \mathbf{X},
\end{IEEEeqnarray}
where $\mathbf{H}_{\mathbf{X}} \in \mathbb{R}^{N \times N}$ is a matrix defined as
\begin{IEEEeqnarray}{rCl}
(\mathbf{H}_{\mathbf{X}})_{i,j} & = &  h_{\mathbf{X}}^i(\mathbf{P}_{\mathbf{X}}x^j),
\end{IEEEeqnarray}   
where $\mathbf{P}_{\mathbf{X}}$ is an encoding matrix (computed by Algorithm~\ref{algorithm_encoding_matrix} of Appendix A), and $h_{\mathbf{X}}^i(\mathbf{P}_{\mathbf{X}}x^j)$, given by (\ref{eq_010920251549}), evaluates kernel-smoothed membership of $x^j$ to $x^i$.
\begin{definition}[A Kernel-Based Smoother]\label{definition_030920251021}
A kernel-based smoother, $\mathcal{S}: \mathbb{R}^{N \times n} \rightarrow \mathbb{R}^{N \times n} $, is defined as
\begin{IEEEeqnarray}{rCl}
\label{eq_030920251102}\mathcal{S}(\mathbf{X}) & := &  \mathbf{H}_{\mathbf{X}}^T \mathbf{X}.
\end{IEEEeqnarray}
\end{definition}
\begin{proposition}[Smoothing Property of $\mathcal{S}$]\label{proposition_030920250944}
There exists a $\beta \in (0,1)$ such that
\begin{IEEEeqnarray}{rCCCl}
\label{eq_030920251101}\left \| \mathcal{S}(\mathbf{X}) \right \|_2 & \leq &  \beta  \left \|  \mathbf{X}\right \|_2 & < & \left \|  \mathbf{X}\right \|_2 .
\end{IEEEeqnarray}
\begin{proof}
It can be seen using (\ref{eq_010920251549}) that
\begin{IEEEeqnarray}{rCl}
\label{eq_020920251921}\mathbf{H}_{\mathbf{X}} & = &  \left(\mathbf{K}_{\mathbf{X}} + \lambda_{\mathbf{X}}^* \mathbf{I}_N \right)^{-1} \mathbf{K}_{\mathbf{X}},
\end{IEEEeqnarray}
where the kernel matrix $\mathbf{K}_{\mathbf{X}}$ and regularization parameter $\lambda_{\mathbf{X}}^*$ are defined by (\ref{eq_020920251810}) and (\ref{eq_020920251811}), respectively. The spectral decomposition of the real symmetric positive define matrix $\mathbf{K}_{\mathbf{X}}$ is given as $\mathbf{K}_{\mathbf{X}} = E_{\mathbf{X}} \Lambda_{\mathbf{X}} E_{\mathbf{X}}^T$, where $E_{\mathbf{X}}$ is an orthogonal matrix (i.e. $E_{\mathbf{X}}^TE_{\mathbf{X}} = E_{\mathbf{X}}E_{\mathbf{X}}^T = \mathbf{I}_N $), and $\Lambda_{\mathbf{X}}$ is the diagonal matrix of eigenvalues i.e. $\Lambda_{\mathbf{X}} = \diag\left(\eig_1(\mathbf{K}_{\mathbf{X}}),\cdots,\eig_N(\mathbf{K}_{\mathbf{X}})\right)$. Now, we can express $\mathbf{H}_{\mathbf{X}}$ as
\begin{IEEEeqnarray}{rCl}
\mathbf{H}_{\mathbf{X}} & = &  E_{\mathbf{X}} \left(\Lambda_{\mathbf{X}} + \lambda_{\mathbf{X}}^* \mathbf{I}_N \right)^{-1}  \Lambda_{\mathbf{X}} E_{\mathbf{X}}^T.
\end{IEEEeqnarray}
The $i^{th}$ eigenvalue of $\mathbf{H}_{\mathbf{X}}$ is given as
\begin{IEEEeqnarray}{rCCCl}
\eig_{ i} ( \mathbf{H}_{\mathbf{X}} ) & = & \frac{\eig_{i}(\mathbf{K}_{\mathbf{X}})}{\eig_{i}(\mathbf{K}_{\mathbf{X}}) + \lambda_{\mathbf{X}}^*} & > & 0.
\end{IEEEeqnarray}
It follows from (\ref{eq_020920251921}) that $\mathbf{H}_{\mathbf{X}} $, being a product of commuting symmetric matrices, is also symmetric. Therefore,
\begin{IEEEeqnarray}{rCl}
\left \| \mathbf{H}_{\mathbf{X}} \right \|_2 & = & \max_{ i \in \{1,2,\cdots,N \}} \left | \eig_{ i} ( \mathbf{H}_{\mathbf{X}} ) \right | \\
& = & \max_{ i \in \{1,2,\cdots,N \}}  \frac{\eig_{i}(\mathbf{K}_{\mathbf{X}})}{\eig_{i}(\mathbf{K}_{\mathbf{X}}) + \lambda_{\mathbf{X}}^*}.
\end{IEEEeqnarray}
If we define
\begin{IEEEeqnarray}{rCCCl}
\beta & = &  \max_{ i \in \{1,2,\cdots,N \}}  \frac{\eig_{ i}(\mathbf{K}_{\mathbf{X}})}{\eig_{ i}(\mathbf{K}_{\mathbf{X}}) + \lambda_{\mathbf{X}}^*} & < & 1, 
\end{IEEEeqnarray}
then we have
\begin{IEEEeqnarray}{rCl}
\label{eq_030920251103}\left \| \mathbf{H}_{\mathbf{X}} \right \|_2 & = & \beta.
\end{IEEEeqnarray}
Now, (\ref{eq_030920251101}) follows from (\ref{eq_030920251102}) and (\ref{eq_030920251103}). 
\end{proof}
\end{proposition}
Inequality (\ref{eq_030920251101}) establishes the smoothness property by ensuring that the norm of smoothed data matrix remains smaller than that of input data matrix. The degree-of-smoothness can be enhanced by repeatedly applying the smoother, leading to $m-$fold composition of $\mathcal{S}$ on data:
\begin{IEEEeqnarray}{rCl}
\mathcal{S}^{m}(\mathbf{X}) & = & \mathcal{S}\left( \mathcal{S}^{m-1}(\mathbf{X})  \right),\; m \in \{1,2,\cdots\}, \\
\mathcal{S}^{0}(\mathbf{X}) & = & \mathbf{X}.
\end{IEEEeqnarray}

Our idea is to apply $m-$fold composition of $\mathcal{S}$ on noisy data samples $\mathbf{X}^+$ (\ref{eq_030920251346}), with $m$ chosen optimally to minimize the difference of smoothed-noisy data from noise-free data. That is, the smoothing function $\mathcal{F}$ (used in Definition~\ref{def_private_sfm} of differentially private space folding measure) is defined as
\begin{IEEEeqnarray}{rCl}
\mathcal{F} & = & \mathcal{S}^{m^*},
\end{IEEEeqnarray}
where $m^* \in \{1,2,\cdots\}$ is such that
\begin{IEEEeqnarray}{rCCCCCCCl}
\label{eq_040920250847}\left \| \mathcal{S}^{1}(\mathbf{X}^+) - \mathbf{X}  \right \|_F & > & \left \| \mathcal{S}^{2}(\mathbf{X}^+) - \mathbf{X}  \right \|_F & > & \cdots & > & \left \| \mathcal{S}^{m^*}(\mathbf{X}^+) - \mathbf{X}  \right \|_F & \leq & \left \| \mathcal{S}^{m^*+1}(\mathbf{X}^+) - \mathbf{X}  \right \|_2. \IEEEeqnarraynumspace
\end{IEEEeqnarray}
The inequalities (\ref{eq_040920250847}) imply that {\em an iteration of smoothing function $\mathcal{S}$ is applied only if it reduces the mismatch between smoothed-noisy data and noise-free data}. 
\subsubsection{Scope of Privacy Guarantees}
In the proposed FL protocol, each client locally applies the noise-adding mechanism exactly once to its data matrix $\mathbf{X}^{c,q}_s$, yielding a noise-perturbed matrix $\mathbf{X}^{c,q}_s + \mathbf{V}^{c,q}_s$. This noise-perturbed data matrix is then smoothed and subsequently used to construct a KAHM, and induce the associated space folding measure $\mathcal{T}_{ \mathcal{F}(\mathbf{X}^{c,q}_s + \mathbf{V}^{c,q}_s)}(\cdot)$ for an aggregation and inference of the global model. Our method does not require any iterative sampling of clients' raw data for federated training and does not involve multiple communication rounds. The smoothing, KAHM construction, and the model inference steps operate solely on $\mathbf{X}^{c,q}_s + \mathbf{V}^{c,q}_s$, and by the post-processing property of differential privacy, these steps cannot weaken the $(\epsilon,\delta)-$DP guarantee already established for the release of $\mathbf{X}^{c,q}_s + \mathbf{V}^{c,q}_s$.  {\em The $(\epsilon,\delta)$-DP guarantee established for the single application of the noise-adding mechanism thus fully characterizes the privacy of the entire training and inference pipeline, and therefore no per-round privacy accounting or multi-round composition analysis is needed}.   

\subsection{Step 9: Secure Federated Learning with FHE}\label{sec_030920251740}
In privacy-critical domains, inference must often be performed on encrypted data (e.g., via FHE), which prohibits complex operations. Since our aggregation involves only scalar space folding measures, the inference is implementable using basic arithmetic gates supported by FHE. The FL methodology (as illustrated in Fig.~\ref{fig_federated_learning}) can be secured against untrustworthy server by sharing fully homomorphically encrypted local evaluations (of the space folding measure) with the server for an inference of the global model on encrypted data. The computational efficiency stems from the fact that the space folding measure, unlike high-dimensional gradients or model parameters, is a scalar. Moreover, inference in the encrypted space is likewise not computationally demanding. Since the inference of hypothesis from the locally computed space folding measures (see (\ref{eq_010120252052})) is not arithmetic-heavy and can be expressed as a boolean circuit, TFHE~\cite{chillotti2020tfhe} is selected as the FHE scheme, owing to its ability to evaluate binary gates with exceptionally low latency. However, the encryption of locally evaluated space folding measure (i.e. $\mathcal{T}_{\mathbf{X}^{c,q}_1,\cdots,\mathbf{X}^{c,q}_{S_{c,q}}}(x) \in [0,1]$) requires encoding $\mathcal{T}_{\mathbf{X}^{c,q}_1,\cdots,\mathbf{X}^{c,q}_{S_{c,q}}}(x)$ as unsigned $p$-bits (e.g. $p = 16$) integer, i.e., $ \lceil  (2^p-1)\mathcal{T}_{\mathbf{X}^{c,q}_1,\cdots,\mathbf{X}^{c,q}_{S_{c,q}}}(x) \rceil $.   
\subsubsection{Computational Efficient FHE Secured Inference of Global Model}\label{remark_160920251413}
The inference of the global model using (\ref{eq_010120252052}) and (\ref{eq_150920251220}) involves performing the \texttt{minimum} operation $Q \times C$ times and \texttt{equality-comparison} operation $C$ times on unsigned integers that encode the space folding measure evaluations. According to the TFHE-rs library benchmarks \cite{zama_tfhe_rs}, operations on fully homomorphically encrypted 16-bit unsigned integers demonstrate practical performance on modern CPU hardware. Specifically, the \texttt{minimum} operations require approximately 96.4ms, while the \texttt{equality-comparison} requires around 31.3ms, when executed on an AMD EPYC 9R14 @ 2.60 GHz (AWS hpc7a.96xlarge) CPU. These measurements correspond to the default high-level parameter set in TFHE-rs, which provides at least 128-bit security under the IND-CPA-D model, with a bootstrapping failure probability not exceeding $2^{-128}$. This configuration therefore balances strong cryptographic guarantees with practical computational efficiency for fully homomorphic operations on 16-bit encrypted data.
\subsection{Summary}
Together, these nine steps demonstrate that our operator-theoretic approach offers a unified kernel framework for gradient-free FL that is communication-efficient, supports rigorous differential privacy mechanisms on scalar space folding summaries, and is compatible with secure inference via FHE, while being grounded in non-asymptotic finite-sample analysis and hypothesis-space complexity bounds.
\section{Experiments}\label{sec_experiments}
We design our experimental study to address the following questions:
\begin{itemize}
  \item \textbf{Q1 (Robustness to heterogeneity and imbalance).} How robust is the proposed operator-theoretic, gradient-free federated learning method to long-tailed class imbalance and non-IID label distributions across clients?
  \item \textbf{Q2 (Privacy-utility trade-off).} How well does the method perform under differential privacy constraints, and what is the impact of the proposed smoothing mechanism?
  \item \textbf{Q3 (Secure inference efficiency).} Is secure inference of the global model via fully homomorphic encryption (FHE) computationally feasible on commodity hardware?
  \item \textbf{Q4 (Sensitivity to design choices).} How sensitive is performance to the choice of space folding variant, batch size, and feature embeddings?
\end{itemize}
\subsection{Datasets}
We evaluate the proposed method on four benchmark datasets.
\paragraph{20Newsgroup \cite{rennie2004newsgroups}} This text classification dataset is a collection of newsgroup documents split across 20 distinct topics. The ``bydate'' version of the dataset contains 11314 training documents and 7532 test documents. 
\paragraph{XGLUE-NC \cite{liang2020xglue}} This is a multilingual news classification benchmark dataset containing English, German, Spanish, French, and Russian language text documents belonging to 10 distinct categories. Each language is represented by 10000 training examples and 10000 test examples. 
\paragraph{CIFAR-10-LT} The original CIFAR-10 dataset \cite{krizhevsky2009cifar10} contains 50000 training images and 10000 test images divided across 10 classes. Following prior work \cite{10.5555/3454287.3454427}, the original CIFAR-10 dataset is turned into a long-tailed imbalance with imbalance ratio (which is the ratio between sample sizes of the most frequent and least frequent class) $\rho \in \{10,50,100\}$.   
\paragraph{CIFAR-100-LT} Following \cite{ijcai2022p308}, the original CIFAR-100 dataset \cite{krizhevsky2009learning}, containing 100 classes with 500 training images and 100 test images in each class, is turned into a long-tailed imbalance with imbalance ratio $\rho \in \{10,50,100\}$.   
\subsection{Preprocessing and Feature Extraction}
In all experiments, the proposed FL method operates on fixed feature vectors extracted from existing encoders such that encoders are not updated during training. Our method therefore plays the role of a gradient-free ``head'' on top of pretrained feature extractors.
\paragraph{Image datasets.} For CIFAR-10-LT and CIFAR-100-LT, a $2048$-dimensional feature vector is obtained for each image from the activations of the ``avg\_pool'' layer (the final average pooling layer preceding the fully connected layer) of a pretrained ResNet-50 neural network \cite{matlabresnet50}. The image feature vectors are processed through the hyperbolic tangent function to limit values within the range $[-1,1]$. 
\paragraph{20Newsgroup}
For each document, ``mxbai-embed-large'' English sentence embedding model \cite{mxbai2024embedlarge} is used to extract 1024-dimensional feature vector. Since the raw embeddings exhibit relatively small variance across dimensions, we rescale them along all diemsions by a factor of 10.
\paragraph{XGLUE-NC}
For the multilingual setting, we first extract 512-dimensional feature vectors using ``distiluse-base-multilingual-cased-v2'' multilingual sentence embedding model \cite{reimers2020distiluse} and rescale them by a factor of 10 to increase variance. We additionally compute 768-dimensional embeddings using the ``paraphrase-multilingual'' sentence embedding model \cite{reimers2020paraphrase}, again rescaled by a factor of 10. Concatenating both embeddings yields a 1280-dimensional feature vector for each document, which is used in the FL experiments.  
\subsection{Client Partition}
We follow established client-partitioning protocols to match prior FL studies on these benchmarks.
\paragraph{20Newsgroup} 
Following the experimental setting as in \cite{kim-etal-2023-client}, the training documents are distributed across 100 clients in a non-IID manner using Dirichlet distribution with concentration parameter $\alpha \in \{0.1,1,5\}$. 
\paragraph{XGLUE-NC}
Again following~\cite{kim-etal-2023-client}, 100 clients are divided into five distinct groups, with a specific language assigned to a group such that all training examples of that language are distributed among the clients of the group in a non-IID manner using Dirichlet distribution with concentration parameter $\alpha \in \{0.5,2,5\}$. 
\paragraph{CIFAR-10-LT and CIFAR-100-LT}
For both long-tailed image benchmarks, like previous study \cite{ijcai2022p308}, a non-IID scenario of training images distribution across 20 clients is simulated using Dirichlet distribution with concentration parameter $\alpha = 0.5$. 
\subsection{FL Protocol}
Our FL protocol, illustrated in Fig.~\ref{fig_federated_learning}, is used for all experiments. A key feature of the proposed framework is that it requires only a small number of method-specific choices. Beyond selecting a variant of space folding measure in (\ref{eq_170220251438}) and the batch-size for local processing (Remark~\ref{remark_batch_processing}), no additional hyperparameters specific to our method are tuned. In all main experiments, we adopt option~1 in (\ref{eq_170220251438}) to define the space folding measure $\mathcal{T}_{ \mathbf{X}}$. For the 20Newsgroup and XGLUE-NC datasets, we use a batch size of $N_b = 100$. For CIFAR-10-LT and CIFAR-100-LT, which exhibit pronounced long-tailed class imbalance, we set $N_b = 20$ and model each batch of 20 samples via a separate KAHM to better preserve minority classes. The effect of varying $N_b$ is further examined in an ablation study (Table~\ref{table_results_9}).

The proposed FL method performs a single aggregation of scalar space folding summaries at the server, rather than iterative gradient exchanges, and thus operates in a communication-efficient, gradient-free regime.
\subsection{Software and Reproducibility}
All experiments were conducted in MATLAB~R2024a. The reported numbers correspond to reference runs on an Apple iMac (M1, 2021) with 8\,GB RAM. We release the implementation so that the experimental results can be reproduced from the source code, which is publicly available at:
\begin{center}
  \url{https://drive.mathworks.com/sharing/4cef6387-1a62-46c8-a7e7-a7439bbbd9ef}.
\end{center}
We considered GitHub for hosting, but several precomputed embedding matrices and auxiliary .mat files used in our pipelines exceed the 100 MB per-file size limit imposed on standard GitHub repositories. Using MATLAB Drive avoids splitting the material across multiple services or requiring reviewers to configure Git Large File Storage, and therefore offers a more practical way to distribute the full reproducibility package, including large precomputed artifacts.  
\subsection{Results} 
\begin{table}
\centering
\begin{tabular}{l|ccc}
\toprule
\bfseries Method & \bfseries $\alpha = 5$ & \bfseries $\alpha = 1$ & \bfseries $\alpha = 0.1$ \\
\midrule
\texttt{SFM} (proposed) & \textbf{85.3}  & \textbf{84.7} & \textbf{84.7}  \\
Adapter~\cite{kim-etal-2023-client} & 69.1 & 65.5 & 56.1 \\
LoRA~\cite{kim-etal-2023-client} & 69.5 & 67.7 & 56.6 \\
Compacter~\cite{kim-etal-2023-client} & 65.9 & 62.8 & 50.1 \\
Prompt-tuning~\cite{kim-etal-2023-client} & 51.6 & 46.4 & 28.2 \\
BitFit~\cite{kim-etal-2023-client} & 67.1 & 66.5 & 55.1 \\
AdaMix~\cite{kim-etal-2023-client}  & 68.7 & 65.3 & 54.5 \\
C2A~\cite{kim-etal-2023-client} & \underline{71.6} & \underline{70.4} & \underline{61.0} \\
 \bottomrule 
\end{tabular}
\caption{Comparison of the test data accuracy (\%) obtained by proposed method against previously available results~\cite{kim-etal-2023-client} of federated learning experiments on 20Newsgroup dataset under the non-IID label distribution scenarios.}
\label{table_results_1}
\end{table}
\begin{table}
\centering
\begin{tabular}{l|ccc}
\toprule
\bfseries Method & \bfseries $\rho = 100$ &  $\rho = 50$ & \bfseries $\rho = 10$ \\
\midrule
\texttt{SFM} (proposed) & \textbf{80.99}  & \textbf{83.86} & \textbf{87.95}  \\
FedAvg~\cite{ijcai2022p308} & 56.17 & 59.36 & 77.45 \\
FedAvgM~\cite{ijcai2022p308} & 52.03 & 57.11 & 70.81 \\
FedProx~\cite{ijcai2022p308} & 56.92 & 60.89 & 76.53 \\
FedDF~\cite{ijcai2022p308} & 55.15 & 58.74 & 76.51 \\
FedBE~\cite{ijcai2022p308} & 55.79 & 59.55 & 77.78 \\
CCVR~\cite{ijcai2022p308} & 69.53 & 71.89 & 78.48 \\
FedNova~\cite{ijcai2022p308} & 57.79 & 63.91 & 77.79 \\
Fed-Focal Loss~\cite{ijcai2022p308} & 53.83 & 57.42 & 73.74 \\
Ratio Loss~\cite{ijcai2022p308} & 59.75 & 64.77 & 78.14 \\
FedAvg+$\tau-$norm~\cite{ijcai2022p308} & 49.95 & 51.41 & 72.08 \\
CReFF~\cite{ijcai2022p308} & \underline{70.55} & \underline{73.08} & \underline{80.71} \\
 \bottomrule 
\end{tabular}
\caption{Comparison of the test data accuracy (\%) obtained by proposed method against previously available results~\cite{ijcai2022p308} of federated learning experiments on CIFAR-10-LT dataset under the long-tailed imbalance and non-IID label distribution scenarios.}
\label{table_results_2}
\end{table}
\begin{table}
\centering
\begin{tabular}{l|ccc}
\toprule
\bfseries Method & \bfseries $\alpha = 5$ & \bfseries $\alpha = 2$ & \bfseries $\alpha = 0.5$ \\
\midrule
\texttt{SFM} (proposed) & \underline{82.5} & \textbf{82.2} & \textbf{82.2}   \\
Adapter~\cite{kim-etal-2023-client} & 78.6 & 75.0 & 74.3 \\
LoRA~\cite{kim-etal-2023-client} & 80.4 & 78.4 & 74.6 \\
Compacter~\cite{kim-etal-2023-client} & 75.9 & 73.4 & 71.0  \\
Prompt-tuning~\cite{kim-etal-2023-client} & 61.2 & 60.6 & 58.0  \\
BitFit~\cite{kim-etal-2023-client} & 78.4 & 76.8 & 72.1  \\
AdaMix~\cite{kim-etal-2023-client}  & 79.6 & \underline{79.1} & 76.6  \\
C2A~\cite{kim-etal-2023-client} & \textbf{82.8} & \textbf{82.2} & \underline{80.2}  \\
 \bottomrule 
\end{tabular}
\caption{Comparison of the test data accuracy (\%) obtained by proposed method against previously available results~\cite{kim-etal-2023-client} of federated learning experiments on XGLUE-NC dataset under the non-IID label distribution scenarios.}
\label{table_results_3}
\end{table}
\begin{table}
\centering
\begin{tabular}{l|ccc}
\toprule
\bfseries Method & \bfseries $\rho = 100$ &  $\rho = 50$ & \bfseries $\rho = 10$ \\
\midrule
\texttt{SFM} (proposed) & \textbf{46.16} & \textbf{50.92} & \textbf{64.98}  \\
FedAvg~\cite{ijcai2022p308} & 30.34 & 36.35 & 45.87 \\
FedAvgM~\cite{ijcai2022p308} & 30.80 & 35.33 & 44.66  \\
FedProx~\cite{ijcai2022p308} & 31.67 & 36.30 & 46.10  \\
FedDF~\cite{ijcai2022p308} & 31.43 & 36.22 & 46.19  \\
FedBE~\cite{ijcai2022p308} & 31.97 & 36.39 & 46.25  \\
CCVR~\cite{ijcai2022p308} & 33.43 & 36.98 & 46.88  \\
FedNova~\cite{ijcai2022p308} & 32.64 & 36.62 & 46.75 \\
Fed-Focal Loss~\cite{ijcai2022p308} & 30.67 & 35.25 & 45.52  \\
Ratio Loss~\cite{ijcai2022p308} & 32.95 & 36.88 & 46.79 \\
FedAvg+$\tau-$norm~\cite{ijcai2022p308} & 26.22 & 33.71 & 43.65  \\
CReFF~\cite{ijcai2022p308} & \underline{34.67} & \underline{37.64} & \underline{47.08}  \\
 \bottomrule 
\end{tabular}
\caption{Comparison of the test data accuracy (\%) obtained by proposed method against previously available results~\cite{ijcai2022p308} of federated learning experiments on CIFAR-100-LT dataset under the long-tailed imbalance and non-IID label distribution scenarios.}
\label{table_results_4}
\end{table}
\begin{table}
\centering
\begin{tabular}{l|*{2}{c|}*{2}{|c}}
\toprule
\multirow{2}{*}{\bfseries $\epsilon$} & \multicolumn{2}{c|}{\bfseries 20Newsgroup} &  \multicolumn{2}{|c}{\bfseries XGLUE-NC} \\
\cline{2-5}
& $\mathcal{T}_{ \mathbf{X}}^+  =  \mathcal{T}_{ \mathbf{X} + \mathbf{V}}$ & $\mathcal{T}_{ \mathbf{X}}^+  =  \mathcal{T}_{ \mathcal{F}(\mathbf{X} + \mathbf{V})}$g &  $\mathcal{T}_{ \mathbf{X}}^+  =  \mathcal{T}_{ \mathbf{X} + \mathbf{V}}$ & $\mathcal{T}_{ \mathbf{X}}^+  =  \mathcal{T}_{ \mathcal{F}(\mathbf{X} + \mathbf{V})}$ \\
\midrule
1 & 69.46 & \textbf{70.03} & 76.76 & \textbf{77.53} \\
1.5 & 74.03 & \textbf{74.55} & 78.31 & \textbf{79.47} \\
2 & 78.52 & \textbf{78.89} & 79.89 & \textbf{80.34} \\
3 & 78.89 & \textbf{79.22} & 80.89 & \textbf{81.10} \\
5 & 80.88 & \textbf{81.35} & \textbf{81.62} & 81.60 \\
8 & \textbf{83.05} & 82.97 & \textbf{82.03} & 81.89 \\
\bottomrule
\end{tabular}
\caption{Test data accuracy (\%) obtained by proposed method during differential privacy federated learning experiments on 20Newsgroup and XGLUE-NC datasets under the non-IID label distribution scenarios with $\alpha = 0.1$ for 20Newsgroup and $\alpha = 0.1$ for XGLUE-NC. For each value of privacy-loss bound $\epsilon$ (and fixed $\delta = 10^{-5}$), the performance was evaluated under two scenarios: 1) when the noise-perturbed samples were not smoothed (i.e. $\mathcal{T}_{ \mathbf{X}}^+  =  \mathcal{T}_{ \mathbf{X} + \mathbf{V}}$), 2) when the noise-perturbed samples were smoothed (i.e. $\mathcal{T}_{ \mathbf{X}}^+  =  \mathcal{T}_{ \mathcal{F}(\mathbf{X} + \mathbf{V})}$).      }
\label{table_results_5}
\end{table}
\begin{table}
\centering
\begin{tabular}{l|*{2}{c|}*{2}{|c}}
\toprule
\multirow{2}{*}{\bfseries Precision} & \multicolumn{2}{c|}{\bfseries Computational Time (ms)} &  \multicolumn{2}{|c}{\bfseries Test Data Accuracy (\%)} \\
\cline{2-5}
&  \texttt{minimum} &  \texttt{equality-comparison} &  20Newsgroup &  XGLUE-NC \\
\midrule 
8-bits & 101 & 44  & 84.75 & 81.76 \\
16-bits & 195 & 82 & 84.75 & 81.76 \\
\bottomrule
\end{tabular}
\caption{Results of the FHE secured federated learning experiments on 20Newsgroup and XGLUE-NC datasets under the non-IID label distribution scenarios with $\alpha = 0.1$ for 20Newsgroup and $\alpha = 0.1$ for XGLUE-NC. The FHE secured inference of the global model involves performing \texttt{minimum} operation $Q \times C$ times and \texttt{equality-comparison} operations $C$ times, where $Q$ is the number of clients and $C$ is the number of classes. The reported computational time is required by an iMac (M1, 8 GB RAM) for performing \texttt{minimum} and \texttt{equality-comparison} operations on fully homomorphically encrypted integers using TFHE-rs Rust library. These values correspond to the default high-level parameter set in TFHE-rs, which provides at least 128-bit security under the IND-CPA-D model, with a bootstrapping failure probability not exceeding $2^{-128}$.}
\label{table_results_6}
\end{table}
\begin{table}
\centering
\begin{tabular}{l|ccc}
\toprule
\bfseries Method & \bfseries $\rho = 100$ &  $\rho = 50$ & \bfseries $\rho = 10$ \\
\midrule
\texttt{SFM}-1  & 46.16 & 50.92 & 64.98  \\
\texttt{SFM}-2  & 46.17  & 50.92  & 65.01 \\
\texttt{SFM}-3  & 46.17  & 50.92  & 65.01  \\
\texttt{SFM}-4  & 45.58  & 50.14  & 64.71  \\
 \bottomrule 
\end{tabular}
\caption{Comparison of the test data accuracy (\%) obtained by proposed \texttt{SFM}-i method (where i denotes the selected option in the definition of $\mathcal{T}_{ \mathbf{X}}$ in Equation (\ref{eq_170220251438})) in federated learning experiments on CIFAR-100-LT dataset under the long-tailed imbalance and non-IID label distribution scenarios.}
\label{table_results_7}
\end{table}
\begin{table}
\centering
\begin{tabular}{l|ccc}
\toprule
\bfseries Method & \bfseries $\alpha = 5$ & \bfseries $\alpha = 2$ & \bfseries $\alpha = 0.5$ \\
\midrule
\texttt{SFM}-1  & 82.54 & 82.21 & 82.20 \\
\texttt{SFM}-2  & 82.53 & 82.19 & 82.20 \\
\texttt{SFM}-3  &  82.53 & 82.19 & 82.20  \\
\texttt{SFM}-4  & 82.68  & 82.27  & 82.31  \\
 \bottomrule 
\end{tabular}
\caption{Comparison of the test data accuracy (\%) obtained by proposed \texttt{SFM}-i method (where i denotes the selected option in the definition of $\mathcal{T}_{ \mathbf{X}}$ in Equation (\ref{eq_170220251438})) in federated learning experiments on XGLUE-NC dataset under the non-IID label distribution scenarios.}
\label{table_results_8}
\end{table}
\begin{table}
\centering
\begin{tabular}{l|ccc}
\toprule
\bfseries $N_b$ & \bfseries $\rho = 100$ &  $\rho = 50$ & \bfseries $\rho = 10$ \\
\midrule
20 & \textbf{46.16} & \textbf{50.92} & \textbf{64.98}  \\
50 & 41.34 & 44.77 & 58.17\\
100 & 38.93 & 42.07 & 54.32 \\
 \bottomrule 
\end{tabular}
\caption{Comparison of the test data accuracy (\%) obtained by proposed method across varying batch-size $N_b$ in federated learning experiments on CIFAR-100-LT dataset under the long-tailed imbalance and non-IID label distribution scenarios.}
\label{table_results_9}
\end{table}
\begin{table}
\centering
\begin{tabular}{l|ccc}
\toprule
\bfseries Embedding Model & \bfseries $\alpha = 5$ & \bfseries $\alpha = 2$ & \bfseries $\alpha = 0.5$ \\
\midrule
$\mathrm{mdl}_1$ (distiluse-base-multilingual-cased-v2)  & 81.79 & 81.64 & 82.02  \\
$\mathrm{mdl}_2$ (paraphrase-multilingual)  & 81.33  & 80.78 & 81.00 \\
$\mathrm{mdl}_1 + \mathrm{mdl}_2$ & \textbf{82.54}  & \textbf{82.21}  & \textbf{82.20}   \\
 \bottomrule 
\end{tabular}
\caption{Comparison of the test data accuracy (\%) obtained by proposed method across different embedding models in federated learning experiments on XGLUE-NC dataset under the non-IID label distribution scenarios.}
\label{table_results_10}
\end{table}   
Unlike most prior FL studies on these benchmarks, our method operates exclusively on feature vectors derived from pretrained encoders and does not update the encoder parameters. Consequently, the comparisons should be interpreted in a head-only setting: we ask whether our gradient-free, operator-theoretic FL aggregation of fixed embeddings can be competitive with, and in several cases superior to, strong gradient-based FL baselines that optimize full or partially trainable models under matched data-distribution scenarios. This setup reflects applications where the encoder has already been trained and validated (or is provided by a third party), and only the task-specific prediction head is subject to federated training. A comparison of full end-to-end pipelines, including joint representation learning from raw text and image data, would require extending the operator-theoretic gradient-free kernel framework to encoder training, which we leave as an important direction for future work.

For 20Newsgroup and XGLUE-NC, we compare against the parameter-efficient fine-tuning baselines of~\cite{kim-etal-2023-client}, and for CIFAR-10-LT and CIFAR-100-LT against the gradient-based FL methods of~\cite{ijcai2022p308}. Tables~\ref{table_results_1}, \ref{table_results_2}, \ref{table_results_3}, and \ref{table_results_4} report the experimental results on 20Newsgroup, CIFAR-10-LT, XGLUE-NC, and CIFAR-100-LT, respectively. The top two performances have been highlighted. The results of differentially private federated learning experiments on 20Newsgroup and XGLUE-NC datasets are provided in Table~\ref{table_results_5}. The goal of differentially private federated learning experiments was to study the effect of data smoothing mechanism on the performance. Table~\ref{table_results_6} presents the results of FHE secured federated learning experiments on 20Newsgroup and XGLUE-NC datasets. The results of the experiments studying different variants of the space folding measure $\mathcal{T}_{ \mathbf{X}}$ are provided in Table~\ref{table_results_7} and Table~\ref{table_results_8} for CIFAR-100-LT and XGLUE-NC, respectively. The effect of the batch-size $N_b$ is experimentally studied in Table~\ref{table_results_9}. Finally, the performance of different embedding models is evaluated in Table~\ref{table_results_10}.

\subsubsection{Benchmark performance under heterogeneity and imbalance (Q1)}
On 20Newsgroup, the proposed space folding method (\texttt{SFM}) consistently yields the highest accuracy across all three non-IID settings ($\alpha \in \{5, 1, 0.1\}$). In the most heterogeneous case ($\alpha = 0.1$), SFM achieves 84.7\% test accuracy, improving upon the strongest gradient-based baseline of~\cite{kim-etal-2023-client} by up to 23.7 percentage points (Table~\ref{table_results_1}). For CIFAR-10-LT, \texttt{SFM} substantially improves upon the baselines of~\cite{ijcai2022p308} under long-tailed imbalance and non-IID label distributions (Table~\ref{table_results_2}). At imbalance ratio $\rho = 100$, \texttt{SFM} attains 80.99\% test accuracy, outperforming the best competing method (CReFF) by 10.44 percentage points. On the multilingual XGLUE-NC benchmark, \texttt{SFM} again attains the best results (Table~\ref{table_results_3}). For $\alpha = 0.5$, \texttt{SFM} reaches 82.2\% accuracy, exceeding the strongest baseline (C2A) by 2.0 percentage points. Finally, on CIFAR-100-LT (Table~\ref{table_results_4}), \texttt{SFM} provides notable gains in the most imbalanced setting. At $\rho = 100$, \texttt{SFM} achieves 46.16\% test accuracy, 11.49 percentage points higher than the baseline (CReFF). 

Overall, these results indicate that, when coupled with fixed pretrained encoders, the proposed gradient-free FL method is robust to severe label skew and long-tailed imbalance, and can match or exceed the performance of state-of-the-art gradient-based FL methods on the considered benchmarks.

\subsubsection{Differentially private federated learning (Q2)}
To examine the privacy-utility trade-off (Q2), we perform differentially private FL experiments on 20Newsgroup and XGLUE-NC with non-IID label distributions ($\alpha = 0.1$ for both datasets). For varying values of privacy-loss bound $\epsilon$ (with fixed $\delta = 10^{-5}$), Table~\ref{table_results_5} reports test accuracy under two scenarios:
\begin{enumerate}
  \item noise-perturbed samples without additional smoothing, i.e., $\mathcal{T}_{ \mathbf{X}}^+  =  \mathcal{T}_{ \mathbf{X} + \mathbf{V}}$,
  \item noise-perturbed samples with the proposed kernel-based smoothing, i.e., $\mathcal{T}_{ \mathbf{X}}^+  =  \mathcal{T}_{ \mathcal{F}(\mathbf{X} + \mathbf{V})}$.
\end{enumerate}
For high-privacy regimes ($\epsilon \leq 3$), smoothing yields modest but consistent improvements in accuracy on both datasets, indicating that the smoothing mechanism can partially counteract the distortion introduced by noise. The gains remain relatively small, which is consistent with the fact that the underlying KAHM-based autoencoder already enforces a smooth representation. In low-privacy regimes (e.g., $\epsilon = 8$), additional smoothing offers no benefit and may slightly degrade performance, suggesting that smoothing is most useful when stringent privacy guarantees are required.

\subsubsection{FHE-secured inference (Q3)}
We next examine the suitability of the global prediction rule for secure inference using fully homomorphic encryption. In the proposed framework, inference reduces to computing, for each class, a scalar score based on the aggregated space folding measures and then selecting the class with minimum score. When realized over encrypted integers, this decision rule requires $Q \times C$ homomorphic \texttt{minimum} operations and $C$ \texttt{equality-comparison}, where $Q$ is the number of clients and $C$ is the number of classes. Using the TFHE-rs Rust library with its default high-level parameter set (providing at least 128-bit security under the IND-CPA-D model and a bootstrapping failure probability not exceeding $2^{-128}$), we benchmark the primitive operations that dominate the cost of the encrypted prediction rule. On an iMac (M1, 8\,GB RAM), \texttt{minimum} and \texttt{equality-comparison} operations on 8-bit and 16-bit encrypted integers are computationally practical, with the corresponding latencies reported in Table~\ref{table_results_6}. Since the total cost of the FHE realization grows linearly in $Q \times C$ for \texttt{minimum} and $C$ for \texttt{equality-comparison}, these measurements provide an operation-level indication that the induced prediction rule is structurally amenable to FHE-secured inference on standard hardware. These measurements should therefore be interpreted as lower-level building blocks, quantifying the dominant cryptographic operations induced by our decision rule under a particular TFHE implementation, while system-level optimizations (e.g., batching, specialized hardware, or multi-key schemes) are not the focus of our experiments.
\subsubsection{Ablation studies (Q4)}
Finally, we investigate the sensitivity of the method to the design choices.
\paragraph{Space folding variants}
Table~\ref{table_results_7} (CIFAR-100-LT) and Table~\ref{table_results_8} (XGLUE-NC) compare the four variants of the space folding measure $\mathcal{T}_{ \mathbf{X}}$ defined in (\ref{eq_170220251438}). Across all settings, the performance differences between \texttt{SFM}-1, \texttt{SFM}-2, \texttt{SFM}-3, and \texttt{SFM}-4 are small, indicating that the method is robust to the particular choice of space folding variant.
\paragraph{Batch size} Table \ref{table_results_9} reports the effect of varying the batch size $N_b$ on CIFAR-100-LT. Smaller batches (e.g., $N_b = 20$) lead to noticeably better performance under strong imbalance, whereas larger batches can hurt accuracy, particularly for the most imbalanced settings. This supports the intuition that, on long-tailed image datasets, modelling smaller batches via separate KAHMs helps capture minority classes more faithfully.
\paragraph{Embedding combinations} 
Table~\ref{table_results_10} examines the effect of different embedding models on performance. Using either \texttt{distiluse-base-multilingual-cased-v2} ($\mathrm{mdl}_1$) or \texttt{paraphrase-multilingual} ($\mathrm{mdl}_2$) alone yields strong performance, but their concatenation ($\mathrm{mdl}_1 + \mathrm{mdl}_2$) systematically improves accuracy for all values of $\alpha$. This suggests that the operator-theoretic gradient-free FL model can effectively exploit complementary information from multiple embedding spaces.

\subsection{Inferences Drawn from Experimental Results} 
We summarize the main empirical findings in terms of the research questions posed at the beginning of this section.
\paragraph{Q1 (Robustness to heterogeneity and imbalance)} 
Across all four datasets, the proposed method achieves accuracy that matches or exceeds strong gradient-based FL baselines under non-IID label distributions and long-tailed class imbalance. The sizeable gains on 20Newsgroup, CIFAR-10-LT, and CIFAR-100-LT, together with the improvements on the multilingual XGLUE-NC benchmark, indicate that the proposed FL method is robust to challenging data heterogeneity when built on top of pretrained encoders.
\paragraph{Q2 (Privacy-utility trade-off)}
In the differentially private FL experiments, the method maintains competitive accuracy even under tight privacy budgets. The kernel-based smoothing improves performance in high-privacy regimes ($\epsilon \leq 3$), while having limited or no benefit when privacy constraints are relaxed. This suggests that smoothing should be applied primarily when strong privacy guarantees are required.
\paragraph{Q3 (Secure inference efficiency)} The global gradient-free FL model admits a simple inference procedure suitable for FHE: for a $C-$class problem with $Q$ clients, encrypted inference requires $Q \times C$ \texttt{minimum} and $C$ \texttt{equality-comparison} operations per test point. The measured latencies for these encrypted primitive operations show that, under the evaluated cryptographic parameter settings and for the dataset and client scales studied here, the resulting FHE-secured inference appears computationally feasible on standard hardware.
\paragraph{Q4 (Sensitivity to design choices)} The ablation studies demonstrate that the method is robust to the choice of space folding variant, benefits from smaller batch sizes in the presence of long-tailed imbalance, and can exploit complementary embeddings to further improve accuracy.
\paragraph{Overall summary} Taken together, the experiments suggest that the proposed operator-theoretic framework offers a favourable combination of robustness to heterogeneity, privacy preservation, and practical efficiency for federated learning, within the scope of the benchmarks and settings considered in this study.

\section{Conclusion}\label{sec_conclusion}
The primary contribution of this study is the development of an operator-theoretic kernel framework for the design and analysis of gradient-free federated learning algorithms. The framework addresses the key requirements identified in Section~\ref{sec_introduction} by reformulating the FL problem in the $L^2$ function space, mapping the $L^2$-optimal solution into a reproducing kernel Hilbert space (RKHS) via an invertible operator, and deriving finite-sample performance guarantees using concentration inequalities over operator norms. This yields a gradient-free learning scheme together with non-asymptotic bounds on risk, prediction error, robustness, and approximation error. Within this formulation, we determine a data-dependent hypothesis space by tuning the kernel to the scale of the data and analyse its complexity via Rademacher complexity. The analysis shows that scalar space folding summaries derived from Kernel Affine Hull Machines (KAHMs) are sufficient for the global task learning solution, characterizing when high-dimensional gradient exchanges and multiple communication rounds are not required. In this way, the framework offers a mathematically grounded alternative to traditional gradient-based FL in heterogeneous settings.

The framework further integrates privacy-enhancing and security mechanisms into FL. Differentially private FL is achieved by applying a single optimized noise-adding mechanism to each client's data matrices, followed by kernel-based smoothing and the computation of scalar space folding summaries. By the post-processing property of differential privacy, the resulting global decision rule inherits the $(\epsilon,\delta)-$DP guarantee. Secure FL is enabled by fully homomorphic encryption (FHE) of space folding measures. Because the global decision rule for a $C-$class problem with $Q$ participating clients can be implemented using $Q \times C$ \texttt{minimum} and $C$ \texttt{equality-comparison} operations per test point, the induced FHE-secured inference has a simple and low-dimensional computational structure. Operation-level benchmarks of these encrypted primitives indicate that, for the problem sizes and cryptographic parameter settings studied here, such FHE-secured inference is practically feasible on standard hardware.

Empirically, when combined with embeddings from existing encoders, the resulting gradient-free FL method is competitive with, and in several settings outperforms, strong gradient-based FL methods on non-IID and long-tailed benchmarks. The experiments also indicate that the proposed smoothing mechanism can mitigate the accuracy loss induced by differential privacy in high-privacy regimes, and that the structural simplicity of the FHE-secured decision rule, together with the measured primitive latencies, supports its practical feasibility at the evaluated scales. 

A limitation of the present framework is that it treats feature extractors as fixed and focuses exclusively on the design and analysis of the federated prediction head. While this matches the settings where pretrained encoders are frozen for regulatory, engineering, or cost reasons, it does not directly address end-to-end representation learning under federated constraints. Extending the operator-theoretic construction to encompass joint encoder and head learning, for example via operator-theoretic formulations of representation learning objectives or hybrid gradient-free/gradient-based schemes, is an interesting avenue for future work.

Overall, the main value of this work lies in the unifying operator-theoretic perspective and the associated guarantees, which are largely architectural and model-agnostic and do not depend on a particular dataset, feature encoder, or hardware platform. While our experiments focus on standard non-IID partitions, long-tailed settings, and a reference implementation for concreteness, the theoretical results are intended to remain informative as future work broadens tasks, systems, and deployment conditions.

\section*{Acknowledgments}
The research reported in this paper has been supported by the state of Upper Austria as part of \#upperVISION2030 under the Secure Prescriptive Analytics (SPA) project; project ARIKI [(FFG grant no. 897904)] funded by the Federal Ministry for Climate Action, Environment, Energy, Mobility, Innovation and Technology (BMK), which is linked to a two-stage German funding call ``Development of Digital Technologies'' of the Federal Ministry for Economic Affairs and Climate Action (BMWK); Austrian Ministry for Transport, Innovation and Technology, the Federal Ministry for Digital and Economic Affairs, and the State of Upper Austria in the frame of the SCCH competence center INTEGRATE [(FFG grant no. 892418)] part of the FFG COMET Competence Centers for Excellent Technologies Programme. 

\appendix

\section*{Appendix A. Description of the KAHM Expression (\ref{eq_220420251843})}
With reference to the KAHM expression (\ref{eq_220420251843}), the following definitions are provided:
\begin{itemize}
\item $\mathbf{P}_{\mathbf{X}} \in \mathbb{R}^{\underline{n} \times n}\: (\underline{n} \in \{1,2,\cdots,n \})$ is an encoding matrix such that product $\mathbf{P}_{\mathbf{X}}x$ is a lower-dimensional (i.e. $\underline{n}-$dimensional) encoding for $x$. The encoding matrix is computed from the data samples $\mathbf{X}$ using the following algorithm:  
\begin{algorithm}
\caption{Determination of Encoding Matrix $\mathbf{P}_{\mathbf{X}}$}
\begin{algorithmic}[1]
\Require Matrix $\mathbf{X} \in \mathbb{R}^{N \times n}$, equivalently represented as dataset $\{x^i \in \mathbb{R}^n\}_{i=1}^{N}$.
\State $\underline{n} \gets \min(20,n,N-1)$.
\State  Define $\mathbf{P}_{\mathbf{X}} \in \mathbb{R}^{\underline{n} \times n}$ such that the $i-$th row of $\mathbf{P}_{\mathbf{X}}$ is equal to transpose of eigenvector corresponding to $i-$th largest eigenvalue of sample covariance matrix of samples $\{x^1,\cdots,x^N \}$. 
\While{$\mathop{\min}_{1\leq j \leq \underline{n}}\left( \mathop{\max}_{1 \leq i \leq N} (\mathbf{P}_{\mathbf{X}}x^i)_j  - \mathop{\min}_{1 \leq i \leq N} (\mathbf{P}_{\mathbf{X}}x^i)_j \right) < 1\mathrm{e}{-3}$}
\State  $\underline{n} \gets \underline{n}-1$.
\State  Define $\mathbf{P}_{\mathbf{X}} \in \mathbb{R}^{\underline{n} \times n}$ such that the $i-$th row of $\mathbf{P}_{\mathbf{X}}$ is equal to transpose of eigenvector corresponding to $i-$th largest eigenvalue of sample covariance matrix of dataset $\{x^1,\cdots,x^N \}$. 
\EndWhile
\State \Return $\mathbf{P}_{\mathbf{X}}$.
\end{algorithmic}  
\label{algorithm_encoding_matrix}
\end{algorithm}
\item We have
   \begin{IEEEeqnarray}{rCl} 
\underline{\mathcal{X}} & := & \{ \mathbf{P}_{\mathbf{X}} x \; \mid \; x \in \mathbb{R}^n \},
   \end{IEEEeqnarray} 
and a positive-definite real-valued kernel, $k_{\mathbf{X}}: \underline{\mathcal{X}} \times \underline{\mathcal{X}} \rightarrow \mathbb{R}$ on $\underline{\mathcal{X}}$ with a corresponding reproducing kernel Hilbert space $\mathcal{H}_{k_{\mathbf{X}}}(\underline{\mathcal{X}})$, as 
   \begin{IEEEeqnarray}{rCl}
k_{\mathbf{X}}(\underline{x}^i,\underline{x}^j) & := & \exp\left(-\frac{1}{2\underline{n}}(\underline{x}^i-\underline{x}^j)^T\theta_{\mathbf{X}}^{-1}(\underline{x}^i-\underline{x}^j)\right), 
  \end{IEEEeqnarray} 
where $\underline{x}^i,\underline{x}^j \in \underline{\mathcal{X}}$ and $\theta_{\mathbf{X}}  \succ 0$ is sample covariance matrix of dataset $\{\mathbf{P}_{\mathbf{X}}x^1,\cdots,\mathbf{P}_{\mathbf{X}}x^N \}$.
\item The function $h_{\mathbf{X}}^i: \underline{\mathcal{X}} \rightarrow \mathbb{R}$, such that $h_{\mathbf{X}}^i \in \mathcal{H}_{k_{\mathbf{X}}}(\underline{\mathcal{X}})$, approximates the indicator function $\mathbbm{1}_{\{\mathbf{P}_{\mathbf{X}}x^i\}}: \underline{\mathcal{X}} \rightarrow \{0,1 \}$ as the solution of following kernel regularized least squares problem:  
  \begin{IEEEeqnarray}{rCl}
h_{\mathbf{X}}^i & = & \arg \; \min_{f \in \mathcal{H}_{k_{\mathbf{X}}}(\underline{\mathcal{X}})} \; \left( \sum_{j=1}^N \left |\mathbbm{1}_{\{\mathbf{P}_{\mathbf{X}}x^i\}}(\mathbf{P_{\mathbf{X}}}x^j) - f(\mathbf{P}_{\mathbf{X}}x^j) \right |^2 + \lambda_{\mathbf{X}}^* \left \| f \right \|^2_{\mathcal{H}_{k_{\mathbf{X}}}(\underline{\mathcal{X}})} \right), \IEEEeqnarraynumspace
  \end{IEEEeqnarray}
where the regularization parameter $\lambda_{\mathbf{X}}^* \in \mathbb{R}_+$ is given as
  \begin{IEEEeqnarray}{rCl}
\label{eq_020920251811} \lambda_{\mathbf{X}}^* & = &  \hat{e} + \frac{2}{nN}\|\mathbf{X} \|^2_F, 
     \end{IEEEeqnarray}   
where $\hat{e}$ is the unique fixed point of the function $r$ such that
  \begin{IEEEeqnarray}{rCl}
\label{eq_090120230831}\hat{e} & = & r(\hat{e},\frac{2}{nN}\|\mathbf{X} \|^2_F),
     \end{IEEEeqnarray}   
with $r: \mathbb{R}_{+} \times \mathbb{R}_{+} \rightarrow \mathbb{R}_{+}$ defined as
   \begin{IEEEeqnarray}{rCl}
r(e,\tau)& := &   \frac{1}{nN} \sum_{j=1}^n \|(\mathbf{X})_{:,j} - \mathbf{K}_{\mathbf{X}} \left(\mathbf{K}_{\mathbf{X}} + (e+\tau) \mathbf{I}_N \right)^{-1} (\mathbf{X})_{:,j}\|^2,
     \end{IEEEeqnarray} 
where $(\mathbf{I}_N)_{i,:}$ denotes the $i-$th row of identity matrix of size $N$ and $\mathbf{K}_{\mathbf{X}}$ is $N \times N$ kernel matrix with its $(i,j)-$th element defined as
   \begin{IEEEeqnarray}{rCl}
\label{eq_020920251810} (\mathbf{K}_{\mathbf{X}})_{ij}& := & k_{\mathbf{X}}(\mathbf{P}_{\mathbf{X}}x^i,\mathbf{P}_{\mathbf{X}}x^j).
  \end{IEEEeqnarray} 
The following iterations
 \begin{IEEEeqnarray}{rCl}
e|_{it+1} & = & r(e|_{it},\frac{2}{nN}\|\mathbf{X} \|^2_F),\; it \in \{0,1,\cdots \}  \\
e|_0 & \in & (0,\frac{1}{nN} \|\mathbf{X} \|^2_F)  
  \end{IEEEeqnarray}
converge to $\hat{e}$. The solution of the kernel regularized least squares problem follows as
  \begin{IEEEeqnarray}{rCl}
\label{eq_010920251549} h_{\mathbf{X}}^i(\cdot) & = & (\mathbf{I}_N)_{i,:} \left(\mathbf{K}_{\mathbf{X}} + \lambda_{\mathbf{X}}^* \mathbf{I}_N \right)^{-1}  \left[\begin{IEEEeqnarraybox*}[][c]{,c/c/c,} k_{\mathbf{X}}(\cdot,\mathbf{P}_{\mathbf{X}}x^1) & \cdots & k_{\mathbf{X}}(\cdot,\mathbf{P}_{\mathbf{X}}x^N)\end{IEEEeqnarraybox*} \right]^T
  \end{IEEEeqnarray} 
The value $ h_{\mathbf{X}}^i(\mathbf{P}_{\mathbf{X}}x)$ represents the kernel-smoothed membership of point $\mathbf{P}_{\mathbf{X}}x$ to the set $\{ \mathbf{P}_{\mathbf{X}}x^i\}$.    
\item The image of $\mathcal{A}_{\mathbf{X}}$ defines a region in the affine hull of $\{x^1,\cdots,x^N\}$. That is,
 \begin{IEEEeqnarray}{rCCCl}
 \mathcal{A}_{\mathbf{X}}[\mathbb{R}^n]& := & \{ \mathcal{A}_{\mathbf{X}}(x) \; \mid \; x \in \mathbb{R}^n  \}  & \subset & \mathrm{aff}(\{x^1,\cdots,x^N \}).  
  \end{IEEEeqnarray}   
\end{itemize} 
\section*{Appendix B. Proof of Equation~(\ref{eq_040920240934})}
Consider
 \begin{IEEEeqnarray}{rCl}
\mathop{\argmin}_{g \in L^2(\mathbb{R}^n,\mathbb{P}_{x})} \left( \int_{\mathbb{R}^n\times \{0,1\}^C} |y_c - g(x)|^2 \diff{\mathbb{P}_{x,y}}(x,y) \right)
 & = & \mathop{\argmin}_{g \in L^2(\mathbb{R}^n,\mathbb{P}_{x})} \left( \mathop{\mathbb{E}}_{x \sim \mathbb{P}_{x}} \left [ \mathop{\mathbb{E}}_{y \sim \mathbb{P}_{y | x}} \left [ |y_c - g(x)|^2 \right ]\right ] \right) \\
 & = & \mathop{\argmin}_{g \in L^2(\mathbb{R}^n,\mathbb{P}_{x})} \left( \mathop{\mathbb{E}}_{x \sim \mathbb{P}_{x}} \left [ |g(x)|^2 - 2 g(x) \mathop{\mathbb{E}}_{y \sim \mathbb{P}_{y | x}} \left[ y_c | x  \right]  \right ] \right) \\
 & = &  \mathop{\mathbb{E}}_{y \sim \mathbb{P}_{y | x}} \left[ y_c | x  \right],
 \end{IEEEeqnarray}
where we have considered that $\mathop{\mathbb{E}}_{y \sim \mathbb{P}_{y | x}} \left[ y_c | x  \right] \in L^2(\mathbb{R}^n,\mathbb{P}_{x})$. Thus, (\ref{eq_040920240934}) follows.

\section*{Appendix C. Proof of $\mathcal{K}_{\Phi_c}$ Being a Positive Semi-Definite Kernel}
$\mathcal{K}_{\Phi_c}$ is a positive semi-definite kernel, since
\begin{itemize}
\item $\mathcal{K}_{\Phi_c}(x^1,x^2) = \mathcal{K}_{\Phi_c}(x^2,x^1)$, and
\item for every $x^1,\cdots,x^N \in \mathbb{R}^n$ and $\alpha_1,\cdots,\alpha_N \in \mathbb{R}$, 
 \begin{IEEEeqnarray}{rCl}
\label{eq_030220251850}\sum_{i,j=1}^N \alpha_i \alpha_j \mathcal{K}_{\Phi_c}(x^i,x^j) & \geq & 0.
  \end{IEEEeqnarray}
Inequality (\ref{eq_030220251850}) can be proved by considering that
 \begin{IEEEeqnarray}{rCl}
\sum_{i,j=1}^N \alpha_i \alpha_j \mathcal{K}_{\Phi_c}(x^i,x^j) & = & \sum_{i,j=1}^N \alpha_i \Phi_c(x^i) \alpha_j \Phi_c(x^j) \\
& = & \left| \sum_{i=1}^N \alpha_i \Phi_c(x^i) \right |^2 \\
& \geq & 0.
   \end{IEEEeqnarray}
\end{itemize}

\section*{Appendix D. Proof of $J:\mathcal{H}_{\Phi_c} \hookrightarrow L^2(\mathbb{R}^n,\mathbb{P}_{x})$ Being Well Defined}
Consider for any $f \in \mathcal{H}_{\Phi_c}$,
\begin{IEEEeqnarray}{rCl}
\| Jf \|_{L^2(\mathbb{R}^n,\mathbb{P}_{x})}^2 & = & \int_{\mathbb{R}^n} |f(x)|^2 \diff{\mathbb{P}_{x}}(x) \\
 & = & \int_{\mathbb{R}^n} |\langle f, \mathcal{K}_{\Phi_c}(x,\cdot)\rangle_{\mathcal{H}_{\Phi_c}}|^2 \diff{\mathbb{P}_{x}}(x)\\
& \leq & \int_{\mathbb{R}^n} \| f \|_{\mathcal{H}_{\Phi_c}}^2 \| \mathcal{K}_{\Phi_c}(x,\cdot)\|_{\mathcal{H}_{\Phi_c}}^2 \diff{\mathbb{P}_{x}}(x) \\
& = & \| f \|_{\mathcal{H}_{\Phi_c}}^2 \int_{\mathbb{R}^n}  \| \mathcal{K}_{\Phi_c}(x,\cdot)\|_{\mathcal{H}_{\Phi_c}}^2 \diff{\mathbb{P}_{x}}(x).
  \end{IEEEeqnarray}
Since 
\begin{IEEEeqnarray}{rCl}
\label{eq_24ß920241408}\sup_{x}\mathcal{K}_{\Phi_c}(x,x) & \leq & 1,
\end{IEEEeqnarray}
we have
 \begin{IEEEeqnarray}{rCl}
\label{eq_210720241134}\| \mathcal{K}_{\Phi_c}(x,\cdot)\|_{\mathcal{H}_{\Phi_c}}^2 & \leq & 1, 
  \end{IEEEeqnarray}
and thus
\begin{IEEEeqnarray}{rCCCl}
\| Jf \|_{L^2(\mathbb{R}^n,\mathbb{P}_{x})}^2 & \leq & \| f \|_{\mathcal{H}_{\Phi_c}}^2  & < & \infty. 
  \end{IEEEeqnarray}
That is $Jf \in L^2(\mathbb{R}^n,\mathbb{P}_{x})$. Hence, $J$ is well defined.

\section*{Appendix E. Proof of $(J^*)^{-1}$ Being Well Defined on the Range of $J^*$}
Consider for any $f \in \mathcal{H}_{\Phi_c}$,
\begin{IEEEeqnarray}{rCl}
\| (J^*)^{-1}f \|_{L^2(\mathbb{R}^n,\mathbb{P}_{x})}^2 & = & \frac{1}{\left(\| \Phi_c \|_{L^2(\mathbb{R}^n,\mathbb{P}_{x})}^2 \right)^2} \int_{\mathbb{R}^n} |f(x)|^2 \diff{\mathbb{P}_{x}}(x) \\
 & = & \frac{1}{\left(\| \Phi_c \|_{L^2(\mathbb{R}^n,\mathbb{P}_{x})}^2 \right)^2} \int_{\mathbb{R}^n} |\langle f, \mathcal{K}_{\Phi_c}(x,\cdot)\rangle_{\mathcal{H}_{\Phi_c}}|^2 \diff{\mathbb{P}_{x}}(x)\\
& \leq & \frac{1}{\left(\| \Phi_c \|_{L^2(\mathbb{R}^n,\mathbb{P}_{x})}^2 \right)^2} \int_{\mathbb{R}^n} \| f \|_{\mathcal{H}_{\Phi_c}}^2 \| \mathcal{K}_{\Phi_c}(x,\cdot)\|_{\mathcal{H}_{\Phi_c}}^2 \diff{\mathbb{P}_{x}}(x) \\
& = & \frac{1}{\left(\| \Phi_c \|_{L^2(\mathbb{R}^n,\mathbb{P}_{x})}^2 \right)^2} \| f \|_{\mathcal{H}_{\Phi_c}}^2 \int_{\mathbb{R}^n}  \| \mathcal{K}_{\Phi_c}(x,\cdot)\|_{\mathcal{H}_{\Phi_c}}^2 \diff{\mathbb{P}_{x}}(x)  \\
& \leq & \frac{1}{\left(\| \Phi_c \|_{L^2(\mathbb{R}^n,\mathbb{P}_{x})}^2 \right)^2} \| f \|_{\mathcal{H}_{\Phi_c}}^2,
  \end{IEEEeqnarray}  
where we have used (\ref{eq_210720241134}). Due to (\ref{eq_270920241923}), we have
\begin{IEEEeqnarray}{rCl}
\| (J^*)^{-1}f \|_{L^2(\mathbb{R}^n,\mathbb{P}_{x})}^2 & < & \infty.
  \end{IEEEeqnarray}
That is, $(J^*)^{-1}f \in L^2(\mathbb{R}^n,\mathbb{P}_{x})$. Hence, $(J^*)^{-1}$ is well defined on the range of $J^*$.

\section*{Appendix F. Proof of Equation (\ref{eq_080220251753})}
Consider
 \begin{IEEEeqnarray}{rCl}
(J^*J)f & = & \mathop{\mathbb{E}}_{x' \sim \mathbb{P}_{x}} \left[ \mathcal{K}_{\Phi_c}(x',\cdot)(Jf)(x') \right] \\
& = & \mathop{\mathbb{E}}_{x' \sim \mathbb{P}_{x}} \left[ \mathcal{K}_{\Phi_c}(x',\cdot)f(x') \right] \\
& = & \mathop{\mathbb{E}}_{x' \sim \mathbb{P}_{x}} \left[ \mathcal{K}_{\Phi_c}(x',\cdot) \langle f, \mathcal{K}_{\Phi_c}(x',\cdot)  \rangle_{\mathcal{H}_{\Phi_c}} \right] \\
& = & \mathop{\mathbb{E}}_{x' \sim \mathbb{P}_{x}} \left[ (\mathcal{K}_{\Phi_c}(x',\cdot)  \otimes \mathcal{K}_{\Phi_c}(x',\cdot) )(f) \right].
   \end{IEEEeqnarray}
Thus, (\ref{eq_080220251753}) follows.

\section*{Appendix G. Proof of Inequality (\ref{eq_260720241423})}
Consider for any $f \in \mathcal{H}_{\Phi_c}$,
 \begin{IEEEeqnarray}{rCl}
\| (J^*J) f \|_{\mathcal{H}_{\Phi_c}}  & = & \left \| \mathop{\mathbb{E}}_{x' \sim \mathbb{P}_{x}} \left[ (\mathcal{K}_{\Phi_c}(x',\cdot)  \otimes \mathcal{K}_{\Phi_c}(x',\cdot) )(f) \right] \right \|_{\mathcal{H}_{\Phi_c}} \\
& \leq & \mathop{\mathbb{E}}_{x' \sim \mathbb{P}_{x}} \left[ \left \|  (\mathcal{K}_{\Phi_c}(x',\cdot)  \otimes \mathcal{K}_{\Phi_c}(x',\cdot) )(f)  \right \|_{\mathcal{H}_{\Phi_c}} \right] \\
& = & \mathop{\mathbb{E}}_{x' \sim \mathbb{P}_{x}} \left[ | \langle f, \mathcal{K}_{\Phi_c}(x',\cdot)  \rangle_{\mathcal{H}_{\Phi_c}} | \|  \mathcal{K}_{\Phi_c}(x',\cdot) \|_{\mathcal{H}_{\Phi_c}} \right] \\
& \leq & \mathop{\mathbb{E}}_{x' \sim \mathbb{P}_{x}} \left[ \| f \|_{\mathcal{H}_{\Phi_c}} \| \mathcal{K}_{\Phi_c}(x',\cdot)  \|_{\mathcal{H}_{\Phi_c}}^2 \right] \\
& \leq & \| f \|_{\mathcal{H}_{\Phi_c}}.
   \end{IEEEeqnarray}
Thus, (\ref{eq_260720241423}) follows.

\section*{Appendix H. Proof of Inequality (\ref{eq_240920241954})}
Consider
\begin{IEEEeqnarray}{rCl}
\mathop{\mathbb{E}}_{x \sim \mathbb{P}_{x}} \left [ \left | h_{x \mapsto y_c}^{\mathcal{H}_{\Phi_c}}(x)-f_{x \mapsto y_c}^{\mathcal{H}_{\Phi_c}}(x) \right |^2 \right]
& = & \left \| J (h_{x \mapsto y_c}^{\mathcal{H}_{\Phi_c}}-f_{x \mapsto y_c}^{\mathcal{H}_{\Phi_c}}) \right \|_{L^2(\mathbb{R}^n,\mathbb{P}_{x})}\\
 & = & \left( \left \langle J (h_{x \mapsto y_c}^{\mathcal{H}_{\Phi_c}}-f_{x \mapsto y_c}^{\mathcal{H}_{\Phi_c}}), J (h_{x \mapsto y_c}^{\mathcal{H}_{\Phi_c}}-f_{x \mapsto y_c}^{\mathcal{H}_{\Phi_c}}) \right \rangle_{L^2(\mathbb{R}^n,\mathbb{P}_{x})} \right)^{1/2} \\
& = & \left( \left \langle  (h_{x \mapsto y_c}^{\mathcal{H}_{\Phi_c}}-f_{x \mapsto y_c}^{\mathcal{H}_{\Phi_c}}), J^*J (h_{x \mapsto y_c}^{\mathcal{H}_{\Phi_c}}-f_{x \mapsto y_c}^{\mathcal{H}_{\Phi_c}}) \right \rangle_{\mathcal{H}_{\Phi_c}} \right)^{1/2} \\
& = & \left \| (J^*J)^{1/2} \left (h_{x \mapsto y_c}^{\mathcal{H}_{\Phi_c}}-J^*f_{x \mapsto y_c} \right) \right \|_{\mathcal{H}_{\Phi_c}}. 
\end{IEEEeqnarray}
Using (\ref{eq_090220151301}), we have
\begin{IEEEeqnarray}{rCl}
\nonumber \lefteqn{\mathop{\mathbb{E}}_{x \sim \mathbb{P}_{x}} \left [ \left | h_{x \mapsto y_c}^{\mathcal{H}_{\Phi_c}}(x)-f_{x \mapsto y_c}^{\mathcal{H}_{\Phi_c}}(x) \right |^2 \right]}\\
\nonumber & = &  \left \| (J^*J)^{1/2}  \left(  \widehat{S}^*_{(x^i)_{i=1}^N} \mathrm{Ev}_{(x^i)_{i=1}^N}  - J^* \right) f_{x \mapsto y_c}  + (J^*J)^{1/2}  \widehat{S}^*_{(x^i)_{i=1}^N} \left( \xi_c(x^{1}, y^{1}),  \cdots,  \xi_c(x^{N}, y^{N}) \right) \right \|_{\mathcal{H}_{\Phi_c}} \\
 & \leq &  \| (J^*J)^{1/2}  \|_{\op} \left \|  \left(  \widehat{S}^*_{(x^i)_{i=1}^N} \mathrm{Ev}_{(x^i)_{i=1}^N}  - J^* \right) f_{x \mapsto y_c}   \right \|_{\mathcal{H}_{\Phi_c}} +   \| (J^*J)^{1/2}  \|_{\op} \left \|     \widehat{S}^*_{(x^i)_{i=1}^N} \left( \xi_c(x^{1}, y^{1}),  \cdots,  \xi_c(x^{N}, y^{N}) \right)  \right \|_{\mathcal{H}_{\Phi_c}}. \IEEEeqnarraynumspace
\end{IEEEeqnarray}
Using (\ref{eq_260720241424}), we get (\ref{eq_240920241954}).

\section*{Appendix I. Proof of Inequality (\ref{eq_100920241949})}
Define, for a given $x \in \mathbb{R}^n$, $F_{x,c} \in \mathcal{H}_{\Phi_c}$ as
 \begin{IEEEeqnarray}{rCl}
\label{eq_011020241233}  F_{x,c}  & := & \mathcal{K}_{\Phi_c}(x,\cdot) f_{x \mapsto y_c}(x)  - J^* f_{x \mapsto y_c}.
   \end{IEEEeqnarray}   
Considering $x$ as a random variable, $ F_{x,c}$ is a random variable taking values in $\mathcal{H}_{\Phi_c}$ with mean equal to the zero function, i.e.,
  \begin{IEEEeqnarray}{rCl}
\label{eq_100920241655} \mathop{\mathbb{E}}_{x \sim \mathbb{P}_x} \left [ F_{x,c}  \right ]& = & \mathbf{0},
   \end{IEEEeqnarray}
where $\mathbf{0}:\mathbb{R}^n\rightarrow 0$. Consider
 \begin{IEEEeqnarray}{rCl}
  \left \|  F_{x,c} \right \|_{\mathcal{H}_{\Phi_c}} & = &  \left \|  f_{x \mapsto y_c}(x) \mathcal{K}_{\Phi_c}(x,\cdot) -  \mathop{\mathbb{E}}_{x \sim \mathbb{P}_{x}} \left[ \mathcal{K}_{\Phi_c}(x,\cdot)f_{x \mapsto y_c}(x) \right] \right \|_{\mathcal{H}_{\Phi_c}} \\
  & \leq & \left \|  f_{x \mapsto y_c}(x) \mathcal{K}_{\Phi_c}(x,\cdot) \right \|_{\mathcal{H}_{\Phi_c}} + \left \|  \mathop{\mathbb{E}}_{x \sim \mathbb{P}_{x}} \left[ \mathcal{K}_{\Phi_c}(x,\cdot)f_{x \mapsto y_c}(x) \right] \right \|_{\mathcal{H}_{\Phi_c}}\\
\label{eq_100920241319}  & \leq & \left \|  f_{x \mapsto y_c}(x) \mathcal{K}_{\Phi_c}(x,\cdot) \right \|_{\mathcal{H}_{\Phi_c}} + \mathop{\mathbb{E}}_{x \sim \mathbb{P}_{x}} \left[ \left \|   \mathcal{K}_{\Phi_c}(x,\cdot)f_{x \mapsto y_c}(x)  \right \|_{\mathcal{H}_{\Phi_c}} \right] \\
\label{eq_100920241320}  & \leq &  2, 
     \end{IEEEeqnarray}   
where (\ref{eq_100920241320}) follows from (\ref{eq_100920241319}) using (\ref{eq_100920241321}) and (\ref{eq_210720241134}). Now, Consider
 \begin{IEEEeqnarray}{rCl}
\mathop{\mathbb{E}}_{\left((x^i,y^i) \sim \mathbb{P}_{x,y} \right)_{i=1}^{N}} \left [ \left \|  \left(  \widehat{S}^*_{(x^i)_{i=1}^N} \mathrm{Ev}_{(x^i)_{i=1}^N}  - J^* \right) f_{x \mapsto y_c}    \right \|_{\mathcal{H}_{\Phi_c}} \right ] & \leq & \left( \mathop{\mathbb{E}}_{\left((x^i,y^i) \sim \mathbb{P}_{x,y } \right)_{i=1}^{N}} \left [  \left \|  \left(  \widehat{S}^*_{(x^i)_{i=1}^N} \mathrm{Ev}_{(x^i)_{i=1}^N}  - J^* \right) f_{x \mapsto y_c}    \right \|_{\mathcal{H}_{\Phi_c}}^2 \right ] \right)^{1/2} \IEEEeqnarraynumspace \\
 & = &  \left(  \frac{1}{N^2} \mathop{\mathbb{E}}_{\left((x^i,y^i) \sim \mathbb{P}_{x,y} \right)_{i=1}^{N}} \left [  \left \|  \sum_{i=1}^{N}  F_{x^{i},c}  \right \|_{\mathcal{H}_{\Phi_c}}^2  \right ] \right)^{1/2} \\
 & = & \left(  \frac{1}{N^2} \mathop{\mathbb{E}}_{\left((x^i,y^i) \sim \mathbb{P}_{x,y} \right)_{i=1}^{N}} \left [  \sum_{i=1}^{N} \sum_{j=1}^{N} \left \langle  F_{x^{i},c}, F_{x^{j},c} \right \rangle_{\mathcal{H}_{\Phi_c}}  \right ] \right)^{1/2} \\
  & = & \left(  \frac{1}{N^2}  \sum_{i=1}^{N} \sum_{j=1}^{N} \mathop{\mathbb{E}}_{(x^{i},y^{i}) \sim \mathbb{P}_{x,y},(x^{j},y^{j}) \sim \mathbb{P}_{x,y} } \left [  \left \langle   F_{x^{i},c}, F_{x^{j},c} \right \rangle_{\mathcal{H}_{\Phi_c}}  \right ] \right)^{1/2} .
    \end{IEEEeqnarray}
Using the independence of the samples $((x^i,y^i))_{i=1}^N$ and (\ref{eq_100920241655}), we have
     \begin{IEEEeqnarray}{rCl}
\mathop{\mathbb{E}}_{(x^{i},y^{i}) \sim \mathbb{P}_{x,y},(x^{j},y^{j}) \sim \mathbb{P}_{x,y}} \left [  \left \langle   F_{x^{i},c}, F_{x^{j},c} \right \rangle_{\mathcal{H}_{\Phi_c}} \right ] & = &   0,\;\mbox{if $i \neq j$.}   
      \end{IEEEeqnarray}
Therefore,
\begin{IEEEeqnarray}{rCl}
\mathop{\mathbb{E}}_{\left((x^i,y^i) \sim \mathbb{P}_{x,y} \right)_{i=1}^{N}} \left [ \left \|  \left(  \widehat{S}^*_{(x^i)_{i=1}^N} \mathrm{Ev}_{(x^i)_{i=1}^N}  - J^* \right) f_{x \mapsto y_c}  \right \|_{\mathcal{H}_{\Phi_c}} \right ] & \leq&  \left(  \frac{1}{N^2}  \sum_{i=1}^{N} \mathop{\mathbb{E}}_{(x^{i},y^{i}) \sim \mathbb{P}_{x,y} } \left [  \left \|  F_{x^{i},c} \right \|_{\mathcal{H}_{\Phi_c}}^2 \right ]  \right)^{1/2} \\ 
 & \leq & \frac{2}{\sqrt{N}}.
     \end{IEEEeqnarray}
where we have used (\ref{eq_100920241320}).
\section*{Appendix J. Proof of Inequality (\ref{eq_011020241149})}
Consider
 \begin{IEEEeqnarray}{rCl}
\nonumber \lefteqn{\mathop{\mathbb{E}}_{\left((x^i, y^i) \sim \mathbb{P}_{x, y} \right)_{i=1}^{N}} \left [ \left \| \widehat{S}^*_{(x^i)_{i=1}^N} ( \xi_c(x^{1}, y^{1}),  \cdots,  \xi_c(x^{N}, y^{N}) ) \right \|_{\mathcal{H}_{\Phi_c}} \right ]}\\
 & \leq & \left( \mathop{\mathbb{E}}_{\left((x^i, y^i) \sim \mathbb{P}_{x, y} \right)_{i=1}^{N}} \left [  \left \| \widehat{S}^*_{(x^i)_{i=1}^N} ( \xi_c(x^{1}, y^{1}),  \cdots,   \xi_c(x^{N}, y^{N}) ) \right \|_{\mathcal{H}_{\Phi_c}}^2 \right ] \right)^{1/2} \\
 & = &  \left( \mathop{\mathbb{E}}_{\left((x^i,y^i) \sim \mathbb{P}_{x,y} \right)_{i=1}^{N}} \left [ \frac{1}{N^2} \sum_{i=1}^{N}\sum_{j=1}^{N} \xi_c(x^{i},y^{i}) \xi_c(x^{j},y^{j})  \mathcal{K}_{\Phi_c}(x^{i},x^{j}) \right ] \right)^{1/2} \\
 & = & \left( \frac{1}{N^2} \sum_{i=1}^{N} \mathop{\mathbb{E}}_{(x^{i},y^{i}) \sim \mathbb{P}_{x, y } } \left [  | \xi_c(x^{i},y^{i})  |^2 \mathcal{K}_{\Phi_c}(x^{i}, x^{i})  \right ]\right)^{1/2},
    \end{IEEEeqnarray}  
where we have used the independence of the samples $((x^i,y^i))_{i=1}^{N}$ and (\ref{eq_240920241406}). Using (\ref{eq_24ß920241408}) and (\ref{eq_240920241255}), we get (\ref{eq_011020241149}).

\section*{Appendix K: Proof of Inequality (\ref{eq_011020241156})}
Define a function $\psi_{c} : (\mathbb{R}^n\times \{0,1\}^C)^N \rightarrow \mathbb{R}_{\geq 0}$ as
 \begin{IEEEeqnarray}{rCl}
\psi_c & :=  & \psi_{1,c}\left((x^1,y^1),\cdots,(x^N,y^N)\right) + \psi_{2,c}\left((x^1,y^1),\cdots,(x^N,y^N)\right),
   \end{IEEEeqnarray} 
where $\psi_{1,c}: (\mathbb{R}^n\times \{0,1\}^C)^N \rightarrow \mathbb{R}_{\geq 0}$ and $\psi_{2,c}: (\mathbb{R}^n\times \{0,1\}^C)^N \rightarrow \mathbb{R}_{\geq 0}$ are defined as 
   \begin{IEEEeqnarray}{rCl}
\label{eq_011020240951}\psi_{1,c}\left((x^1,y^1),\cdots,(x^N,y^N)\right)& : =  & \left \|  \left(  \widehat{S}^*_{(x^i)_{i=1}^N} \mathrm{Ev}_{(x^i)_{i=1}^N}  - J^* \right) f_{x \mapsto y_c} \right \|_{\mathcal{H}_{\Phi_c}} \\
\label{eq_011020240952}\psi_{2,c}\left( (x^1, y^1),\cdots, (x^N, y^N) \right) & : = & \left \| \widehat{S}^*_{(x^i)_{i=1}^N} \left ( \xi_c(x^{1}, y^1),  \cdots,   \xi_c(x^{N}, y^N) \right ) \right \|_{\mathcal{H}_{\Phi_c}}.
       \end{IEEEeqnarray} 
It can be seen that 
   \begin{IEEEeqnarray}{rCl}
\psi_{1,c}\left((x^1,y^1),\cdots,(x^N,y^N)\right)& : =  &  \frac{1}{N}  \left \|   \sum_{i=1}^{N}  F_{x^{i},c}  \right \|_{\mathcal{H}_{\Phi_c}},
       \end{IEEEeqnarray} 
where $F_{x,c}$ is defined as in (\ref{eq_011020241233}). Consider
\begin{IEEEeqnarray}{rCl}
\nonumber \lefteqn{  \left | \frac{\left \|     F_{x^{1},c} + \cdots +  F_{x^{i},c} + \cdots +  F_{x^{N},c}   \right \|_{\mathcal{H}_{\Phi_c}}}{N}    - \frac{\left \|     F_{x^{1},c} + \cdots +  F_{x'^{i},c} + \cdots +  F_{x^{N},c}   \right \|_{\mathcal{H}_{\Phi_c}}}{N}   \right |} \\
& \leq  & \frac{1}{N} \left \|  F_{x^{i},c}  - F_{x'^{i},c} \right \|_{\mathcal{H}_{\Phi_c}} \\
& = & \frac{1}{N} \left \| f_{x \mapsto y_c}(x^{i})\mathcal{K}_{\Phi_c}(x^{i},\cdot) - f_{x \mapsto y_c}(x'^{i})\mathcal{K}_{\Phi_c}(x'^{i},\cdot)   \right \|_{\mathcal{H}_{\Phi_c}} \\
& \leq & \frac{1}{N} \left( \left | f_{x \mapsto y_c}(x^{i}) \right| \left \| \mathcal{K}_{\Phi_c}(x^{i},\cdot)\right \|_{\mathcal{H}_{\Phi_c}} + \left | f_{x \mapsto y_c}(x'^{i}) \right| \left \| \mathcal{K}_{\Phi_c}(x'^{i},\cdot)\right \|_{\mathcal{H}_{\Phi_c}}\right) \\
& \leq & \frac{2}{N}.
\end{IEEEeqnarray} 
Thus,
\begin{IEEEeqnarray}{rCl}
\nonumber \frac{2}{N} & \geq & \sup_{(x'^i, y'^i) \in \mathbb{R}^n\times  \{0,1 \}^C} \left( \left | \psi_{1,c}\left((x^1,y^1),\cdots,(x^i,y^i),\cdots,(x^N,y^N)\right) \right. \right. \\
\label{eq_011020241004} && \left. \left. - \psi_{1,c}\left((x^1,y^1),\cdots,(x'^i,y'^i),\cdots,(x^N,y^N)\right) \right | \right).
\end{IEEEeqnarray} 
Now consider
 \begin{IEEEeqnarray}{rCl}
\nonumber \lefteqn{\left | \left \| \widehat{S}^*_{(x^{1},\cdots,x^{i},\cdots,x^{N})}\left(\xi_c(x^{1}, y^{1}),  \cdots, \xi_c(x^{i}, y^{i}), \cdots,   \xi_c(x^{N}, y^{N}) \right) \right \|_{\mathcal{H}_{\Phi_c}} \right.} \\
\nonumber && \left. - \left \| \widehat{S}^*_{(x^{1},\cdots,x'^{i},\cdots,x^{N})} ( \xi_c(x^{1}, y^{1}),  \cdots, \xi_c(x'^{i},y'^{i}),\cdots,  \xi_c(x^{N}, y^{N}) ) \right \|_{\mathcal{H}_{\Phi_c}}  \right | \\
\nonumber & \leq & \left \| \widehat{S}^*_{(x^{1},\cdots,x^{i},\cdots,x^{N})}\left(\xi_c(x^{1}, y^{1}),  \cdots, \xi_c(x^{i}, y^{i}), \cdots,   \xi_c(x^{N}, y^{N}) \right)\right. \\
&& \left. - \widehat{S}^*_{(x^{1},\cdots,x'^{i},\cdots,x^{N})} ( \xi_c(x^{1}, y^{1}),  \cdots, \xi_c(x'^{i},y'^{i}),\cdots,  \xi_c(x^{N}, y^{N}) )  \right \|_{\mathcal{H}_{\Phi_c}} \\
& = & \frac{1}{N} \left \|   \xi_c(x^{i},y^{i})\mathcal{K}_{\Phi_c}(x^{i},\cdot) - \xi_c(x'^{i},y'^{i}) \mathcal{K}_{\Phi_c}(x'^{i},\cdot) \right \|_{\mathcal{H}_{\Phi_c}} \\
& \leq & \frac{1}{N} | \xi_c(x^{i},y^{i}) | \left \|  \mathcal{K}_{\Phi_c}(x^{i},\cdot) \right \|_{\mathcal{H}_{\Phi_c}} + \frac{1}{N} | \xi_c(x'^{i},y'^{i}) |  \| \mathcal{K}_{\Phi_c}(x'^{i},\cdot)  \|_{\mathcal{H}_{\Phi_c}}  \\
& \leq & \frac{2}{N},
   \end{IEEEeqnarray} 
where we have used (\ref{eq_210720241134}) and (\ref{eq_240920241255}). Thus,
 \begin{IEEEeqnarray}{rCl}
\nonumber \frac{2}{N} & \geq & \sup_{(x'^i, y'^i) \in \mathbb{R}^n\times  \{0,1 \}^C } \left( \left | \psi_{2,c}\left( (x^1, y^1),\cdots, (x^i,  y^i), \cdots, (x^N,  y^N) \right) \right. \right. \\
\label{eq_011020241005} && \left. \left. - \psi_{2,c}\left( (x^1,  y^1),\cdots,(x'^i,  y'^i),\cdots  (x^N,  y^N) \right) \right | \right).
   \end{IEEEeqnarray} 
It follows from (\ref{eq_011020241004}) and (\ref{eq_011020241005}) that
 \begin{IEEEeqnarray}{rCl}
\nonumber \frac{4}{N} & \geq & \sup_{(x'^i, y'^i) \in \mathbb{R}^n\times  \{0,1 \}^C } \left( \left | \psi_{c}\left( (x^1, y^1),\cdots, (x^i,  y^i), \cdots, (x^N,  y^N) \right) \right. \right. \\
\label{eq_011020241006} && \left. \left. - \psi_{c}\left( (x^1,  y^1),\cdots,(x'^i,  y'^i),\cdots  (x^N,  y^N) \right) \right | \right).
   \end{IEEEeqnarray} 
Thus, $\psi_{c}$ satisfies the bounded differences property with bound $4/N$, and therefore by McDiarmid's inequality, for any $\epsilon > 0$, with probability at most $\exp(- 0.125N \epsilon^2)$, the following holds:
 \begin{IEEEeqnarray}{rCl}
 \epsilon & \leq & \psi_{c}\left((x^1,y^1),\cdots,(x^i,y^i),\cdots,(x^N,y^N)\right) -  \mathop{\mathbb{E}}_{\left((x^i,y^i) \sim \mathbb{P}_{x,y} \right)_{i=1}^{N}} \left [ \psi_{c}\left((x^1,y^1),\cdots,(x^i,y^i),\cdots,(x^N,y^N)\right)  \right ]. \IEEEeqnarraynumspace
   \end{IEEEeqnarray}  
That is, with probability at most $\exp(- 0.125N \epsilon^2)$, the following holds:
 \begin{IEEEeqnarray}{rCl}
\nonumber \epsilon & \leq & \left \|  \left(  \widehat{S}^*_{(x^i)_{i=1}^N} \mathrm{Ev}_{(x^i)_{i=1}^N}  - J^* \right) f_{x \mapsto y_c} \right \|_{\mathcal{H}_{\Phi_c}} + \left \| \widehat{S}^*_{(x^i)_{i=1}^N} \left ( \xi_c(x^{1}, y^1),  \cdots,   \xi_c(x^{N}, y^N) \right ) \right \|_{\mathcal{H}_{\Phi_c}} \\
 && {-}\: \mathop{\mathbb{E}}_{\left((x^i,y^i) \sim \mathbb{P}_{x,y} \right)_{i=1}^{N}} \left [ \left \|  \left(  \widehat{S}^*_{(x^i)_{i=1}^N} \mathrm{Ev}_{(x^i)_{i=1}^N}  - J^* \right) f_{x \mapsto y_c} \right \|_{\mathcal{H}_{\Phi_c}} + \left \| \widehat{S}^*_{(x^i)_{i=1}^N} \left ( \xi_c(x^{1}, y^1),  \cdots,   \xi_c(x^{N}, y^N) \right ) \right \|_{\mathcal{H}_{\Phi_c}}   \right ].
 \end{IEEEeqnarray}  
That is, with probability at most $\delta > 0$, the following holds:
 \begin{IEEEeqnarray}{rCl}
\nonumber \lefteqn{\sqrt{\frac{8\log(1/\delta)}{N}}} \\
\nonumber & \leq & \left \|  \left(  \widehat{S}^*_{(x^i)_{i=1}^N} \mathrm{Ev}_{(x^i)_{i=1}^N}  - J^* \right) f_{x \mapsto y_c} \right \|_{\mathcal{H}_{\Phi_c}} + \left \| \widehat{S}^*_{(x^i)_{i=1}^N} \left ( \xi_c(x^{1}, y^1),  \cdots,   \xi_c(x^{N}, y^N) \right ) \right \|_{\mathcal{H}_{\Phi_c}} \\
 && {-}\: \mathop{\mathbb{E}}_{\left((x^i,y^i) \sim \mathbb{P}_{x,y} \right)_{i=1}^{N}} \left [ \left \|  \left(  \widehat{S}^*_{(x^i)_{i=1}^N} \mathrm{Ev}_{(x^i)_{i=1}^N}  - J^* \right) f_{x \mapsto y_c} \right \|_{\mathcal{H}_{\Phi_c}} + \left \| \widehat{S}^*_{(x^i)_{i=1}^N} \left ( \xi_c(x^{1}, y^1),  \cdots,   \xi_c(x^{N}, y^N) \right ) \right \|_{\mathcal{H}_{\Phi_c}}   \right ].
 \end{IEEEeqnarray}  
In other words, with probability at least $1-\delta$, the following holds:  
 \begin{IEEEeqnarray}{rCl}
\nonumber \lefteqn{\sqrt{\frac{8\log(1/\delta)}{N}}} \\
\nonumber & \geq & \left \|  \left(  \widehat{S}^*_{(x^i)_{i=1}^N} \mathrm{Ev}_{(x^i)_{i=1}^N}  - J^* \right) f_{x \mapsto y_c} \right \|_{\mathcal{H}_{\Phi_c}} + \left \| \widehat{S}^*_{(x^i)_{i=1}^N} \left ( \xi_c(x^{1}, y^1),  \cdots,   \xi_c(x^{N}, y^N) \right ) \right \|_{\mathcal{H}_{\Phi_c}} \\
\label{eq_011020241250} && {-}\: \mathop{\mathbb{E}}_{\left((x^i,y^i) \sim \mathbb{P}_{x,y} \right)_{i=1}^{N}} \left [ \left \|  \left(  \widehat{S}^*_{(x^i)_{i=1}^N} \mathrm{Ev}_{(x^i)_{i=1}^N}  - J^* \right) f_{x \mapsto y_c} \right \|_{\mathcal{H}_{\Phi_c}} + \left \| \widehat{S}^*_{(x^i)_{i=1}^N} \left ( \xi_c(x^{1}, y^1),  \cdots,   \xi_c(x^{N}, y^N) \right ) \right \|_{\mathcal{H}_{\Phi_c}}   \right ].
 \end{IEEEeqnarray}  
Using (\ref{eq_100920241949}) and (\ref{eq_011020241149}) in (\ref{eq_011020241250}), we get (\ref{eq_011020241156}).

\section*{Appendix L: Proof of Theorem~\ref{theorem_110220250938}}
Consider
 \begin{IEEEeqnarray}{rCl}
\widehat{\mathcal{R}}_{\mathcal{D}}( \mathcal{M}_{c} ) & = & \frac{1}{N} \mathop{\mathbb{E}}_{\sigma } \left[ \sup_{h_{x \mapsto y_c} \in \mathcal{M}_{c}  } \sum_{i=1}^N \sigma_i \: h_{x \mapsto y_c}(x^i) \right]\\
& = & \frac{1}{N} \mathop{\mathbb{E}}_{\sigma } \left[ \sup_{h_{x \mapsto y_c} \in \mathcal{M}_{c}  } \sum_{i=1}^N \sigma_i \: \left \langle h_{x \mapsto y_c}, \mathcal{K}_{\Phi_c}(\cdot,x^i) \right \rangle_{ \mathcal{H}_{\Phi_c}} \right],
  \end{IEEEeqnarray} 
where we have used the reproducing property of the kernel, since $h_{x \mapsto y_c} \in \mathcal{H}_{\Phi_c}$. That is,
\begin{IEEEeqnarray}{rCl}
\label{eq_111220241244}\widehat{\mathcal{R}}_{\mathcal{D}}( \mathcal{M}_{c} ) & = & \frac{1}{N} \mathop{\mathbb{E}}_{\sigma } \left[ \sup_{h_{x \mapsto y_c} \in \mathcal{M}_{c}  }  \: \left \langle h_{x \mapsto y_c}, \sum_{i=1}^N \sigma_i \mathcal{K}_{\Phi_c}(\cdot,x^i) \right \rangle_{ \mathcal{H}_{\Phi_c}} \right] \\
\label{eq_111220241245}& \leq & \frac{1}{N} \mathop{\mathbb{E}}_{\sigma } \left[ \sup_{h_{x \mapsto y_c} \in \mathcal{M}_{c}  }  \: \left( \| h_{x \mapsto y_c} \|_{\mathcal{H}_{\Phi_c}} \left \| \sum_{i=1}^N \sigma_i \mathcal{K}_{\Phi_c}(\cdot,x^i) \right \|_{\mathcal{H}_{\Phi_c}} \right)  \right],
 \end{IEEEeqnarray}
where (\ref{eq_111220241245}) follows from (\ref{eq_111220241244}) due to Cauchy–Schwarz inequality. Also,
\begin{IEEEeqnarray}{rCl}
\label{eq_111220241437}\left \| \sum_{i=1}^N \sigma_i \mathcal{K}_{\Phi_c}(\cdot,x^i) \right \|_{\mathcal{H}_{\Phi_c}} & = & \left | \sum_{i=1}^N \sigma_i \Phi_c(x^i) \right |.
  \end{IEEEeqnarray} 
Using (\ref{eq_111220241416}) and (\ref{eq_111220241437}),   
\begin{IEEEeqnarray}{rCl}
\widehat{\mathcal{R}}_{\mathcal{D}}( \mathcal{M}_{c} ) & \leq & \frac{1}{N} \mathop{\mathbb{E}}_{\sigma } \left[ \sup_{\Phi_c: \mathbb{R}^n \rightarrow [0,1] }  \: \left(  \frac{\sum_{i=1}^{N_c} \Phi_c(x^{\mathrm{I}_i^{c}})}{N_c} \left | \sum_{i=1}^N \sigma_i \Phi_c(x^i) \right | \right) \right] \\
\label{eq_111220241841}& \leq  &  \frac{1}{N}  \mathop{\mathbb{E}}_{\sigma } \left[  \sup_{\Phi_c: \mathbb{R}^n \rightarrow [0,1] }  \: \left( \left | \sum_{i=1}^N \sigma_i \Phi_c(x^i) \right | \right) \right].
 \end{IEEEeqnarray}
For any $\epsilon > 0$, let $\overline{\Phi}_{c,\sigma} : \mathbb{R}^n \rightarrow [0,1]$ be such that
\begin{IEEEeqnarray}{rCl}
\label{eq_111220241840}\sup_{\Phi_c: \mathbb{R}^n \rightarrow [0,1] }  \: \left( \left | \sum_{i=1}^N \sigma_i \Phi_c(x^i) \right | \right) & = & \left | \sum_{i=1}^N \sigma_i \overline{\Phi}_{c,\sigma}(x^i) \right | + \epsilon.
 \end{IEEEeqnarray}
Using (\ref{eq_111220241840}) in (\ref{eq_111220241841}),
\begin{IEEEeqnarray}{rCl}
\widehat{\mathcal{R}}_{\mathcal{D}}( \mathcal{M}_{c} ) & \leq & \frac{1}{N} \mathop{\mathbb{E}}_{\sigma } \left[  \left | \sum_{i=1}^N \sigma_i \overline{\Phi}_{c,\sigma}(x^i) \right | \right] + \frac{\epsilon}{N}.
\end{IEEEeqnarray}
As per Jensen's inequality,
 \begin{IEEEeqnarray}{rCl}
 \left( \mathop{\mathbb{E}}_{\sigma } \left[  \left | \sum_{i=1}^N \sigma_i \overline{\Phi}_{c,\sigma}(x^i) \right | \right]\right)^2 & \leq & \mathop{\mathbb{E}}_{\sigma } \left[  \left | \sum_{i=1}^N \sigma_i \overline{\Phi}_{c,\sigma}(x^i) \right |^2 \right],
 \end{IEEEeqnarray} 
and thus 
\begin{IEEEeqnarray}{rCl}
\label{eq_121220240924}\widehat{\mathcal{R}}_{\mathcal{D}}( \mathcal{M}_{c} ) & \leq & \frac{1}{N}  \sqrt{\mathop{\mathbb{E}}_{\sigma } \left[  \left | \sum_{i=1}^N \sigma_i \overline{\Phi}_{c,\sigma}(x^i) \right |^2 \right]} + \frac{\epsilon}{N}.
 \end{IEEEeqnarray} 
Consider
\begin{IEEEeqnarray}{rCl}
\mathop{\mathbb{E}}_{\sigma } \left[  \left | \sum_{i=1}^N \sigma_i \overline{\Phi}_{c,\sigma}(x^i) \right |^2 \right] & = & \mathop{\mathbb{E}}_{\sigma } \left[  \sum_{i,j=1}^N \sigma_i \sigma_j \overline{\Phi}_{c,\sigma}(x^i) \overline{\Phi}_{c,\sigma}(x^j)  \right] \\
& = & \sum_{i,j=1}^N \mathop{\mathbb{E}}_{\sigma } \left[ \sigma_i \sigma_j \right ] \overline{\Phi}_{c,\sigma}(x^i) \overline{\Phi}_{c,\sigma}(x^j).
 \end{IEEEeqnarray} 
Since $\sigma_1,\cdots,\sigma_N$ are independent random variables drawn from the Rademacher distribution, we have
\begin{IEEEeqnarray}{rCl}
\mathop{\mathbb{E}}_{\sigma } \left[  \left | \sum_{i=1}^N \sigma_i \overline{\Phi}_{c,\sigma}(x^i) \right |^2 \right] & = & \sum_{i=1}^N \left |\overline{\Phi}_{c,\sigma}(x^i) \right|^2.
 \end{IEEEeqnarray} 
As $\overline{\Phi}_{c,\sigma} : \mathbb{R}^n \rightarrow [0,1]$, we have
\begin{IEEEeqnarray}{rCl}
\label{eq_121220240923}\mathop{\mathbb{E}}_{\sigma } \left[  \left | \sum_{i=1}^N \sigma_i \overline{\Phi}_{c,\sigma}(x^i) \right |^2 \right] & \leq & N.
 \end{IEEEeqnarray}
Using (\ref{eq_121220240923}) in (\ref{eq_121220240924}), we get
\begin{IEEEeqnarray}{rCl}
\label{eq_050120251512}\widehat{\mathcal{R}}_{\mathcal{D}}( \mathcal{M}_{c} ) & \leq & \frac{1}{ \sqrt{N}} + \frac{\epsilon}{N}.
 \end{IEEEeqnarray} 
Since the inequality (\ref{eq_050120251512}) holds for all $\epsilon > 0$, we have (\ref{eq_121220240925}).

\section*{Appendix M: Proof of Theorem~\ref{theorem_120220250928}}
Define, for a given dataset $\mathcal{D}$ (as defined in (\ref{eq_080620242021})),
\begin{IEEEeqnarray}{rCl}
\label{eq_301220241244}\widehat{\mathbb{E}}_{\mathcal{D}}(h_{x \mapsto y_c}) & = & \frac{1}{N} \sum_{i=1}^N  |y^i_c - h_{x \mapsto y_c}(x^i)|^2.
\end{IEEEeqnarray}
Consider a function assessing the supremum of difference of expected loss value from empirically averaged loss value:
\begin{IEEEeqnarray}{rCl}
\label{eq_301220241241}g_c(\mathcal{D}) & : = & \sup_{h_{x \mapsto y_c} \in \mathcal{M}_{c} } \left( \mathop{\mathbb{E}}_{(x,y) \sim \mathbb{P}_{x,y}} \left [ \left | y_c - h_{x \mapsto y_c}(x) \right |^2 \right ] - \widehat{\mathbb{E}}_{\mathcal{D}}(h_{x \mapsto y_c})\right). 
 \end{IEEEeqnarray}
Let $\mathcal{D}' = \left\{(x^1,y^1),\cdots, (x^{i-1},y^{i-1}),(x'^i,y'^i),(x^{i+1},y^{i+1}),\cdots, (x^N,y^N) \right\}$ be the {\em neighboring} set of $\mathcal{D}$ such that $\mathcal{D}'$ and $\mathcal{D}$ differ by only a single entry, i.e. the entry $(x'^i,y'^i) \notin \mathcal{D}$. As the difference of suprema can't exceed the supremum of the difference, we have
\begin{IEEEeqnarray}{rCl}
g_c(\mathcal{D}') - g_c(\mathcal{D}) & \leq & \sup_{h_{x \mapsto y_c} \in \mathcal{M}_{c} } \left( \frac{|y^i_c - h_{x \mapsto y_c}(x^i)|^2 - |y'^i_c - h_{x \mapsto y_c}(x'^i)|^2}{N}\right) \\
& \leq & \frac{1}{N},
 \end{IEEEeqnarray}
where we have used the facts that $y^i_c,y'^i_c \in \{0,1 \}$ and (\ref{eq_251220241757}). Similarly, we can obtain 
\begin{IEEEeqnarray}{rCl}
g_c(\mathcal{D}) - g_c(\mathcal{D}') & \leq & \frac{1}{N}.
\end{IEEEeqnarray}
Thus
\begin{IEEEeqnarray}{rCl}
\left|g_c(\mathcal{D}) - g_c(\mathcal{D}')\right| & \leq & \frac{1}{N}.
\end{IEEEeqnarray}
Thus, $g_c$ satisfies the bounded differences property with bound $1/N$, and therefore by Mc-Diarmid’s inequality, for any $\epsilon > 0$, with probability at most $\exp\left(- 2N \epsilon^2 \right)$, the following holds:
\begin{IEEEeqnarray}{rCl}
g_c(\mathcal{D}) - \mathop{\mathbb{E}}_{\mathcal{D} \sim (\mathbb{P}_{x,y})^N}\left[ g_c(\mathcal{D})\right] & \geq & \epsilon.
\end{IEEEeqnarray}
That is, with probability at most $\delta \in (0,1)$, the following holds: 
\begin{IEEEeqnarray}{rCl}
g_c(\mathcal{D}) - \mathop{\mathbb{E}}_{\mathcal{D} \sim (\mathbb{P}_{x,y})^N}\left[ g_c(\mathcal{D})\right] & \geq &  \sqrt{\frac{\log(1/\delta)}{2N}}.
\end{IEEEeqnarray}
In other words, with probability at least $1-\delta$, the following holds:
\begin{IEEEeqnarray}{rCl}
\label{eq_301220241200} g_c(\mathcal{D}) & \leq & \mathop{\mathbb{E}}_{\mathcal{D} \sim (\mathbb{P}_{x,y})^N}\left[ g_c(\mathcal{D})\right] +  \sqrt{\frac{\log(1/\delta)}{2N}}.
\end{IEEEeqnarray}
Let 
\begin{IEEEeqnarray}{rCl}
\tilde{\mathcal{D}} & = & \left \{(\tilde{x}^i,\tilde{y}^i)\; \mid \; i \in \{1,2,\cdots,N \} \right \} \sim (\mathbb{P}_{x,y})^N
\end{IEEEeqnarray}
be another set of IID samples and consider
\begin{IEEEeqnarray}{rCl}
\nonumber \lefteqn{\mathop{\mathbb{E}}_{\mathcal{D} \sim (\mathbb{P}_{x,y})^N}\left[ g_c(\mathcal{D})\right]}\\
& = & \mathop{\mathbb{E}}_{\mathcal{D} \sim (\mathbb{P}_{x,y})^N}\left[ \sup_{h_{x \mapsto y_c} \in \mathcal{M}_{c} } \left( \mathop{\mathbb{E}}_{(x,y) \sim \mathbb{P}_{x,y}} \left [ \left | y_c - h_{x \mapsto y_c}(x) \right |^2 \right ] - \widehat{\mathbb{E}}_{\mathcal{D}}(h_{x \mapsto y_c})\right) \right] \\
& = & \mathop{\mathbb{E}}_{\mathcal{D} \sim (\mathbb{P}_{x,y})^N}\left[ \sup_{h_{x \mapsto y_c} \in \mathcal{M}_{c} } \left( \mathop{\mathbb{E}}_{ \tilde{\mathcal{D}} \sim (\mathbb{P}_{x,y})^N} \left [ \widehat{\mathbb{E}}_{\tilde{\mathcal{D}}} \left( h_{x \mapsto y_c} \right) \right ] - \widehat{\mathbb{E}}_{\mathcal{D}}(h_{x \mapsto y_c})\right) \right] \\
& = & \mathop{\mathbb{E}}_{\mathcal{D} \sim (\mathbb{P}_{x,y})^N}\left[ \sup_{h_{x \mapsto y_c} \in \mathcal{M}_{c} } \left( \mathop{\mathbb{E}}_{ \tilde{\mathcal{D}} \sim (\mathbb{P}_{x,y})^N} \left [ \widehat{\mathbb{E}}_{\tilde{\mathcal{D}}} \left( h_{x \mapsto y_c} \right) - \widehat{\mathbb{E}}_{\mathcal{D}}(h_{x \mapsto y_c}) \right ] \right) \right] \\
& \leq & \mathop{\mathbb{E}}_{\mathcal{D} \sim (\mathbb{P}_{x,y})^N,\tilde{\mathcal{D}} \sim (\mathbb{P}_{x,y})^N } \left [ \sup_{h_{x \mapsto y_c} \in \mathcal{M}_{c} } \left( \widehat{\mathbb{E}}_{\tilde{\mathcal{D}}} \left( h_{x \mapsto y_c} \right) - \widehat{\mathbb{E}}_{\mathcal{D}}(h_{x \mapsto y_c}) \right) \right] \\
\label{eq_261220242202}& = & \mathop{\mathbb{E}}_{\mathcal{D} \sim (\mathbb{P}_{x,y})^N,\tilde{\mathcal{D}} \sim (\mathbb{P}_{x,y})^N } \left [ \sup_{h_{x \mapsto y_c} \in \mathcal{M}_{c} } \left( \frac{1}{N} \sum_{i=1}^N \left( |\tilde{y}^i_c - h_{x \mapsto y_c}(\tilde{x}^i)|^2 - |y^i_c - h_{x \mapsto y_c}(x^i)|^2 \right) \right) \right]. \IEEEeqnarraynumspace
\end{IEEEeqnarray}
Consider
\begin{IEEEeqnarray}{rCl}
\nonumber \lefteqn{\mathop{\mathbb{E}}_{\mathcal{D} \sim (\mathbb{P}_{x,y})^N,\tilde{\mathcal{D}} \sim (\mathbb{P}_{x,y})^N, \sigma } \left [ \sup_{h_{x \mapsto y_c} \in \mathcal{M}_{c} } \left( \frac{1}{N} \sum_{i=1}^N \sigma_i \left( |\tilde{y}^i_c - h_{x \mapsto y_c}(\tilde{x}^i)|^2 - |y^i_c - h_{x \mapsto y_c}(x^i)|^2 \right) \right) \right]}\\
\nonumber& = & \frac{1}{2} \mathop{\mathbb{E}}_{\mathcal{D} \sim (\mathbb{P}_{x,y})^N,\tilde{\mathcal{D}} \sim (\mathbb{P}_{x,y})^N } \left [ \sup_{h_{x \mapsto y_c} \in \mathcal{M}_{c} } \left( \frac{1}{N} \sum_{i=1}^N \left( |\tilde{y}^i_c - h_{x \mapsto y_c}(\tilde{x}^i)|^2 - |y^i_c - h_{x \mapsto y_c}(x^i)|^2 \right) \right) \right] \\
\label{eq_261220252144}& & + \frac{1}{2} \mathop{\mathbb{E}}_{\mathcal{D} \sim (\mathbb{P}_{x,y})^N,\tilde{\mathcal{D}} \sim (\mathbb{P}_{x,y})^N } \left [ \sup_{h_{x \mapsto y_c} \in \mathcal{M}_{c} } \left( \frac{1}{N} \sum_{i=1}^N \left( |y^i_c - h_{x \mapsto y_c}(x^i)|^2 - |\tilde{y}^i_c - h_{x \mapsto y_c}(\tilde{x}^i)|^2  \right) \right) \right] \IEEEeqnarraynumspace
\end{IEEEeqnarray}
where we have used the fact that $\sigma_1,\cdots,\sigma_N$ are Rademacher variables (i.e. taking value in $\{-1,1 \}$ with probability equal to 1/2). Further, due to the fact that
\begin{IEEEeqnarray}{rCl}
\lefteqn{\mathop{\mathbb{E}}_{\mathcal{D} \sim (\mathbb{P}_{x,y})^N,\tilde{\mathcal{D}} \sim (\mathbb{P}_{x,y})^N } \left [ \sup_{h_{x \mapsto y_c} \in \mathcal{M}_{c} } \left( \frac{1}{N} \sum_{i=1}^N \left( |y^i_c - h_{x \mapsto y_c}(x^i)|^2 - |\tilde{y}^i_c - h_{x \mapsto y_c}(\tilde{x}^i)|^2  \right) \right) \right]}\\
& = & \mathop{\mathbb{E}}_{\tilde{\mathcal{D}} \sim (\mathbb{P}_{x,y})^N ,\mathcal{D} \sim (\mathbb{P}_{x,y})^N} \left [ \sup_{h_{x \mapsto y_c} \in \mathcal{M}_{c} } \left( \frac{1}{N} \sum_{i=1}^N \left( |\tilde{y}^i_c - h_{x \mapsto y_c}(\tilde{x}^i)|^2 - |y^i_c - h_{x \mapsto y_c}(x^i)|^2 \right) \right) \right], \IEEEeqnarraynumspace
\end{IEEEeqnarray}
(\ref{eq_261220252144}) becomes
\begin{IEEEeqnarray}{rCl}
\nonumber \lefteqn{\mathop{\mathbb{E}}_{\mathcal{D} \sim (\mathbb{P}_{x,y})^N,\tilde{\mathcal{D}} \sim (\mathbb{P}_{x,y})^N, \sigma } \left [ \sup_{h_{x \mapsto y_c} \in \mathcal{M}_{c} } \left( \frac{1}{N} \sum_{i=1}^N \sigma_i \left( |\tilde{y}^i_c - h_{x \mapsto y_c}(\tilde{x}^i)|^2 - |y^i_c - h_{x \mapsto y_c}(x^i)|^2 \right) \right) \right]}\\
\label{eq_261220242201}& = & \mathop{\mathbb{E}}_{\mathcal{D} \sim (\mathbb{P}_{x,y})^N,\tilde{\mathcal{D}} \sim (\mathbb{P}_{x,y})^N } \left [ \sup_{h_{x \mapsto y_c} \in \mathcal{M}_{c} } \left( \frac{1}{N} \sum_{i=1}^N \left( |\tilde{y}^i_c - h_{x \mapsto y_c}(\tilde{x}^i)|^2 - |y^i_c - h_{x \mapsto y_c}(x^i)|^2 \right) \right) \right]. \IEEEeqnarraynumspace
\end{IEEEeqnarray}
Using (\ref{eq_261220242201}) in (\ref{eq_261220242202}), we have
\begin{IEEEeqnarray}{rCl}
\nonumber \lefteqn{\mathop{\mathbb{E}}_{\mathcal{D} \sim (\mathbb{P}_{x,y})^N}\left[ g_c(\mathcal{D})\right]}\\
& \leq & \mathop{\mathbb{E}}_{\mathcal{D} \sim (\mathbb{P}_{x,y})^N,\tilde{\mathcal{D}} \sim (\mathbb{P}_{x,y})^N, \sigma } \left [ \sup_{h_{x \mapsto y_c} \in \mathcal{M}_{c} } \left( \frac{1}{N} \sum_{i=1}^N \sigma_i \left( |\tilde{y}^i_c - h_{x \mapsto y_c}(\tilde{x}^i)|^2 - |y^i_c - h_{x \mapsto y_c}(x^i)|^2 \right) \right) \right] \IEEEeqnarraynumspace \\
& \leq & \mathop{\mathbb{E}}_{\tilde{\mathcal{D}} \sim (\mathbb{P}_{x,y})^N, \sigma } \left [ \sup_{h_{x \mapsto y_c} \in \mathcal{M}_{c} } \left( \frac{1}{N} \sum_{i=1}^N \sigma_i  |\tilde{y}^i_c - h_{x \mapsto y_c}(\tilde{x}^i)|^2   \right) \right] \\
& & + \mathop{\mathbb{E}}_{\mathcal{D} \sim (\mathbb{P}_{x,y})^N, \sigma } \left [ \sup_{h_{x \mapsto y_c} \in \mathcal{M}_{c} } \left( \frac{1}{N} \sum_{i=1}^N -\sigma_i   |y^i_c - h_{x \mapsto y_c}(x^i)|^2  \right) \right].
\end{IEEEeqnarray}
Since $\sigma_i$ and $-\sigma_i$ are identically distributed, we have
\begin{IEEEeqnarray}{rCl}
\mathop{\mathbb{E}}_{\mathcal{D} \sim (\mathbb{P}_{x,y})^N}\left[ g_c(\mathcal{D})\right] & \leq & 2\mathop{\mathbb{E}}_{\mathcal{D} \sim (\mathbb{P}_{x,y})^N, \sigma } \left [ \sup_{h_{x \mapsto y_c} \in \mathcal{M}_{c} } \left( \frac{1}{N} \sum_{i=1}^N \sigma_i   |y^i_c - h_{x \mapsto y_c}(x^i)|^2  \right) \right] \\
\label{eq_291220242219}& = & 2\mathop{\mathbb{E}}_{\mathcal{D} \sim (\mathbb{P}_{x,y})^N } \left[ \frac{1}{N}  \mathop{\mathbb{E}}_{\sigma } \left[ \sup_{h_{x \mapsto y_c} \in \mathcal{M}_{c} } \left( \sum_{i=1}^N \sigma_i   |y^i_c - h_{x \mapsto y_c}(x^i)|^2  \right)\right ]   \right].
\end{IEEEeqnarray}
Define
\begin{IEEEeqnarray}{rCl}
u_j(h_{x \mapsto y_c}) & := & \sum_{i=1}^j \sigma_i   |y^i_c - h_{x \mapsto y_c}(x^i)|^2
\end{IEEEeqnarray}
to express
\begin{IEEEeqnarray}{rCl}
\lefteqn{  \mathop{\mathbb{E}}_{\sigma } \left[ \sup_{h_{x \mapsto y_c} \in \mathcal{M}_{c} } \left( \sum_{i=1}^N \sigma_i   |y^i_c - h_{x \mapsto y_c}(x^i)|^2  \right) \right] }\\
& = &   \mathop{\mathbb{E}}_{\sigma_1,\cdots, \sigma_{N-1} } \left[ \mathop{\mathbb{E}}_{\sigma_N} \left [ \sup_{h_{x \mapsto y_c} \in \mathcal{M}_{c} } \left( u_{N-1}(h_{x \mapsto y_c}) +  \sigma_N   |y^N_c - h_{x \mapsto y_c}(x^N)|^2 \right) \right ]\right]
\end{IEEEeqnarray}
For any $\epsilon > 0$, let $h^1, h^2 \in \mathcal{M}_{c} $ be such that
\begin{IEEEeqnarray}{rCl}
\sup_{h_{x \mapsto y_c} \in \mathcal{M}_{c} } \left( u_{N-1}(h_{x \mapsto y_c}) +   |y^N_c - h_{x \mapsto y_c}(x^N)|^2 \right) & = & u_{N-1}(h^1)  +   |y^N_c - h^1(x^N)|^2 + \epsilon \\
\sup_{h_{x \mapsto y_c} \in \mathcal{M}_{c} } \left( u_{N-1}(h_{x \mapsto y_c}) -   |y^N_c - h_{x \mapsto y_c}(x^N)|^2 \right) & = & u_{N-1}(h^2)  -   |y^N_c - h^2(x^N)|^2 + \epsilon.
\end{IEEEeqnarray}
Now, consider
\begin{IEEEeqnarray}{rCl}
\lefteqn{\mathop{\mathbb{E}}_{\sigma_N} \left [ \sup_{h_{x \mapsto y_c} \in \mathcal{M}_{c} } \left( u_{N-1}(h_{x \mapsto y_c}) +  \sigma_N   |y^N_c - h_{x \mapsto y_c}(x^N)|^2 \right) \right ]} \\
 & = & \frac{1}{2} \sup_{h_{x \mapsto y_c} \in \mathcal{M}_{c} } \left( u_{N-1}(h_{x \mapsto y_c}) +     |y^N_c - h_{x \mapsto y_c}(x^N)|^2 \right) + \frac{1}{2} \sup_{h_{x \mapsto y_c} \in \mathcal{M}_{c} } \left( u_{N-1}(h_{x \mapsto y_c}) -   |y^N_c - h_{x \mapsto y_c}(x^N)|^2 \right) \IEEEeqnarraynumspace \\
 & = & \frac{1}{2} \left(u_{N-1}(h^1) + u_{N-1}(h^2) \right) + \frac{1}{2}\left( (h^1(x^N) + h^2(x^N) - 2 y_c^N) (h^1(x^N) - h^2(x^N)) \right) + \epsilon. 
\end{IEEEeqnarray}
Define
\begin{IEEEeqnarray}{rCl}
\eta & = & \sign\left( h^1(x^N) - h^2(x^N) \right),
\end{IEEEeqnarray}
and consider (\ref{eq_251220241757}) and $y_c^N \in \{0,1 \}$, leading to
\begin{IEEEeqnarray}{rCl}
h^1(x^N) + h^2(x^N) - 2 y_c^N & \leq & 2, 
\end{IEEEeqnarray}
so that
\begin{IEEEeqnarray}{rCl}
\nonumber \lefteqn{\mathop{\mathbb{E}}_{\sigma_N} \left [ \sup_{h_{x \mapsto y_c} \in \mathcal{M}_{c} } \left( u_{N-1}(h_{x \mapsto y_c}) +  \sigma_N   |y^N_c - h_{x \mapsto y_c}(x^N)|^2 \right) \right ]} \\
& \leq & \frac{1}{2} \left(u_{N-1}(h^1) + u_{N-1}(h^2) \right) + \frac{1}{2}\left( 2  \eta  (h^1(x^N) - h^2(x^N)) \right) + \epsilon \\
& = & \frac{1}{2} \left(u_{N-1}(h^1) + 2  \eta h^1(x^N) \right) + \frac{1}{2} \left(  u_{N-1}(h^2) - 2  \eta h^2(x^N) \right) + \epsilon \\
 & \leq & \frac{1}{2} \sup_{h_{x \mapsto y_c} \in \mathcal{M}_{c} } \left( u_{N-1}(h_{x \mapsto y_c}) + 2  \eta h_{x \mapsto y_c}(x^N)  \right)  + \frac{1}{2} \sup_{h_{x \mapsto y_c} \in \mathcal{M}_{c} } \left( u_{N-1}(h_{x \mapsto y_c}) - 2  \eta h_{x \mapsto y_c}(x^N) \right) + \epsilon \\
& = & \mathop{\mathbb{E}}_{\sigma_N} \left[ \sup_{h_{x \mapsto y_c} \in \mathcal{M}_{c} } \left( u_{N-1}(h_{x \mapsto y_c}) + \sigma_N 2   h_{x \mapsto y_c}(x^N) \right)  \right] + \epsilon.
\end{IEEEeqnarray}
Since the inequality holds for all $\epsilon > 0$, we have
\begin{IEEEeqnarray}{rCl}
\nonumber \lefteqn{\mathop{\mathbb{E}}_{\sigma_N} \left [ \sup_{h_{x \mapsto y_c} \in \mathcal{M}_{c} } \left( u_{N-1}(h_{x \mapsto y_c}) +  \sigma_N   |y^N_c - h_{x \mapsto y_c}(x^N)|^2 \right) \right ]} \\
\label{eq_291220242052}& \leq & \mathop{\mathbb{E}}_{\sigma_N} \left[ \sup_{h_{x \mapsto y_c} \in \mathcal{M}_{c} } \left( u_{N-1}(h_{x \mapsto y_c}) + \sigma_N 2   h_{x \mapsto y_c}(x^N) \right)  \right].
\end{IEEEeqnarray}
Equivalently,
\begin{IEEEeqnarray}{rCl}
\mathop{\mathbb{E}}_{\sigma_N} \left [ \sup_{h_{x \mapsto y_c} \in \mathcal{M}_{c} } \left( \sum_{i=1}^N \sigma_i   |y^i_c - h_{x \mapsto y_c}(x^i)|^2 \right) \right ]
& \leq & \mathop{\mathbb{E}}_{\sigma_N} \left[ \sup_{h_{x \mapsto y_c} \in \mathcal{M}_{c} } \left( u_{N-1}(h_{x \mapsto y_c}) + \sigma_N 2   h_{x \mapsto y_c}(x^N) \right)  \right].
\end{IEEEeqnarray}
That is,
\begin{IEEEeqnarray}{rCl}
\nonumber \lefteqn{\mathop{\mathbb{E}}_{\sigma_{N-1},\sigma_N} \left [ \sup_{h_{x \mapsto y_c} \in \mathcal{M}_{c} } \left( \sum_{i=1}^N \sigma_i   |y^i_c - h_{x \mapsto y_c}(x^i)|^2 \right) \right ]} \\
& \leq & \mathop{\mathbb{E}}_{\sigma_N} \left[ \mathop{\mathbb{E}}_{\sigma_{N-1}} \left[ \sup_{h_{x \mapsto y_c} \in \mathcal{M}_{c} } \left( u_{N-2}(h_{x \mapsto y_c}) + \sigma_{N-1}   |y^{N-1}_c - h_{x \mapsto y_c}(x^{N-1})|^2 + \sigma_N 2   h_{x \mapsto y_c}(x^N)  \right) \right] \right ] \IEEEeqnarraynumspace
\end{IEEEeqnarray}
Following the same procedure for $\sigma_{N-1}$, as that for $\sigma_N$ to derive the inequality (\ref{eq_291220242052}), we will get
\begin{IEEEeqnarray}{rCl}
\nonumber \lefteqn{\mathop{\mathbb{E}}_{\sigma_{N-1}} \left[ \sup_{h_{x \mapsto y_c} \in \mathcal{M}_{c} } \left( u_{N-2}(h_{x \mapsto y_c}) + \sigma_{N-1}   |y^{N-1}_c - h_{x \mapsto y_c}(x^{N-1})|^2 + \sigma_N 2   h_{x \mapsto y_c}(x^N)  \right) \right]} \\
& \leq & \mathop{\mathbb{E}}_{\sigma_{N-1}} \left[ \sup_{h_{x \mapsto y_c} \in \mathcal{M}_{c} } \left( u_{N-2}(h_{x \mapsto y_c}) + \sigma_{N-1} 2  h_{x \mapsto y_c}(x^{N-1}) + \sigma_N 2   h_{x \mapsto y_c}(x^N)   \right) \right],
\end{IEEEeqnarray}
resulting in
\begin{IEEEeqnarray}{rCl}
\nonumber \lefteqn{\mathop{\mathbb{E}}_{\sigma_{N-1},\sigma_N} \left [ \sup_{h_{x \mapsto y_c} \in \mathcal{M}_{c} } \left( \sum_{i=1}^N \sigma_i   |y^i_c - h_{x \mapsto y_c}(x^i)|^2 \right) \right ]} \\
& \leq & \mathop{\mathbb{E}}_{\sigma_N} \left[ \mathop{\mathbb{E}}_{\sigma_{N-1}} \left[ \sup_{h_{x \mapsto y_c} \in \mathcal{M}_{c} } \left( u_{N-2}(h_{x \mapsto y_c}) + \sigma_{N-1} 2  h_{x \mapsto y_c}(x^{N-1}) + \sigma_N 2   h_{x \mapsto y_c}(x^N)   \right) \right] \right ] 
\end{IEEEeqnarray}
Following the same procedure for $\sigma_{N-2},\cdots,\sigma_1$, we will get
\begin{IEEEeqnarray}{rCl}
\mathop{\mathbb{E}}_{\sigma} \left [ \sup_{h_{x \mapsto y_c} \in \mathcal{M}_{c} } \left( \sum_{i=1}^N \sigma_i   |y^i_c - h_{x \mapsto y_c}(x^i)|^2 \right) \right ] & \leq & 2  \mathop{\mathbb{E}}_{\sigma} \left [ \sup_{h_{x \mapsto y_c} \in \mathcal{M}_{c} } \left(  \sum_{i=1}^N \sigma_i  h_{x \mapsto y_c}(x^i) \right) \right ] \\
& = & 2 N \widehat{\mathcal{R}}_{\mathcal{D}}( \mathcal{M}_{c} ).
\end{IEEEeqnarray}
Using (\ref{eq_121220240925}), we get
\begin{IEEEeqnarray}{rCl}
\label{eq_291220242220}\mathop{\mathbb{E}}_{\sigma} \left [ \sup_{h_{x \mapsto y_c} \in \mathcal{M}_{c} } \left( \sum_{i=1}^N \sigma_i   |y^i_c - h_{x \mapsto y_c}(x^i)|^2 \right) \right ] & \leq & 2  \sqrt{N}.
\end{IEEEeqnarray}
Combining (\ref{eq_291220242220}) and (\ref{eq_291220242219}), we have
\begin{IEEEeqnarray}{rCl}
\label{eq_301220241201}\mathop{\mathbb{E}}_{\mathcal{D} \sim (\mathbb{P}_{x,y})^N}\left[ g_c(\mathcal{D})\right] & \leq &   \frac{4}{\sqrt{N}}  .
\end{IEEEeqnarray}
Using (\ref{eq_301220241201}) in (\ref{eq_301220241200}), we have with probability at least $1-\delta$, 
\begin{IEEEeqnarray}{rCl}
g_c(\mathcal{D}) & \leq &  \frac{4}{ \sqrt{N}} + \sqrt{\frac{\log(1/\delta)}{2N}}.
\end{IEEEeqnarray}
Considering the definition of $g_c$ (as given in (\ref{eq_301220241241})) and (\ref{eq_301220241244}), we have with probability at least $1-\delta$, 
\begin{IEEEeqnarray}{rCl}
\label{eq_161220241845}\mathop{\mathbb{E}}_{(x,y) \sim \mathbb{P}_{x,y}} \left [ \left | y_c - h_{x \mapsto y_c}(x) \right |^2  \right] & \leq & \frac{1}{N} \sum_{i=1}^N |y_c^i - h_{x \mapsto y_c}(x^i) |^2 +  \frac{4}{\sqrt{N}} +  \sqrt{\frac{\log(1/\delta)}{2N}}.
 \end{IEEEeqnarray} 
Using (\ref{eq_100220251930}) together with Theorem~\ref{theorem_090220251718}, we have with probability at least $1-\delta$:
\begin{IEEEeqnarray}{rCl}
 \label{eq_120220251947} \mathop{\mathbb{E}}_{(x,y) \sim \mathbb{P}_{x,y}} \left [ \left | y_c - h_{x \mapsto y_c}(x) \right |^2 \right]  & \leq &   \mathop{\mathbb{E}}_{(x,y) \sim \mathbb{P}_{x,y}} \left [ \left | y_c - \mathop{\mathbb{E}}_{y \sim \mathbb{P}_{y | x}} \left[ y_c | x  \right] \right |^2 \right] + \frac{1}{\left(N_c/N \right)^2} \left( \frac{3}{\sqrt{N}} + \sqrt{\frac{8  \log(1/\delta)}{N}} \right).
   \end{IEEEeqnarray}
Combining (\ref{eq_161220241845}) and (\ref{eq_120220251947}) leads to the result.

\section*{Appendix N: Proof of Theorem~\ref{theorem_120220251923}}
Consider
 \begin{IEEEeqnarray}{rCl}
\nonumber \lefteqn{\mathop{\mathbb{E}}_{(x,y) \sim \mathbb{P}_{x,y}} \left [ \left | y_c - h_{x \mapsto y_c}(x) \right |^2 \right] - \mathop{\mathbb{E}}_{(x,y) \sim \mathbb{P}_{x,y}} \left [ \left | y_c - \mathop{\mathbb{E}}_{y \sim \mathbb{P}_{y | x}} \left[ y_c | x  \right] \right |^2 \right]}\\
 & = & \mathop{\mathbb{E}}_{(x,y) \sim \mathbb{P}_{x,y}} \left [ \left | y_c - h_{x \mapsto y_c}(x) \right |^2 \right] - \mathop{\mathbb{E}}_{(x,y) \sim \mathbb{P}_{x,y}} \left [ \left | y_c - \mathbb{P}_{y | x}(y_c = 1 | x) \right |^2 \right] \\
 & = & \mathop{\mathbb{E}}_{(x,y) \sim \mathbb{P}_{x,y}} \left [ \left| h_{x \mapsto y_c}(x) \right|^2 - \left| \mathbb{P}_{y | x}(y_c = 1 | x) \right|^2 - 2y_c(h_{x \mapsto y_c}(x) - \mathbb{P}_{y | x}(y_c = 1 | x)) \right ] \\
 & = & \mathop{\mathbb{E}}_{x \sim \mathbb{P}_x}\left[ \left| h_{x \mapsto y_c}(x) \right|^2\right] - \mathop{\mathbb{E}}_{x \sim \mathbb{P}_x}\left[ \left| \mathbb{P}_{y | x}(y_c = 1 | x) \right|^2 \right ]  - 2 \mathop{\mathbb{E}}_{x \sim \mathbb{P}_x}\left[ \mathbb{P}_{y | x}(y_c = 1 | x) (h_{x \mapsto y_c}(x) - \mathbb{P}_{y | x}(y_c = 1 | x))  \right] \\
\label{eq_070120251035}& = & \mathop{\mathbb{E}}_{x \sim \mathbb{P}_x}\left[ \left| h_{x \mapsto y_c}(x) - \mathbb{P}_{y | x}(y_c = 1 | x) \right|^2\right].
 \end{IEEEeqnarray} 
It follows from (\ref{eq_070120251035}) that 
\begin{IEEEeqnarray}{rCl}
\label{eq_120220252031}\mathop{\mathbb{E}}_{x \sim \mathbb{P}_x}\left[ \left| h_{x \mapsto y_c}(x) - \mathbb{P}_{y | x}(y_c = 1 | x) \right|^2\right] & \leq & \mathop{\mathbb{E}}_{(x,y) \sim \mathbb{P}_{x,y}} \left [ \left | y_c - h_{x \mapsto y_c}(x) \right |^2  \right].
 \end{IEEEeqnarray}
Since (\ref{eq_161220241845}) holds with probability at least $1-\delta$, using (\ref{eq_120220252031}), we have with probability at least $1-\delta$:  
\begin{IEEEeqnarray}{rCl}
\label{eq_120220252037}\mathop{\mathbb{E}}_{x \sim \mathbb{P}_x}\left[ \left| h_{x \mapsto y_c}(x) - \mathbb{P}_{y | x}(y_c = 1 | x) \right|^2\right] & \leq & \frac{1}{N} \sum_{i=1}^N |y_c^i - h_{x \mapsto y_c}(x^i) |^2 +  \frac{4}{\sqrt{N}} +  \sqrt{\frac{\log(1/\delta)}{2N}}. \IEEEeqnarraynumspace
 \end{IEEEeqnarray}
Using (\ref{eq_100220251930}) together with Theorem~\ref{theorem_100220251831}, we have with probability at least $1-\delta$: 
\begin{IEEEeqnarray}{rCl}
\label{eq_130220250847}\mathop{\mathbb{E}}_{x \sim \mathbb{P}_x}\left[ \left| h_{x \mapsto y_c}(x) - \mathbb{P}_{y | x}(y_c = 1 | x) \right|^2\right] & \leq & \frac{1}{\left(N_c/N \right)^2} \left( \frac{3}{\sqrt{N}} + \sqrt{\frac{8  \log(1/\delta)}{N}} \right).
   \end{IEEEeqnarray}
Combining (\ref{eq_120220252037}) and (\ref{eq_130220250847}) leads to the result.

\bibliographystyle{ACM-Reference-Format}  
\bibliography{references}

\end{document}